\documentclass{article}

\PassOptionsToPackage{numbers, compress}{natbib}

 \usepackage[main, final]{neurips_2026}

\usepackage[utf8]{inputenc} 
\usepackage[T1]{fontenc}    
\usepackage{url}            
\usepackage{booktabs}       
\usepackage{amsfonts}       
\usepackage{nicefrac}       
\usepackage{url}
\usepackage{anyfontsize}

\usepackage{appendix}
\usepackage{titletoc}
\usepackage{enumitem}
\usepackage{tabularx}    
\usepackage{float}
\usepackage[table]{xcolor}
\usepackage{booktabs}    
\usepackage{caption}     
\usepackage{lipsum}      
\usepackage{wrapfig}      
\usepackage{kotex}
\usepackage{array}
\usepackage{graphicx}
\usepackage{subcaption}
\usepackage{multirow}
\usepackage[most]{tcolorbox}

\definecolor{citecolor}{HTML}{0071bc}
\usepackage[
    breaklinks=true,
    colorlinks=true,
    citecolor=citecolor
]{hyperref}

\usepackage{cleveref}

\definecolor{darkgreen}{RGB}{0,100,0} 

\definecolor{gtgreen}{HTML}{006400}
\definecolor{deepred}{rgb}{0.7,0.0,0.0}
\definecolor{mildyellow}{rgb}{0.92,0.9,0.88}
\definecolor{steelblue}{rgb}{0.27,0.51,0.71}
\definecolor{darkcyan}{rgb}{0.0,0.55,0.55}

\definecolor{orange}{rgb}{1.0,0.45,0.0}
\definecolor{navy}{rgb}{0.0,0.0,0.5}

\title{VLMs Trace Without Tracking: \\ Diagnosing Failures in Visual Path Following}

\author{
Hyesoo Hong\textsuperscript{1} \hspace{0.5em}
Min Soo Kim\textsuperscript{1} \hspace{0.5em}
Wonje Jeung\textsuperscript{1} \\ [0.5em]
\textbf{Sangyeon Yoon\textsuperscript{1}} \hspace{0.5em}
\textbf{Dongjae Jeon\textsuperscript{2}} \hspace{0.5em}
\textbf{Albert No\textsuperscript{1}}\thanks{Correspondence to: Albert No \texttt{albertno@yonsei.ac.kr}}
\\[0.5em]
\textsuperscript{1}Department of Artificial Intelligence, Yonsei University \\
\textsuperscript{2}Department of Computer Science, Yonsei University
}

\begin{document}

\maketitle

\begin{abstract}
Vision-language models (VLMs) perform strongly on multimodal benchmarks, yet their ability to reliably follow continuous visual structure remains unclear. We focus on \textit{path following}, a basic operation that requires following a selected path through successive visual continuations. Before considering complex factors such as crossings or overlaps, we first test whether current VLMs can robustly follow a simple continuous line under controlled settings that remove such confounds. Across frontier models, we find that tracing is already fragile, with this vulnerability manifesting as unstable path preservation in the presence of other nearby paths. Analysis of internal representations shows that information needed to determine the correct path continuation becomes weaker, with this degradation already emerging in the visual encoding stage. Common remedies do not resolve this weakness: model scaling provides limited improvement, additional reasoning provides uneven gains at substantial cost, and explicit tracing instructions fail to stabilize performance. The same path-preservation failures and sensitivity to nearby paths persist in visually richer and more realistic settings, suggesting that current VLMs lack a robust mechanism for following a selected path.
\end{abstract}

\section{Introduction}

VLMs now achieve strong performance on many multimodal benchmarks~\citep{liu2023mmbench,lu2024mathvista,yue2024mmmu}, yet this progress does not imply mastery of basic visual operations. A growing body of work shows that even frontier VLMs fail on visually simple tasks that humans solve easily~\citep{fu2024blink, chandhok2025response,hou2024vision,huang2025vision,tong2024eyes}, often because models lean on language priors, semantic context, or coarse image statistics rather than genuine visual analysis~\citep{guan2024hallusionbench,lee2025vlind,luo2024probing,zhou2025visualcorrespondence,vo2025vision}. Elementary capabilities such as individuating objects, sustaining attention across space, and following continuous structure remain underdeveloped~\citep{campbell2024understanding,rahmanzadehgervi2024vision,chen2025spatial,chen2026babyvision}.

In this work, we investigate whether VLMs can reliably follow a visual path, beginning with its simplest form: a continuous line. This operation matters disproportionately because VLMs are increasingly deployed in settings where they must follow connected visual paths, such as following a route on a transit map~\citep{mukhopadhyay2024unraveling, feng2025rewardmap}, tracking a deformable object through a cluttered scene~\citep{viswanath2023handloom}, or reading a flow chart~\citep{chen2024spatialvlm, hou2024vision, li2024visiongraph, tang2026chartmuseum, zhang2025mllms}. The shared primitive across these tasks is simple: Given a starting point, the model must stay on the selected path as it unfolds through space. As such applications become more common, it is important to ask whether VLMs can reliably perform this basic visual operation rather than merely succeed on high-level task outputs.

Prior works frame VLM failures in path following as limits of nonlocal reasoning under difficult conditions such as long paths, crossings, and visual shortcuts~\citep{berman2025vlms, chen2026babyvision}. Before attributing failures to such complex factors, however, it is necessary to first establish whether VLMs possess the more fundamental ability to follow a simple continuous line. We therefore evaluate six frontier VLMs---Claude Sonnet-4.5, GPT-5.4, Gemini3-Flash, Kimi-K3, GLM-4.6V, and Qwen3-VL-235B---under strictly controlled settings and find that most already struggle to maintain the selected path in the mere presence of nearby paths, even without crossings or overlaps.

\begin{figure}[t!]
\centering
\includegraphics[width=\textwidth]{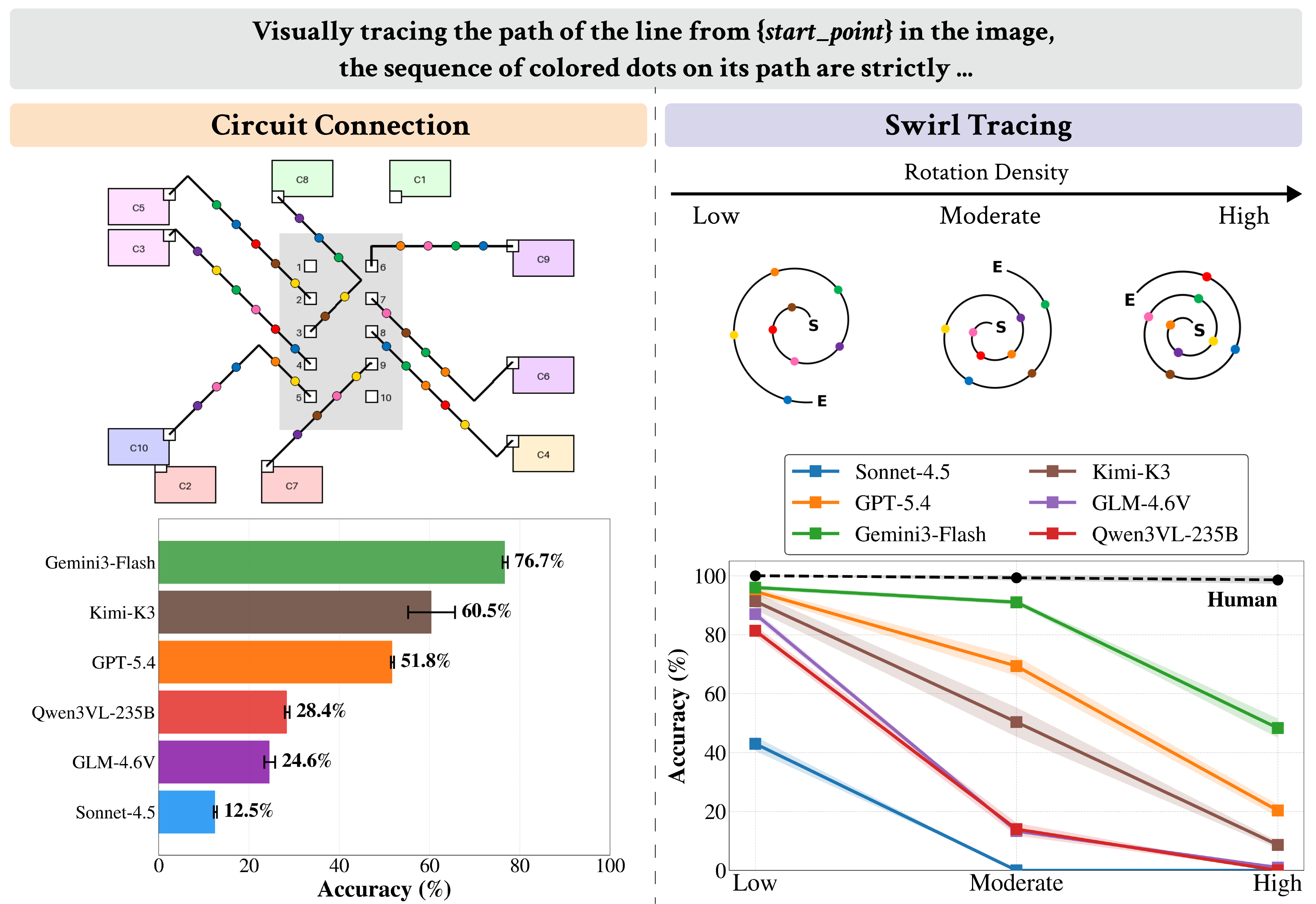}
\caption{
\textbf{Two tracing settings and the corresponding model performance.}
In both settings, the prompt asks the model to begin from a specified start point and output the sequence of colored dots along the corresponding path in order.
Left: Example of the \textit{Circuit Connections} task and model performance across all evaluated frontier models.
Right: Examples of \textit{Swirl} tracing at \textit{Low}, \textit{Moderate}, and \textit{High} rotation density and model performance across these three levels.
}
\label{fig:overview}
\vspace{-1.0em}
\end{figure}

Specifically, we first adapt the \textit{Circuit Connections} task from prior path-following work~\citep{berman2025vlms}, removing crossings, overlaps, and semantic shortcuts to isolate basic path following. Despite this simplification, all models remain below 80\% accuracy, with wide variation across models, showing that preserving a selected path is already unreliable. This failure often manifests as an \textit{adjacent-wire jump}, where the model begins on the correct wire but later switches to a nearby wire and continues toward its endpoint. Motivated by this sensitivity to nearby paths, we design the \textit{Swirl} dataset to systematically stress path preservation within one continuous path by varying rotation density. Human accuracy stays near 99\% across densities, whereas model accuracy drops by more than 40 percentage points from \textit{Low} to \textit{High} density. This loss appears as increasingly frequent \textit{adjacent-turn jumps}, where models leave the selected spiral path for a neighboring turn at the same angular direction. Together, these results reveal a lack of robust path following, exposed through the systematic destabilization of path preservation when nearby competing paths are introduced.

Our claim is supported by internal analysis and standard-remedy tests. Internal analyses show that next-step information for tacing weakens under controlled variations in nearby-path presence and appearance similarity across the vision encoder and LLM. Standard remedies do not resolve this limitiation: scaling the vision encoder provides limited improvement in connectivity preservation; reasoning shifts from the required tracing procedure toward alternative strategies at substantial computational cost; and explicit tracing instructions fail to stabilize tracing performance. Notably, model-generated tracing instructions closely resemble human tracing procedures, yet yield only limited gains when fed back to the same model, revealing a gap between high-level tracing knowledge and stable visual execution. Case studies on deformable line objects and metro maps show that the same tendency persists in visually richer and more realistic settings.

Overall, our findings from using nearby paths as a stressor provide converging behavioral and representational evidence of a limitation in VLM tracing robustness. Its persistence across controlled perturbations, standard remedies, and realistic settings suggests that overcoming this limitation will require more fundamental solutions than scaling, reasoning, or prompting alone.

\section{Related Work}

\paragraph{Basic Visual Abilities in VLMs.}
Recent work shows that strong performance on broad multimodal benchmarks can coexist with failures on visually simple tasks. Prior studies have documented broad shortcomings in the visual abilities of multimodal models~\citep{al2024unibench,fu2024blink,tong2024eyes,chandhok2025response,fu2025hidden,hou2024vision,luo2026vision}, as well as failures on basic geometric judgments, visual-only perception, and core visual tasks designed to reduce linguistic shortcuts~\citep{rahmanzadehgervi2024vision,kamoi2025visonlyqa,chen2026babyvision}. Related analyses connect such failures to limitations in binding, correspondence, and spatial reasoning~\citep{kamath2023whatsup,campbell2024understanding,chen2025spatial,zhou2025visualcorrespondence,assouel2025visual}.
These findings motivate controlled evaluations that minimize semantic and linguistic shortcuts to probe basic visual abilities. We adopt this approach to study a capability: maintaining a selected path through nearby competing continuations.

\paragraph{Path Following and Nonlocal Visual Reasoning.}
A related line of work studies path following as a form of nonlocal visual reasoning, where success depends on integrating evidence across distant image regions. Prior work identifies smooth visual search as a weakness, including contour-following and circuit-connection tasks~\citep{berman2025vlms}; map-based reasoning tasks also test route planning over transit maps~\citep{feng2025rewardmap}. Contour following has been studied as a long-range dependency in \emph{Pathfinder}~\citep{linsley2018learning} and as deformable-object tracing in \emph{HANDLOOM}~\citep{viswanath2023handloom}. Concurrent work, \emph{TraversalBench}~\citep{petrova2026traversalbenchchallengingpathsfollow}, studies exact path traversal across factors such as crossings, tortuosity, vertex count, and confounding lines. Our work is complementary, isolating path preservation in a crossing-free, color-only setting and analyzing failures through controlled similarity, internal-state, and reasoning analyses.

\paragraph{Tracing as a Sequential Visual Operation.}
Classic work on visual cognition provides a precedent for viewing tracing as a sequential operation over connected structure. One account proposes \textit{incremental grouping}: while simple structures can be grouped using local visual features, nearby competing paths introduce clutter and ambiguous continuations, making incremental grouping important for maintaining the selected path through successive local choices~\citep{roelfsema2006cortical,roelfsema2011incremental}. Behavioral curve-tracing studies show that tracing depends on serial attentional processing and becomes harder as paths become longer, more cluttered, or more ambiguous~\citep{jolicoeur1986curvetracing,pringle1988mentalcurvetracing,scholte2001spatial,houtkamp2003gradual}. We use this literature as conceptual grounding: it motivates interpreting VLM tracing errors as failures to remain on the selected path when nearby paths offer competing continuations, rather than as generic spatial mistakes.

\section{VLM Tracing Lacks Robust Path Preservation}
\label{tracing_ex}

Before evaluating path following in more complex visual settings, we first ask whether VLMs can follow its simplest form: a single continuous line. We therefore begin with a simplified adaptation of the \textit{Circuit Connections} task~\citep{berman2025vlms} to examine basic path following under minimal visual complexity as an initial robustness check. We then introduce \textit{Swirl} to systematically reproduce the failure factor observed in \textit{Circuit Connections} as a controlled stressor and examine how path preservation changes as its strength is progressively increased.

\subsection{VLMs Fail Even at Basic Path Following}
\label{circuit}

\paragraph{Path Following Under Minimal Visual Complexity.}
To isolate basic path following, we adapt the \textit{Circuit Connections} task by removing wire crossings and overlaps, as shown in \Cref{fig:overview} (left), and place colored dots along each wire to reveal the model's followed path. Given a queried wire, the model must follow it from a port on the breadboard to its endpoint component and report every colored dot encountered in order. A prediction is counted as correct only if it reports the complete ordered sequence and terminates with \textit{``end''}. Detailed task construction is provided in Appendix~\ref{app:circuit_construction}.

As shown in \Cref{fig:overview} (left), performance remains limited across all models even in this simplified path-following setting. The best-performing model, Gemini3-Flash, reaches 76.7\% accuracy, while GPT-5.4 achieves 51.8\% and Sonnet-4.5 performs lowest at 12.5\%. Even the strongest model fails on nearly one quarter of traces, while the wide performance spread shows that basic path following remains unreliable across frontier VLMs. Understanding what drives these failures is therefore critical for identifying the underlying weakness in path preservation more precisely.


\begin{figure}[t]
    \centering
    \includegraphics[width=\linewidth]{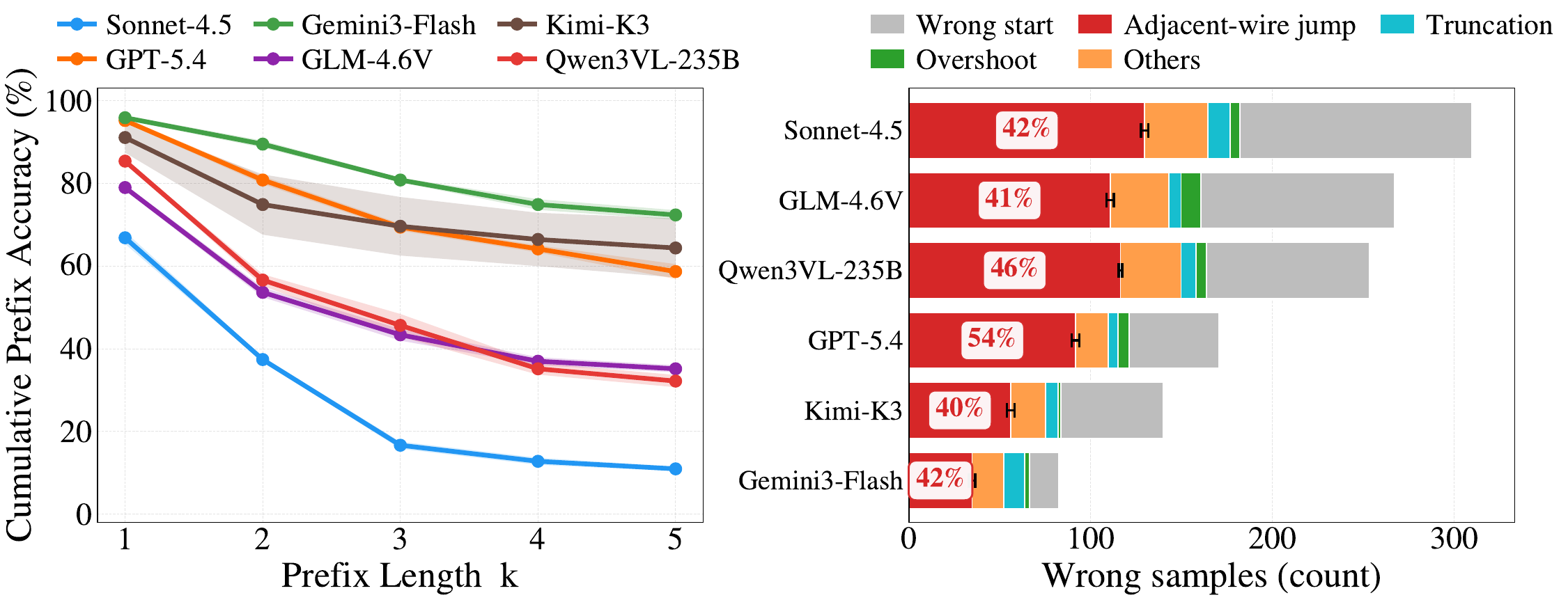}
    \caption{
\textbf{Path-preservation and error analysis on Circuit Connections.}
Left: Cumulative prefix accuracy across tracing steps for wires containing five colored dots across all evaluated models.
Right: Error-type distribution for failed traces. Errors are assigned sequentially to the first applicable category for each failed trace: \textit{Wrong start} when the initial dot is incorrect; \textit{Adjacent-wire jump} when the trace switches to a nearby wire within the predefined pixel-distance threshold; \textit{Truncation} when the queried wire is followed correctly but the prediction terminates early; \textit{Overshoot} when the correct sequence is completed but additional dots are generated; and \textit{Others} for all remaining errors.
}
\label{fig:circuit_path_preservation}
\vspace{-0.5em}
\end{figure}

\paragraph{Path Preservation Breaks Down During Tracing.}
To understand the source of this unstable performance, we analyze cumulative accuracy as tracing progresses and the types of errors that occur. Cumulative prefix accuracy at step $k$ is defined as the proportion of traces whose first $k$ predicted dots exactly match the ground-truth sequence. Therefore, an error at any earlier step also makes all subsequent prefix steps incorrect. As shown in \Cref{fig:circuit_path_preservation} (left), models often recover the initial segment correctly, but cumulative prefix accuracy consistently declines as tracing proceeds, even under this simplified setting. This pattern shows that extending the correct sequence becomes progressively more difficult as tracing proceeds. Full cumulative prefix accuracy results for wires containing different numbers of dots are provided in Appendix~\ref{app:circuit_prefix}.

We next analyze the error types underlying this cumulative decline. As shown in \Cref{fig:circuit_path_preservation} (right), \textit{Wrong start} errors are common, indicating that some failures occur before tracing is correctly initialized. Our main focus, however, is the \textit{Adjacent-wire jump}, where a model starts on the correct wire but subsequently switches to a nearby wire and continues tracing from there. These jumps account for 40--54\% of all errors across models, making them one of the dominant failure modes. Thus, even when the correct path is initially identified, models frequently fail to preserve it by switching to nearby paths. Details of the error-type classification are provided in Appendix~\ref{app:circuit_error_analysis}.

\subsection{Swirl: A Controlled Stress Test of Path Preservation}
\label{sec:swirl}

\paragraph{Systematically Varying Nearby-Path Proximity.}
Previous recurring failure makes nearby paths a natural controlled stressor for testing whether VLM path following remains stable. Therefore, we design \textit{Swirl} as a controlled stress test of path preservation. Using a single continuous spiral and keeping path length, stroke width, and dot count fixed, we vary the spacing between neighboring turns by progressively increasing the number of turns. We use this inter-turn spacing to define \textit{Low}, \textit{Moderate}, and \textit{High} rotation-density levels (\Cref{fig:overview}, right), with smaller spacing corresponding to higher density and a stronger nearby-path stressor. The design also avoids introducing multiple independent paths and removes factors unrelated to path following, such as domain-specific objects, labels, ports, and background structure. Each image contains the same eight uniquely colored, non-repeating dots as trace verifiers. Detailed dataset construction is provided in Appendix~\ref{app:swirl_task_construction}.

\begin{figure}[t]
    \centering
    \includegraphics[width=0.80\linewidth]{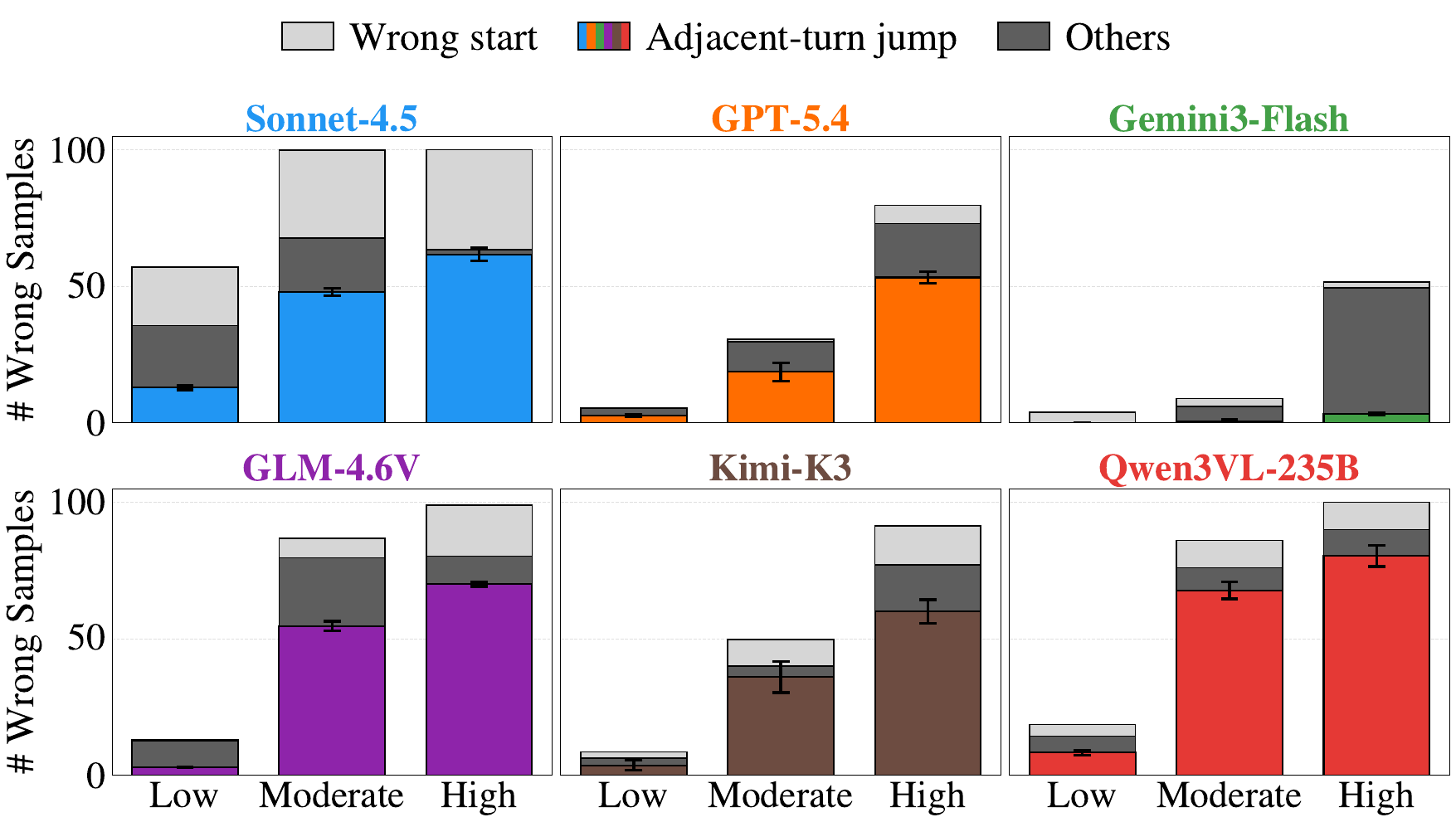}
    \caption{
    \textbf{Error analysis on Swirl.}
    Shown errors are assigned sequentially to the first applicable category: \textit{Wrong start} when the initial dot is incorrect; \textit{Adjacent-turn jump} when the trace moves to a dot located at the same angular direction on the immediately adjacent outer turn of the spiral; and \textit{Others} for the remaining errors.
    \textit{Truncation} and \textit{Overshoot} were rarely observed and are  omitted.
    }
    \vspace{-0.5em}
    \label{fig:swirl_path_preservation}
\end{figure}

\paragraph{Tracing Degrades as the Nearby-Path Stressor Strengthens.}
If models possessed robust path-following ability, increasing rotation density should not systematically disrupt path preservation. As shown in \Cref{fig:overview} (right), the task remains trivial for humans across density levels, with accuracy changing only from 100.0\% to 98.6\%. Against this baseline, the sharp degradation of all evaluated VLMs is striking: full-sequence accuracy falls from 94.7\% to 20.3\% for GPT-5.4, from 96.0\% to 48.3\% for Gemini3-Flash, and from 91.3\% to 8.7\% for Kimi-K3 between \textit{Low} and \textit{High} density. This contrast highlights the lack of path-following robustness in current VLMs. Human-baseline details, full prefix-accuracy, and error-type classification are provided in Appendices~\ref{app:human_baseline} to \ref{app:swirl_error_analysis}.

As rotation density increases, a growing fraction of failures in most models, except Gemini3-Flash, take the form of an \textit{adjacent-turn jump} (\Cref{fig:swirl_path_preservation}). In this failure, instead of following the next dot along the current spiral turn, the model jumps outward to a dot at the same angular direction on the immediately adjacent outer turn and continues tracing from there. The systematic increase in adjacent-path jumps as nearby-path competition grows demonstrates its effectiveness as a controlled stressor for revealing the fragility of path preservation in current VLMs. We further verify that these results are not primarily driven by insufficient input resolution: no patch contains multiple colored dots, and doubling the input resolution yields no substantial recovery, as described in Appendix~\ref{app:swirl_resolution}.

\section{Diagnosing Path Preservation Under Nearby-Path Competition}
\label{sec:distractor_analysis}

We next examine how nearby, non-overlapping paths affect tracing and its internal representations, focusing on Qwen3-VL-8B~\citep{qwen3_vl}. We additionally conduct the same analysis with InternVL3.5-8B~\citep{internvl3.5}, with the additional results and full experimental details provided in Appendix~\ref{app:nearby_exp}.

\subsection{Effect of Nearby-Path Presence on Path Preservation}
\label{sec:nearby_line_probe}

\begin{wrapfigure}{r}{0.35\textwidth}
    \centering
    \vspace{-1.5em}
    \includegraphics[width=\linewidth]{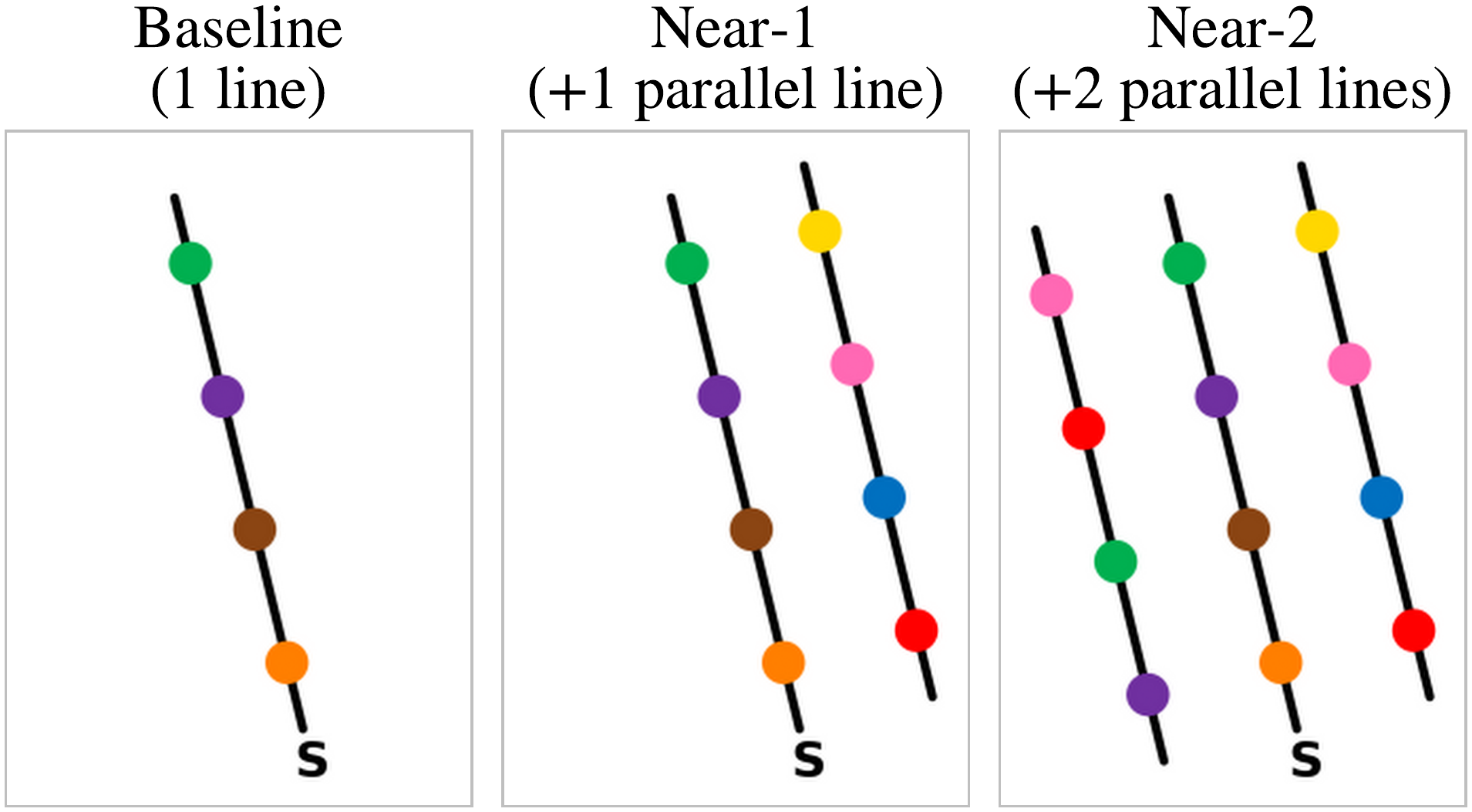}
    \caption{Examples from the Baseline, Near-1, and Near-2.}
    \label{fig:examples_3cond}
    \vspace{-1.5em}
\end{wrapfigure}


We treat the presence and number of nearby competing paths as the controlled variable, comparing \textit{Baseline}, \textit{Near-1}, and \textit{Near-2} (Figure~\ref{fig:examples_3cond}) while keeping the queried path fixed. Tracing accuracy decreases from 100\% in \textit{Baseline} to 98\% in \textit{Near-1} and 28\% in \textit{Near-2}, revealing that tracing is highly sensitive to nearby paths. To examine how tracing-related information evolves through the model, we apply simple linear probes at each evaluated vision-encoder block and LLM layer using representations of the current dot's image patch. \textit{Probe~A} predicts the color of the dot within the current patch, testing whether the model retains information about its current position on the queried path, while \textit{Probe~B} predicts the color of the next dot along the queried path, testing whether the representation contains information needed to follow the correct continuation. 

Probe~A remains at 100\% across all three conditions, whereas Probe~B accuracy at the strongest vision-encoder block decreases from 57.0\% to 48.4\% and 36.6\% as nearby-path competition increases. These results suggest that the mere presence of nearby paths can weaken continuation information while leaving local dot identity intact, with further degradation as more paths are introduced, already evident in the vision encoder. Detailed task construction, experimental settings, and probing analyses are provided in Appendix~\ref{app:probing_details}. We further confirm that replacing nearby paths with non-line controls has little effect on tracing performance, suggesting that nearby visual content alone does not account for the observed failures, as detailed in Appendix~\ref{app:distractor_type_controls}.

\subsection{Effect of Nearby-Path Similarity on Path Preservation}

\begin{figure*}[t!]
\centering
\includegraphics[width=\textwidth]{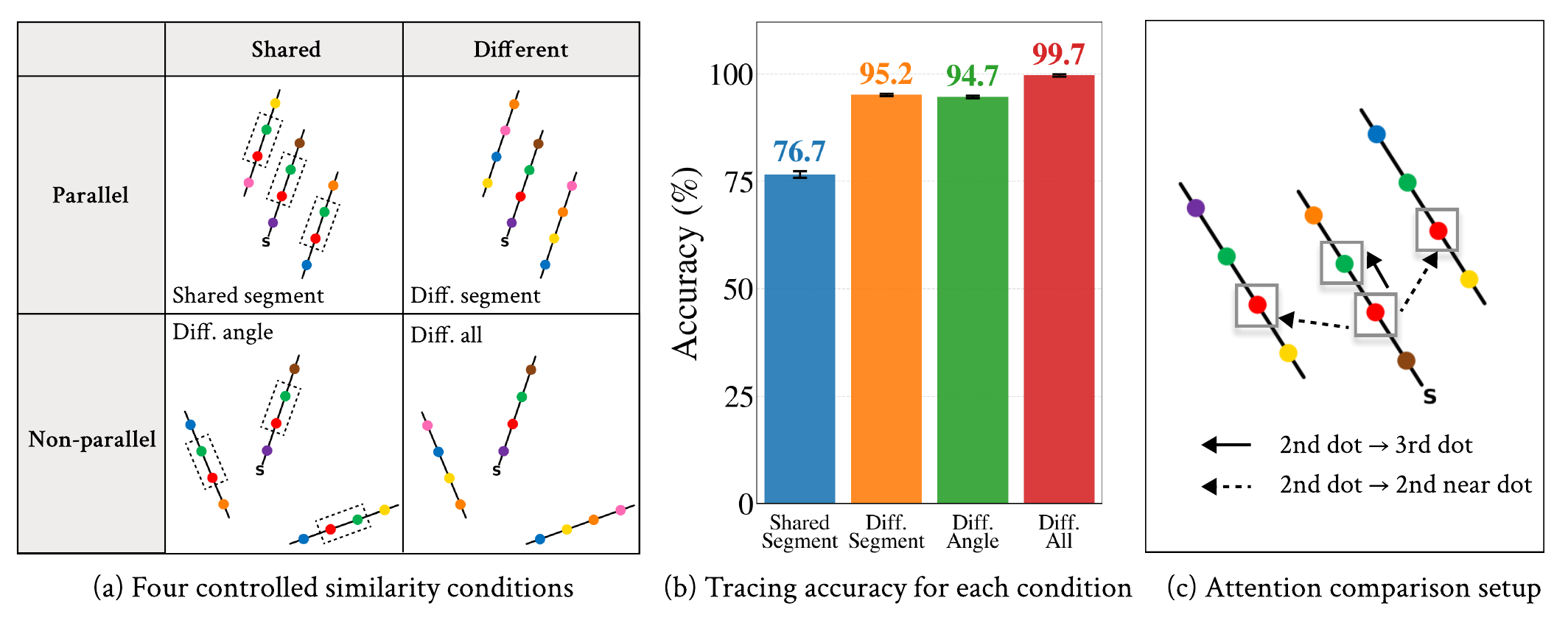}
\caption{
\textbf{Effect of nearby-path similarity on tracing performance and internal selectivity.}
(a) Four controlled similarity conditions varying whether nearby paths share local structure and orientation with the queried path.
(b) Tracing accuracy across conditions increases as local similarity between the queried and nearby paths decreases.
(c) Attention comparison setup, measuring attention from the second dot on the queried path to the correct third dot versus the nearby dot.
}
\label{fig:causal_conditions_acc_attndir}
\vspace{-0.5em}
\end{figure*}

Beyond the presence of nearby paths, we further examine how their visual and structural similarity to the queried path affects path preservation. We construct four conditions (\Cref{fig:causal_conditions_acc_attndir}a) in which each path contains four colored dots: the first verifies whether the model starts on the correct path, the second and third manipulate local continuation similarity, and the final dot has a path-specific color that reveals which path the model follows. In \textit{Shared segment}, nearby paths contain matching color steps at the second and third positions and run parallel to the queried path, creating the strongest local similarity. In \textit{Different segment}, matching color steps are removed while the paths remain locally parallel and retain directional similarity. In \textit{Different angle}, the matching color steps are retained but the nearby paths differ in orientation from the queried path. Finally, \textit{Different all} removes matching color steps and shared orientation, resulting in the weakest resemblance to the queried path.

As shown in \Cref{fig:causal_conditions_acc_attndir}b, accuracy drops to 76.7\% in the \textit{Shared segment} condition, compared to 99.7\% in \textit{Different all}. This suggests that the model's tracing becomes less stable not only in the presence of nearby paths but also as their structural similarity to the queried path increases.

\paragraph{Nearby-Path Similarity Weakens True-Continuation Preference.}
\label{sec:internal}

\begin{figure}[t!]
\centering

\includegraphics[width=0.8\textwidth]{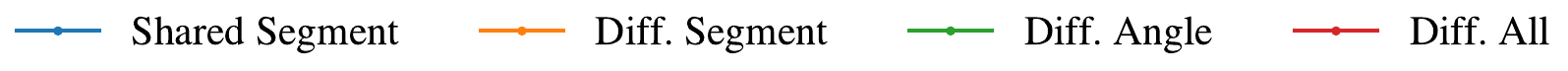}

\vspace{0.4em}

\includegraphics[width=\textwidth]{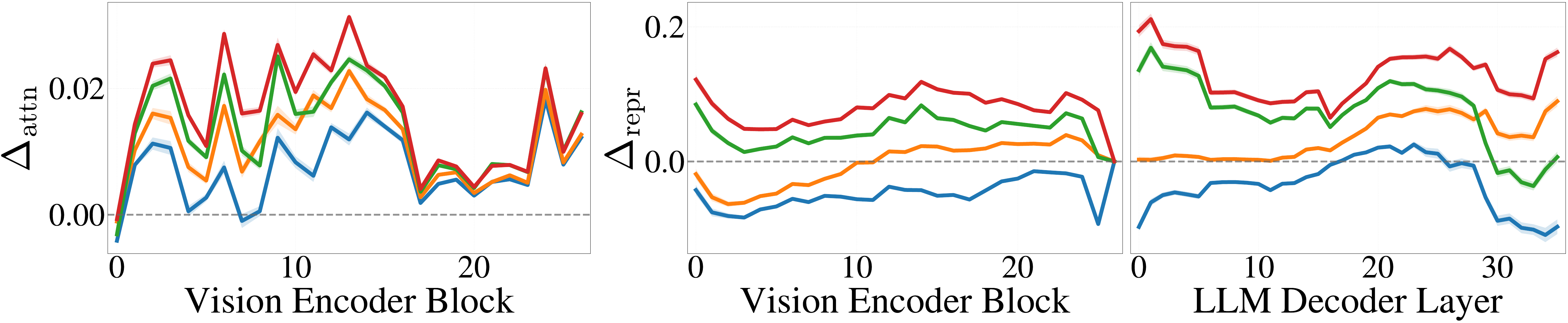}

\vskip -0.1em
\caption{
\textbf{Internal selectivity under increasing shared local structure.}
Left: block-wise vision attention margin ($\Delta_{\text{attn}}$) from \textit{Red} on the queried path to the correct next dot \textit{Green} versus the nearby competing distractor \textit{Dist-Red}.
Middle and right: cosine similarity margin ($\Delta_{\text{repr}}$) between the current queried-path state \textit{Red} and the true next state \textit{Green} versus \textit{Dist-Red}, shown separately for the vision encoder and LLM decoder, respectively.
}
\label{fig:internal_preference}
\vspace{-0.5em}
\end{figure}

To examine the internal signals underlying the behavioral accuracy changes, we analyze two complementary measures. \textit{Attention preference} in the vision encoder measures whether attention from the current queried-path position favors the true next dot over a competing dot on the nearby path, while \textit{representational preference} across the vision encoder and LLM decoder measures whether the hidden state at the current queried-path position is closer to the true continuation than to the nearby-path alternative. Together, these diagnostics characterize how increasing nearby-path similarity is reflected in the model's internal signals for path preservation at both the attention and representation levels. The compared regions are illustrated in \Cref{fig:causal_conditions_acc_attndir}c; implementation details are provided in Appendix~\ref{app:distractor_analysis}.

Let \(q_2\) and \(q_3\) denote the second and third dots on the queried path, and \(d_{1,2}\) and \(d_{2,2}\) the corresponding second dots on the two nearby paths.

For attention preference, we compute
\[
\Delta_{\mathrm{attn}}^{(\ell)}
=
a_{q_2\rightarrow q_3}^{(\ell)}
-
\frac{1}{2}
\left(
a_{q_2\rightarrow d_{1,2}}^{(\ell)}
+
a_{q_2\rightarrow d_{2,2}}^{(\ell)}
\right).
\]

For representational preference, we compute
\[
\Delta_{\mathrm{repr}}^{(\ell)}
=
\cos\!\big(\tilde{h}_{q_2}^{(\ell)},\, \tilde{h}_{q_3}^{(\ell)}\big)
-
\max\!\left\{
\cos\!\big(\tilde{h}_{q_2}^{(\ell)},\, \tilde{h}_{d_{1,2}}^{(\ell)}\big),
\cos\!\big(\tilde{h}_{q_2}^{(\ell)},\, \tilde{h}_{d_{2,2}}^{(\ell)}\big)
\right\},
\]
where \(\cos(\cdot,\cdot)\) denotes cosine similarity and \(\tilde{h}^{(\ell)}\) denotes the normalized hidden representation at layer \(\ell\). In both preferences, positive values indicate preference for the true continuation, values near zero indicate weak discrimination, and negative values indicate association with the nearby path.

If path following were robust, the model should maintain a consistent preference for the true continuation across all four conditions, regardless of nearby-path appearance. However, \Cref{fig:internal_preference} (left) shows that attention preference varies systematically across early-to-middle vision-encoder blocks despite the queried path remaining unchanged. The preference is lowest in the \textit{Shared segment} condition and highest in \textit{Different all}. A similar pattern appears in representational preference across both the vision encoder and LLM decoder (\Cref{fig:internal_preference}, middle and right), suggesting that nearby-path similarity can affect how tracing-relevant information is processed at both stages. These internal patterns align with the behavioral trend observed above (\Cref{fig:causal_conditions_acc_attndir}b), providing further experimental evidence consistent with limited robustness of path preservation in current VLMs.

\section{Standard Remedies Do Not Resolve Fragile Visual Tracing}
\label{sec:remedy}

We test three commonly used approaches for improving VLM performance: scaling up the model, increasing reasoning effort, and prompt engineering. We therefore ask whether these standard remedies can compensate for the lack of robustness in visual tracing under nearby visual competition.

\subsection{Vision-Encoder Scaling Does Not Improve Connectivity Preservation}

\begin{figure}[t!]
\centering

\includegraphics[width=\textwidth]{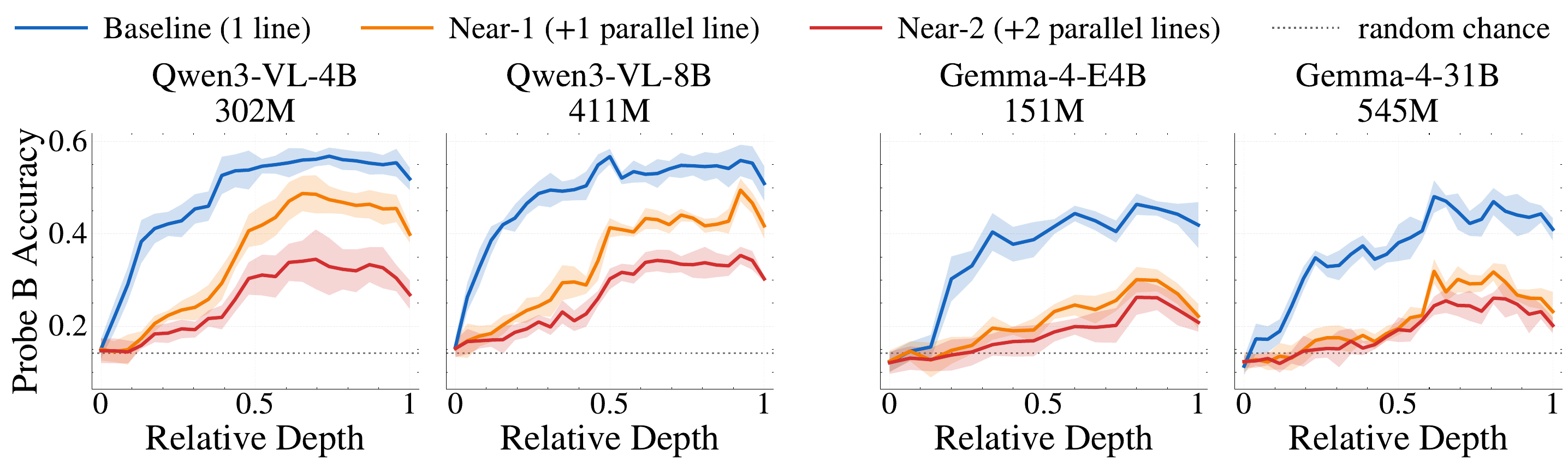}

\vskip -0.1em
\caption{
\textbf{Probe~B across vision-encoder scales.}
Probe~B accuracy across vision-encoder blocks for model sizes within the Gemma 4 and Qwen3-VL families under \textit{Baseline}, \textit{Near-1}, and \textit{Near-2}, which vary the number of nearby paths. Probe~B measures how well the representation of the queried-path dot encodes information about the next dot's color needed to continue tracing the path.
}
\label{fig:probeB_vis_scaling}
\vspace{-0.5em}
\end{figure}

Our preceding probing results suggest that tracing-relevant information is already weakened in the vision encoder, motivating a scaling analysis at the encoder level. We therefore apply Probe~B (in~\Cref{sec:nearby_line_probe}), which predicts the next dot's color from the image patch containing the current dot, across models with different vision-encoder scales from the Qwen3-VL (302M and 411M) and Gemma 4 (151M and 545M) families (\Cref{fig:probeB_vis_scaling}). Despite these increases in vision-encoder size, larger and smaller models show little difference in Probe~B accuracy, and the same degradation persists when nearby paths are introduced. Thus, although the vision encoder becomes larger, the improvement in connectivity representation remains limited. We further examine full-model scaling across the Qwen3-VL family, jointly scaling the vision encoder and LLM, and observe similarly limited gains across scales on the Swirl task; detailed results are provided in Appendix~\ref{app:accuracy_variability}.


\subsection{Reasoning Does Not Reliably Compensate for Tracing Failures}
\label{sec:reasoning}

\begin{figure*}[t]
\centering
\includegraphics[width=\textwidth]{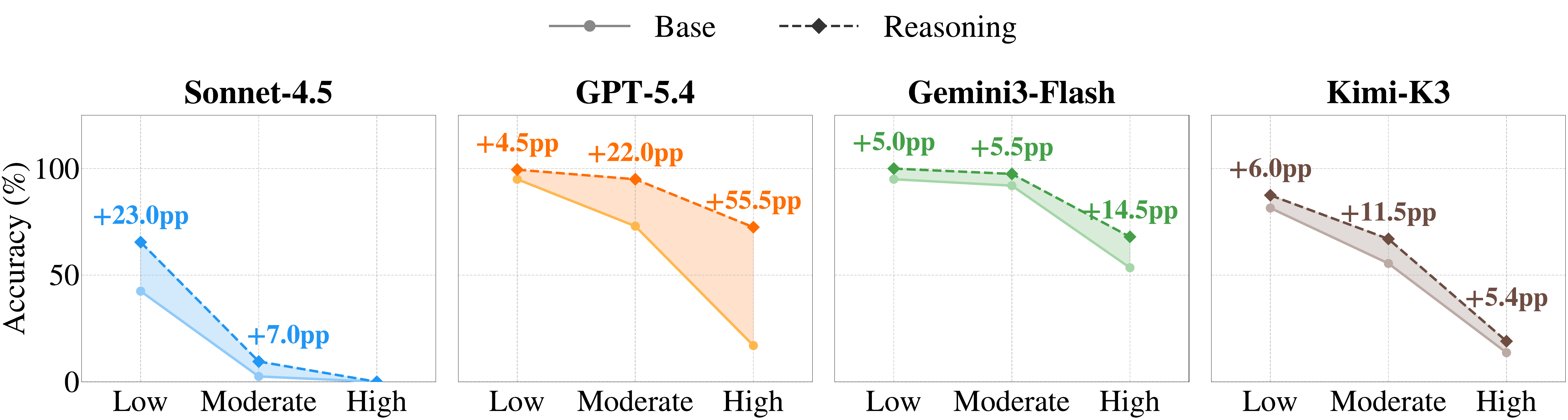}

\caption{
\textbf{Effect of reasoning on Swirl performance.}
Accuracy of \textit{Base} (non-reasoning) and \textit{Reasoning} models across Swirl difficulty levels, highlighting reasoning-induced performance changes.
}
\label{fig:reasoning_accuracy}
\end{figure*}


We investigate whether longer and deeper reasoning can support tracing and compensate for its lack of robustness. We evaluate reasoning-enabled VLMs on the \textit{Swirl} task (\Cref{fig:overview}, right). Results for additional models, including GLM-4.6V and Qwen3-VL-235B, are provided in Appendix~\ref{app:reasoning_trace_analysis}.


\paragraph{Reasoning Provides Uneven Benefits at Substantial Cost.}
As shown in \Cref{fig:reasoning_accuracy}, most models show only limited gains from reasoning, with GPT-5.4 as a notable exception, improving by 55.5 pp at \textit{High} density. However, these gains come at substantial computational cost. As rotation density rises, all models consume increasingly more tokens and exhibit more explicit self-corrections, such as \textit{``wait,''} or \textit{``let me check''} (\Cref{tab:reasoning_cost_selfcorrection}), despite the task requiring the same tracing procedure and output format. Even GPT-5.4, which achieves the largest accuracy gain, increases its average token usage from 501 at \textit{Low} density to 3,443 at \textit{High}, nearly a sevenfold increase. This sharp growth in reasoning length and repeated reconsideration warrants closer examination of how the reasoning process itself changes. Further details on self-correction frequency are provided in Appendix~\ref{app:self_correction_rate}.

\paragraph{Reasoning Strategies Shift with Density.}
The tracing task requires the same sequential procedure at every density level: starting from S, the model must identify the next connected dot and continue along the same path until reaching E. At \textit{Low} density, reasoning resembles this procedure through direct order inference or step-by-step spatial descriptions, whereas higher density increasingly induces qualitatively different, non-sequential strategies (\Cref{tab:reasoning_pattern_examples}). For example, GPT-5.4 directly lists the path order at \textit{Low} density but shifts at \textit{High} density toward geometric reconstruction, such as decomposing the spiral into multiple rings. Kimi-K3 uses spatial information at \textit{Low} density to describe how one point connects to the next, but at \textit{High} density instead attempts to reconstruct the spiral using inferred centers, radial coordinates, and radius-based ordering, relying on geometric criteria that do not necessarily preserve line connectivity. The sharp increase in token usage and self-correction is therefore largely driven by these substitute strategies, as additional reasoning is spent on geometric reconstruction rather than sequential tracing. Thus, changing only rotation density substantially alters how models approach the same tracing task, a shift that is difficult to reconcile with a stable and robust tracing mechanism. Additional reasoning examples are provided in Appendix~\ref{app:reasoning-trace-examples}.

\definecolor{costred}{HTML}{B2182B}

\begin{table*}[t!]
\centering
\caption{
\textbf{Reasoning length and self-correction frequency on Swirl.}
Average reasoning-token usage and self-correction markers per sample are reported across density levels for all evaluated reasoning models. Darker shading indicates a larger increase from each model's \textit{Low}-density baseline.
}
\label{tab:reasoning_cost_selfcorrection}
\small
\setlength{\tabcolsep}{5pt}
\renewcommand{\arraystretch}{1.15}

\begin{tabular}{lcccccc}
\toprule
& \multicolumn{3}{c}{\textbf{Reasoning tokens}}
& \multicolumn{3}{c}{\textbf{Self-corrections / sample}} \\
\cmidrule(lr){2-4}
\cmidrule(lr){5-7}
\textbf{Reasoning Model}
& \textbf{Low} & \textbf{Moderate} & \textbf{High}
& \textbf{Low} & \textbf{Moderate} & \textbf{High} \\
\midrule

Sonnet-4.5
& 390
& \cellcolor{costred!10}510
& \cellcolor{costred!10}537
& 4.20
& \cellcolor{costred!10}6.14
& \cellcolor{costred!15}6.38 \\

GPT-5.4
& 501
& \cellcolor{costred!25}1,373
& \cellcolor{costred!50}3,443
& 0.12
& \cellcolor{costred!35}0.53
& \cellcolor{costred!50}1.45 \\

Gemini3-Flash
& 557
& \cellcolor{costred!10}800
& \cellcolor{costred!25}1,337
& 0.47
& \cellcolor{costred!15}0.88
& \cellcolor{costred!35}1.66 \\

GLM-4.6V
& 477
& \cellcolor{costred!10}706
& \cellcolor{costred!15}854
& 9.25
& \cellcolor{costred!15}15.64
& \cellcolor{costred!25}19.49 \\

Kimi-K3
& 2,565
& \cellcolor{costred!5}2,984
& \cellcolor{costred!15}4,631
& 9.44
& \cellcolor{costred!5}10.07
& \cellcolor{costred!15}16.48 \\

Qwen3-VL-235B
& 657
& \cellcolor{costred!15}998
& \cellcolor{costred!15}1,311
& 10.23
& \cellcolor{costred!15}16.70
& \cellcolor{costred!25}24.23 \\

\bottomrule
\end{tabular}
\end{table*}

\definecolor{positiveblue}{HTML}{0072B2}
\definecolor{negativeorange}{HTML}{D55E00}

\begin{table*}[t]
\centering
\caption{
\textbf{Representative reasoning patterns across Swirl density levels.}
Reasoning traces from GPT-5.4 and Kimi-K3 illustrate a shift from brief sequential tracing at \textit{Low} density, where models follow the visible path from one continuation to the next (blue), toward geometric reconstruction at \textit{High} density, where models infer global structure using loops, radii, or spatial layout (orange).
}
\label{tab:reasoning_pattern_examples}
\small
\setlength{\tabcolsep}{5pt}
\renewcommand{\arraystretch}{1.15}
\begin{tabular}{
p{0.12\textwidth}
p{0.08\textwidth}
p{0.73\textwidth}}
\toprule
\textbf{Model} & \textbf{Density} & \textbf{Representative reasoning summary} \\
\midrule

\multirow{3}{*}{GPT-5.4}
& Low
& ``\textcolor{positiveblue}{\textbf{The line goes right/upward to }}purple
\textcolor{positiveblue}{\textbf{, curves to }}blue
\textcolor{positiveblue}{\textbf{, then left to }}orange
\textcolor{positiveblue}{\textbf{, moving down-left to }}brown
\textcolor{positiveblue}{\textbf{, down-right to }}yellow
\textcolor{positiveblue}{\textbf{, up-right to }}pink
\textcolor{positiveblue}{\textbf{, up-left to }}green
\textcolor{positiveblue}{\textbf{, left to }}red
\textcolor{positiveblue}{\textbf{, and finally down-left to }}E.'' \\

& \multirow{2}{*}{High}
& ``\textcolor{negativeorange}{\textbf{I see three loops}}: the
\textcolor{negativeorange}{\textbf{inner ring}} has red and pink, the
\textcolor{negativeorange}{\textbf{middle}} goes through yellow and blue, and ... ends at E. I need to figure out
\textcolor{negativeorange}{\textbf{how the spiral transitions from the first to the second loop}}...'' \\

\midrule

\multirow{2}{*}{Kimi-K3}
& Low
& ``Actually after blue
at bottom, \textcolor{positiveblue}{\textbf{ line continues to the right and up along outer right side to }}purple (453,355)
\textcolor{positiveblue}{\textbf{, then up-left to }}pink (429,211), \ldots\
\textcolor{positiveblue}{\textbf{ Let's determine exact order maybe }}S
\textcolor{positiveblue}{\textbf{ -> }}yellow
\textcolor{positiveblue}{\textbf{ -> }}orange
 ...
\textcolor{positiveblue}{\textbf{ -> }}red
\textcolor{positiveblue}{\textbf{ -> }}brown
\textcolor{positiveblue}{\textbf{ -> }}E.'' \\

& High
& ``Black spiral roughly centered at (330,320), radii: 
\textcolor{negativeorange}{\textbf{inner $\sim$42, middle $\sim$75, outer $\sim$110}}.
\textcolor{negativeorange}{\textbf{Let's reconstruct spiral mathematically}} from
endpoints and dots. \ldots path could go outward in spiral,
\textcolor{negativeorange}{\textbf{encountering dots by radius order:}} red (\textcolor{negativeorange}{\textbf{r27}}),
orange (\textcolor{negativeorange}{\textbf{r32}}), brown (\textcolor{negativeorange}{\textbf{r56}}),
...,
purple (\textcolor{negativeorange}{\textbf{r87}}), yellow (\textcolor{negativeorange}{\textbf{r96}}),
green (\textcolor{negativeorange}{\textbf{r100}}).'' \\

\bottomrule
\end{tabular}
\end{table*}

\begin{figure}[t]
    \centering
    \includegraphics[width=0.95\linewidth]{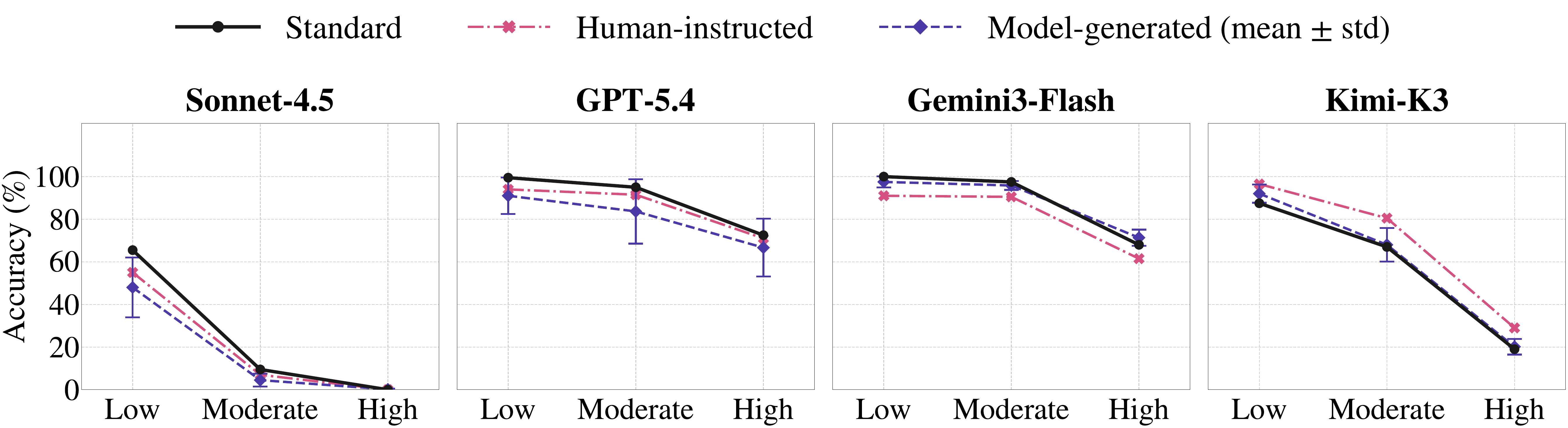}
    \caption{
\textbf{Effect of explicit tracing instructions on model accuracy.}
\textit{Standard} uses the original prompt without procedural guidance; \textit{Human-instructed} provides a human-designed explicit sequential tracing procedure; and \textit{Model-generated} uses model-specific instructions, reporting mean accuracy across three generated versions with standard-deviation error bars per model.
}
    \label{fig:prompting_intervention}
\end{figure}

\subsection{Explicit Tracing Instructions Do Not Improve Tracing Robustness}

We next test whether explicitly specifying how to trace the path can stabilize the reasoning shifts observed above and improve tracing robustness. We compare three prompting conditions. \textit{Standard} uses the original tracing prompt without procedural guidance. \textit{Human-instructed} provides a tracing procedure explicitly specified by a human. For \textit{Model-generated}, each model is shown a Swirl example and asked to describe a general tracing procedure that could be applied consistently across different Swirl instances and across rotation densities; this procedure is then provided back to the same model during evaluation. The full prompts and accuracy results are provided in Appendix~\ref{app:prompting_intervention_details}.

As shown in \Cref{fig:prompting_intervention}, neither human- nor model-generated instructions provide substantial performance gains or stabilize performance as density increases. Token usage continues to rise sharply with density, and reasoning still shifts from sequential tracing toward global reconstruction strategies (Appendix~\ref{app:prompting_reasoning_stability}). Thus, explicitly specifying the tracing procedure does not recover robust tracing in practice or stabilize the reasoning process. Notably, the model-generated instructions themselves largely recover the core human-designed tracing procedure: starting from S, following visible connectivity, and proceeding along the selected path. Models therefore appear to possess high-level procedural knowledge of tracing, yet fail to execute that procedure robustly as rotation density changes, revealing a gap between \textit{knowing what to do} and \textit{being able to execute it visually}.

\section{Case Study: Tracing Failures in Real-World Settings}
\vspace{-0.5em}

\begin{wrapfigure}{r}{0.60\textwidth}
    \centering
    \vspace{-1.2em}
    \includegraphics[width=\linewidth]{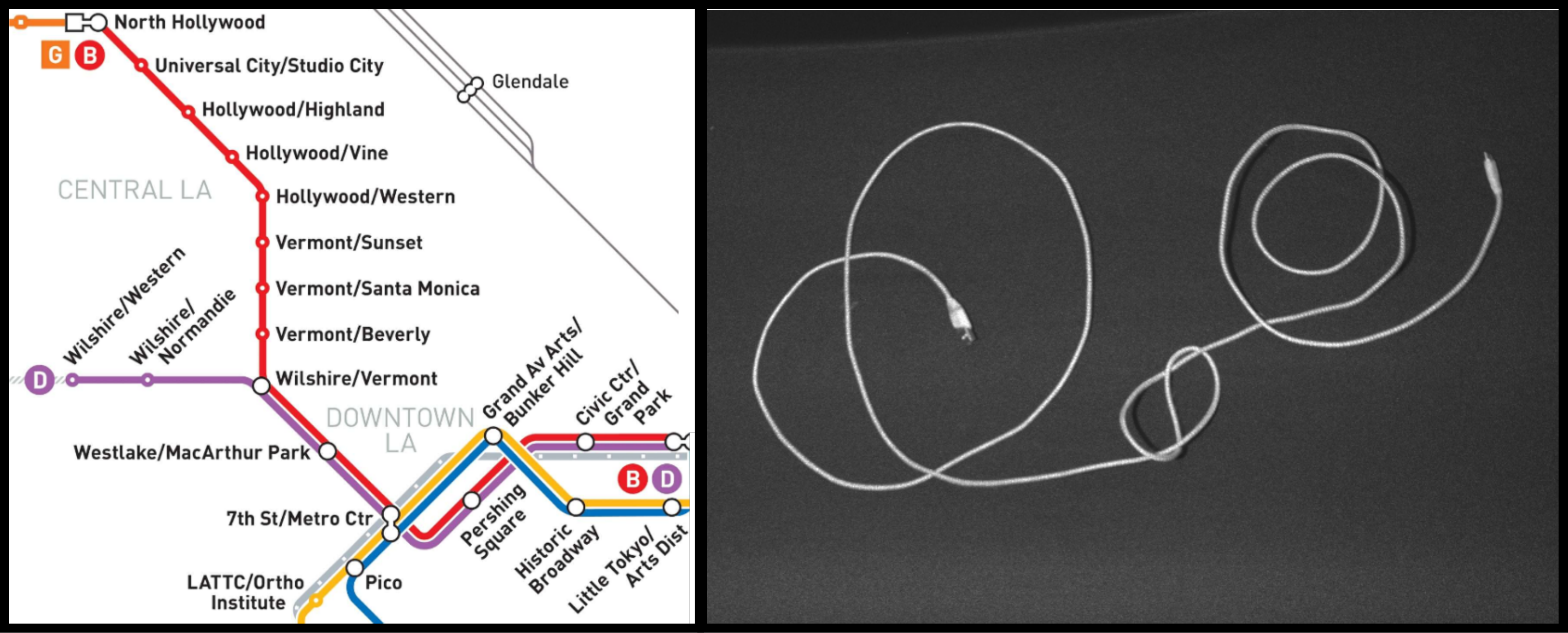}
    \vspace{-1.2em}
    \caption{
    \textbf{Realistic tracing examples.}
Metro maps (left) and tangled HANDLOOM cables (right), requiring tracing through crossings, overlaps, and nearby competing paths.}
    \label{fig:real_world_examples}
    \vspace{-1.5em}
\end{wrapfigure}

In previous sections, we stress-tested diverse VLMs and found that they fail to robustly preserve the selected path as tracing progresses. We examine whether the same path-preservation failure appears in more realistic tracing settings, where multiple structures cross, overlap, or run in close proximity. We evaluate two such settings: tangled cables from the HANDLOOM dataset~\citep{viswanath2023handloom}, and metro maps with realistic diagrams  (See~\Cref{fig:real_world_examples}).

As in our controlled experiments, performance remains unreliable in these realistic settings under richer visual conditions, with models exhibiting switches from the selected path to nearby visually competing paths. In the tangled-cable setting, we observe internal signal leakage toward an adjacent segment rather than remaining concentrated on the next dot along the selected path. In the metro map case, reasoning again falls back on substitute strategies, such as pruning candidate station names rather than tracing the line itself. The consistency of this behavior across both types of scenes indicates that the failures we identify are not artifacts of synthetic tasks, but reflect fragile path preservation in VLMs. Full experimental results for both datasets are provided in Appendix~\ref{app:real_world}.

\vspace{-0.5em}
\section{Conclusion}
\vspace{-0.5em}

We studied \textit{path following} to test whether current VLMs can robustly maintain a selected path during tracing. Across controlled settings, tracing performance degrades and path preservation becomes increasingly unstable in the presence of nearby paths and as their influence grows, while internal probing shows weakened information about the correct continuation despite preserved current-path information. Standard remedies, including vision-encoder scaling, reasoning, and explicit tracing instructions, do not reliably resolve this limitation. More broadly, as VLMs are increasingly deployed in settings that demand reliable path following, addressing this failure becomes a practical necessity. We hope this work provides a concrete target for future research on robust visual tracing.

\section*{Acknowledgment}

This work was supported in part by Institute of Information \& communications Technology Planning \& Evaluation (IITP) grant funded by the Korea government (MSIT) (No. RS-2024-00457882, AI Research Hub Project), IITP grant funded by the Korean Government (MSIT) (No. RS-2020-II201361, Artificial Intelligence Graduate School Program (Yonsei University)), and KCHIPS (Korea Collaborative \& High-tech Initiative for Prospective Semiconductor Research) (RS2024-00405946, 24052- 15TC) funded by the Ministry of Trade, Industry \& Energy (MOTIE, Korea).

\section*{Competing interests}
The authors declare that, during the 36 months preceding submission, they had no financial relationships outside the submitted work that could reasonably be perceived as influencing this research.

{
\bibliographystyle{plainnat}
\bibliography{neurips}
}

\newpage
\appendix



\section{Limitations}
\label{app:limitations}

Our study focuses on a specific visual primitive: preserving the identity of a selected path under nearby local competition. The controlled circuit and Swirl tasks are designed to isolate this operation, rather than to span the full diversity of visual reasoning. The color-dot sequence interface similarly operationalizes step-by-step tracing in a measurable form. While this abstracts away many properties of natural visual tasks, our HANDLOOM and metro-map evaluations provide complementary evidence that the same failure pattern appears in richer tracing settings.
Our proprietary-model results should be interpreted with the usual caveat that closed-source systems may change across API versions and inference settings. Hidden reasoning configurations may also affect both accuracy and generated reasoning traces. We therefore emphasize qualitative trends and recurring error modes in addition to exact performance numbers.
Finally, our failure-mode interpretations are based on interventions and systematic error patterns rather than full mechanistic identification. For open-source models, internal diagnostics provide supporting evidence that nearby distractors compete with the true continuation. For closed-source models, where internal states are unavailable, we treat reasoning traces and errors as behavioral evidence rather than direct evidence of internal mechanisms.

\section{Additional Implementation Details}
\label{app:implementation_details}

\subsection{Evaluated Models}
We evaluate a mixture of proprietary and open-source vision-language models.
Depending on the model and experiment, we access models through provider APIs or OpenRouter, or run publicly available Hugging Face checkpoints locally. Local checkpoints are used where access to internal model representations is required.
All models are used in accordance with the corresponding public release licenses, model cards, API terms, or service terms.

The evaluated models are as follows:
\begin{itemize}
    \item \textbf{GPT-5.4} \citep{openai_gpt5.4}
    \item \textbf{Claude Sonnet 4.5} \citep{anthropic_sonnet4.5}
    \item \textbf{Gemini3-Flash} \citep{google2025gemini3flash}
    \item \textbf{Kimi-K3} \citep{kimiteam2026kimik3openfrontier}
    \item \textbf{GLM-4.6V} \citep{glmvteam2025glm}
    \item \textbf{Qwen3-VL-4B-Instruct} \citep{qwen3_vl}
    \item \textbf{Qwen3-VL-8B-Instruct} \citep{qwen3_vl}
    \item \textbf{Qwen3-VL-8B-Thinking} \citep{qwen3_vl}
    \item \textbf{Qwen3-VL-32B-Instruct} \citep{qwen3_vl}
    \item \textbf{Qwen3-VL-235B-A22B-Instruct} \citep{qwen3_vl}
    \item \textbf{Qwen3-VL-235B-A22B-Thinking} \citep{qwen3_vl}
    \item \textbf{InternVL3.5-8B-Instruct} \citep{internvl3.5}
    \item \textbf{Gemma4-E4B} \citep{gemmateam2026gemma4}
    \item \textbf{Gemma4-31B} \citep{gemmateam2026gemma4}
\end{itemize}


\subsection{Inference Details}
Unless otherwise specified, we use the default generation settings for each model.
GPT-5.4 is accessed through the official OpenAI API, while other API-based evaluations use OpenRouter.
For models that support a \textit{medium} reasoning-effort setting, we use \textit{medium}; otherwise, we use the model's native reasoning configuration. We run each Circuit Connections and Swirl evaluation three times and report the mean accuracy and standard deviation across runs (Appendix~\ref{app:accuracy_variability}).

\paragraph{Compute Resources.}
Proprietary-model experiments were run through external model APIs. Open-source model evaluations and internal analyses were run on NVIDIA L40S GPUs. The internal attention and representation analyses used a single L40S GPU per run. We did not train or fine-tune any VLM; only the diagnostic linear probes were trained.


\section{Circuit Connections}
\label{app:circuit_task}
\subsection{Circuit Connections Task Construction}
\label{app:circuit_construction}

We adapt the \textit{Circuit Connections} task from prior work on smooth visual search~\citep{berman2025vlms}. The original task asks the model to identify the component connected to a queried breadboard port by following a wire through a circuit diagram. In that setting, wires are visually distinguished by their colors, ports are shown as circular terminals, and the answer only requires the final connected component label. The original generator also allows a broader range of visual configurations, including crossing wires.

\begin{figure}[H]
    \centering
    \includegraphics[width=\textwidth]{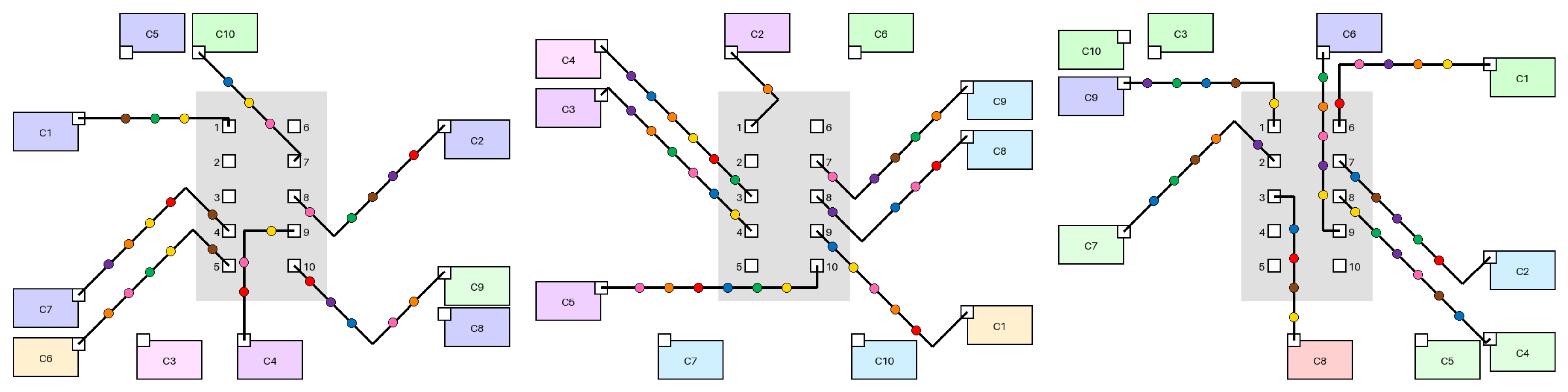}
    \vskip -0.5em
    \caption{
    \small
    \textbf{Examples from the modified Circuit Connections task.}
    Each example contains a central breadboard with numbered square ports, external component labels, single-color wires, and colored dots placed along each wire. The model is queried with a breadboard port and must report the colored dots on the corresponding wire in order, followed by the connected component label.
    }
    \label{fig:circuit_examples}
\end{figure}

Our version modifies this task to focus more directly on basic path following. First, we render all wires in the same color, so the model cannot solve the task by matching wire color across the image. Second, we remove wire crossings and overlaps, and reject candidate wires that collide with existing wires, pass through unrelated ports, overlap labels, cross component interiors, or run along the outline of the breadboard or components. This reduces explicit topological ambiguity and makes the task depend primarily on following the queried wire itself. Third, we replace circular ports with square white ports to make port locations visually explicit and consistent with the prompt. Finally, we place colored dots along each wire and ask the model to report the ordered dot sequence before giving the connected component label. This changes the task from endpoint identification to directly verifiable sequential tracing. Representative examples from our modified task are shown in \Cref{fig:circuit_examples}.

Each image contains a central breadboard, shown as a gray rectangle with numbered square ports, and several external components labeled with identifiers such as C1, C2, and so on. A queried wire is defined as a connected line that links one breadboard port to one component port. Each queried wire connects exactly two endpoints: the queried breadboard port and the component at the other end.

Colored dots are placed along each wire at regular intervals, with margins near endpoints and corners to avoid ambiguity from labels, ports, and sharp turns. The dot colors are sampled from a fixed color set: red, orange, yellow, green, blue, violet, pink, and brown. For each queried wire, the ground-truth answer is the ordered sequence of dot colors along that wire, followed by the connected component label and the token ``end.''

The final dataset contains 50 circuit images and 354 queried wires in total. Each wire connected to a breadboard port is queried once. A response is counted as correct only when it exactly recovers the full ordered color sequence on the queried wire, the correct component label, and the final ``end'' token. Responses with missing colors, extra colors, incorrect order, an incorrect component label, or missing termination are counted as incorrect.

\clearpage
\paragraph{Prompt.} The prompt is designed to force an explicit trace rather than a component-only answer.

\begin{tcolorbox}[
    colback=gray!5,
    colframe=black!60,
    boxrule=0.5pt,
    arc=2pt,
    left=6pt,
    right=6pt,
    top=6pt,
    bottom=6pt,
    title={Prompt for Circuit Connections}
]
\small
\textbf{Task:} A wire is a series of connected, same colored lines that go from the center of a port, represented on the screen as a white square, to another port. Each wire only connects two ports, one at either end. You must complete the sentence by listing all the colors of the dots on that wire in sequential order, and then give the label of the component connected at the other end. Append `end' at the very end. No explanations are allowed. 

\medskip

\textbf{Output format:} color1, color2, ..., colorN, \{component\_label\}, end

\medskip

Visually tracing the wire from port \{port\_num\} on the breadboard, which is the gray rectangle with numbered ports, the sequence of colored dots on its wire is strictly
\end{tcolorbox}

\subsection{Prefix Accuracy Across Circuit Wires}
\label{app:circuit_prefix}

\begin{figure}[htbp!]
\centering
\includegraphics[width=\linewidth]{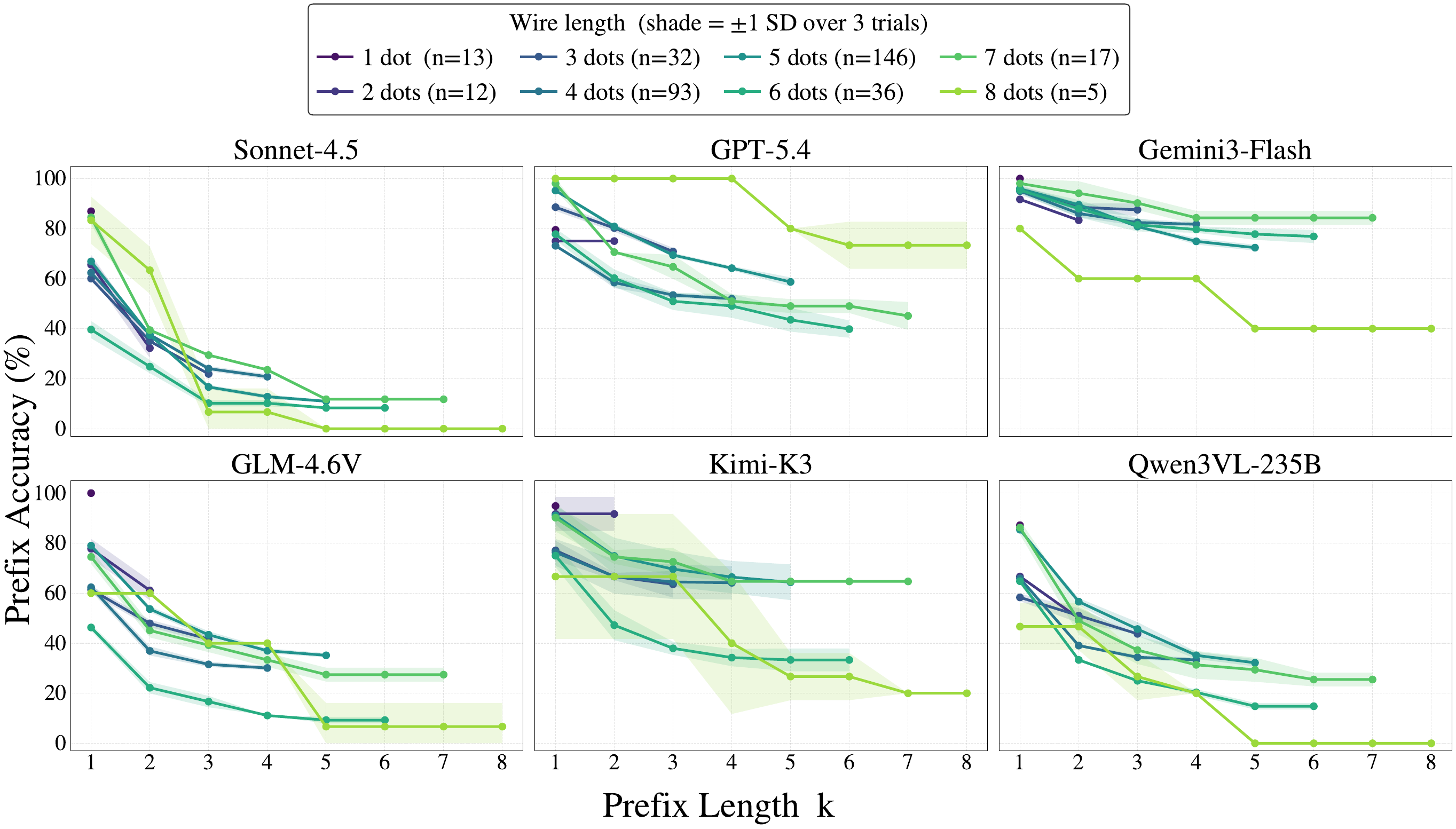}
\caption{\textbf{Cumulative prefix accuracy across wire lengths.}
Results are grouped by the number of colored dots on each wire. At each step, accuracy requires the model to have correctly recovered every preceding dot in the sequence. Across wire lengths, accuracy decreases as tracing proceeds, showing the accumulation of path-preservation errors.}
\label{fig:circuit_prefix_all}
\end{figure}

Because the \textit{Circuit Connections} dataset contains wires with different numbers of colored dots, we report cumulative prefix accuracy separately for each group. At each tracing step, a prediction is counted as correct only when all preceding dots have also been recovered correctly. As shown in \Cref{fig:circuit_prefix_all}, cumulative accuracy generally declines as tracing proceeds across the different wire groups. This pattern shows that failures accumulate as models attempt to extend the correct sequence to subsequent steps.

\subsection{Error-Type Classification for Circuit Connections}
\label{app:circuit_error_analysis}

To classify incorrect responses, we first compute the longest common prefix between the predicted and ground-truth color sequences. The first position at which the two sequences differ is defined as the divergence point. Errors are then assigned to mutually exclusive categories using a fixed priority order.

We first classify \textit{wrong-start} errors, in which the first predicted dot does not match the ground-truth first dot. Because these responses fail to initialize tracing on the queried wire, their subsequent trajectories cannot be meaningfully interpreted as path-following failures. We therefore identify wrong-start errors using only the first-dot prediction and apply the coordinate-based divergence analysis only to the remaining responses.

The next category is an \textit{adjacent-wire jump}. At the divergence point, we identify dots with the predicted color that belong to wires other than the queried wire. A response is classified as an adjacent-wire jump when the nearest candidate dot within a Euclidean distance of $\tau=135$ pixels from the last correctly traced position is selected, and the prediction subsequently follows that wire for at least two consecutive dots in either direction. The threshold corresponds to three grid intervals, where each interval is 45 pixels. We also assign this category when the minimum continuation length is not satisfied but the model predicts the endpoint component of the candidate wire, indicating that it followed the wire to its destination. 

The remaining errors are classified as \textit{truncation}, where the response terminates before reporting all dots in the ground-truth sequence; \textit{overshoot}, where additional dots are produced after the full correct sequence; or \textit{others}, which includes all other deviations that do not exhibit a consistent transition pattern.

\begin{table*}[htbp]
\centering
\caption{\textbf{Error-type distribution on Circuit Connections.}
Mean $\pm$ standard deviation of raw error counts over three trials is reported for each mutually exclusive error category. Each trial contains 354 evaluated wires.}
\label{tab:circuit_error_type_all}
\resizebox{\textwidth}{!}{
\begin{tabular}{lccccc}
\toprule
Model & Adjacent-wire jump & Wrong start & Truncation & Overshoot & Others \\
\midrule

Sonnet-4.5
& 129.67 $\pm$ 2.05
& 127.67 $\pm$ 1.25
& 12.33 $\pm$ 1.25
& 5.33 $\pm$ 0.47
& 34.67 $\pm$ 1.25 \\

GPT-5.4
& 91.67 $\pm$ 2.05
& 49.67 $\pm$ 1.89
& 5.33 $\pm$ 2.49
& 6.00 $\pm$ 0.82
& 18.00 $\pm$ 2.83 \\

Gemini3-Flash
& 34.67 $\pm$ 1.70
& 16.00 $\pm$ 3.27
& 11.67 $\pm$ 3.09
& 2.67 $\pm$ 0.47
& 17.33 $\pm$ 1.25 \\

Qwen3-VL-235B
& 116.33 $\pm$ 0.94
& 89.67 $\pm$ 0.94
& 8.33 $\pm$ 0.47
& 5.67 $\pm$ 0.47
& 33.33 $\pm$ 3.40 \\

Kimi-K3
& 56.00 $\pm$ 2.16
& 56.33 $\pm$ 11.26
& 7.00 $\pm$ 0.82
& 1.67 $\pm$ 0.47
& 19.00 $\pm$ 4.55 \\

GLM-4.6V
& 110.67 $\pm$ 2.05
& 106.33 $\pm$ 2.05
& 6.67 $\pm$ 0.94
& 11.00 $\pm$ 2.16
& 32.33 $\pm$ 3.40 \\

\bottomrule
\end{tabular}
}
\end{table*}

\section{Swirl}
\label{app:swirl_task}
\subsection{Swirl Task Construction}
\label{app:swirl_task_construction}

We design the \textit{Swirl} task as an abstract tracing benchmark with no semantic objects, background clutter, or path crossings. Each image contains a single spiral path marked with ``S'' at the start and ``E'' near the end. The model is instructed to trace the path from ``S'' and report the colored dots encountered along the path in order. All images in the main evaluation are rendered at a resolution of $640 \times 640$ pixels.

Each image contains exactly eight colored dots. To avoid ambiguity from repeated color identities, we use a fixed set of eight colors and assign each color only once within an image: red, orange, yellow, green, blue, purple, pink, and brown. Thus, errors reflect incorrect ordering or loss of the traced path rather than confusion caused by duplicate color labels.

\begin{tcolorbox}[
    colback=gray!5,
    colframe=black!60,
    boxrule=0.5pt,
    arc=2pt,
    left=6pt,
    right=6pt,
    top=6pt,
    bottom=6pt,
    title={Prompt for Swirl}
]
\small
\textbf{Task:} You must complete the sentence by listing the colors of the dots that satisfy the sentence's logic, in sequential order, and append `end' at the very end. No explanations are allowed.

\medskip

Visually tracing the path of the line from `S' in the image, the sequence of colored dots on its path is strictly
\end{tcolorbox}

We control density while keeping the path length, line width, number of dots, and image resolution fixed. Rotation density is defined by the spacing between neighboring turns, which is varied by changing the number of turns while preserving the overall path length. Specifically, the number of turns ranges from 1.4--1.8 for \textit{Low}, 1.8--2.2 for \textit{Moderate}, and 2.2--2.7 for \textit{High}. Increasing the number of turns compresses the spacing between neighboring tracks, allowing us to evaluate path preservation as nearby continuations become more closely spaced. For a spiral with $T$ turns, the dots are distributed across $8/T$ angular directions, separated by $360^\circ/(8/T)$, and the entire pattern is randomly rotated for each sample.

For the main evaluation in \Cref{sec:swirl}, we use 100 images per level. For the reasoning analysis in \Cref{sec:reasoning}, we use 200 images per level to examine a wider range of generated reasoning traces and failure patterns.

\subsection{Human Baseline}
\label{app:human_baseline}
We additionally measure a human baseline on the Swirl dataset. Ten human participants each completed 50 images randomly sampled from each rotation density set. Participants traced the spiral from the starting point and entered the eight encountered colors in order by selecting the corresponding color buttons. Each sample was limited to 20 seconds, while the average completion time was 11.3 seconds per sample in \textit{High} density level. Human participants maintained near-perfect prefix accuracy throughout the trace (\Cref{tab:swirl_prefix_all}), providing a reference for the difficulty of the task under the same sequential tracing requirement.

\subsection{Prefix Accuracy}
\label{app:swirl_prefix}
As shown in \Cref{tab:swirl_prefix_all}, the same cumulative degradation is observed across density levels, but becomes substantially stronger as density increases. At \textit{Low} density, several models maintain high prefix accuracy throughout the trace, whereas \textit{Moderate} and especially \textit{High} density produce much steeper declines. This further supports that path preservation becomes less robust as nearby competing turns are brought closer together.

\begin{table*}[!htbp]
\centering
\caption{
\textbf{Cumulative prefix accuracy on Swirl across density levels.}
Mean $\pm$ standard deviation (\%) is reported at each tracing step. Model results are computed over three trials, while the human baseline is computed over 10 participants. Prefix-$k$ accuracy requires all dots from the first through the $k$-th dot to be predicted correctly.
}
\label{tab:swirl_prefix_all}

\setlength{\tabcolsep}{6pt}
\renewcommand{\arraystretch}{0.95}

\begin{tabular}{llcccc}
\toprule
Density & Model & Prefix-1 & Prefix-2 & Prefix-3 & Prefix-4 \\
\midrule

\multirow{7}{*}{Low}
& Sonnet-4.5
& 78.67 $\pm$ 0.94
& 65.67 $\pm$ 1.70
& 62.33 $\pm$ 1.25
& 55.00 $\pm$ 1.63 \\

& GPT-5.4
& 100.00 $\pm$ 0.00
& 98.67 $\pm$ 1.25
& 97.67 $\pm$ 0.94
& 97.67 $\pm$ 0.94 \\

& Gemini3-Flash
& 96.00 $\pm$ 0.82
& 96.00 $\pm$ 0.82
& 96.00 $\pm$ 0.82
& 96.00 $\pm$ 0.82 \\

& Qwen3-VL-235B
& 95.67 $\pm$ 1.89
& 95.67 $\pm$ 1.89
& 95.67 $\pm$ 1.89
& 93.33 $\pm$ 0.94 \\

& Kimi-K3
& 97.33 $\pm$ 2.05
& 96.67 $\pm$ 2.36
& 96.67 $\pm$ 2.36
& 95.00 $\pm$ 2.45 \\

& GLM-4.6V
& 99.67 $\pm$ 0.47
& 96.00 $\pm$ 1.41
& 96.00 $\pm$ 1.41
& 96.00 $\pm$ 1.41 \\

& \textbf{Human}
& \textbf{100.00 $\pm$ 0.00}
& \textbf{100.00 $\pm$ 0.00}
& \textbf{100.00 $\pm$ 0.00}
& \textbf{100.00 $\pm$ 0.00} \\

\midrule

\multirow{7}{*}{Moderate}
& Sonnet-4.5
& 67.67 $\pm$ 1.89
& 52.00 $\pm$ 1.41
& 35.00 $\pm$ 2.45
& 18.67 $\pm$ 3.40 \\

& GPT-5.4
& 99.00 $\pm$ 0.00
& 88.67 $\pm$ 1.70
& 87.00 $\pm$ 1.63
& 85.67 $\pm$ 1.70 \\

& Gemini3-Flash
& 97.00 $\pm$ 1.63
& 97.00 $\pm$ 1.63
& 97.00 $\pm$ 1.63
& 97.00 $\pm$ 1.63 \\

& Qwen3-VL-235B
& 90.00 $\pm$ 0.00
& 84.67 $\pm$ 1.25
& 77.67 $\pm$ 0.94
& 67.33 $\pm$ 1.25 \\

& Kimi-K3
& 90.33 $\pm$ 1.70
& 88.67 $\pm$ 1.25
& 83.67 $\pm$ 0.94
& 77.00 $\pm$ 0.82 \\

& GLM-4.6V
& 93.00 $\pm$ 1.63
& 81.67 $\pm$ 1.25
& 80.00 $\pm$ 1.63
& 69.00 $\pm$ 0.82 \\

& \textbf{Human}
& \textbf{100.00 $\pm$ 0.00}
& \textbf{99.80 $\pm$ 0.60}
& \textbf{99.80 $\pm$ 0.60}
& \textbf{99.60 $\pm$ 0.80} \\

\midrule

\multirow{7}{*}{High}
& Sonnet-4.5
& 63.33 $\pm$ 1.25
& 41.33 $\pm$ 3.40
& 4.00 $\pm$ 0.82
& 1.00 $\pm$ 0.00 \\

& GPT-5.4
& 93.33 $\pm$ 0.47
& 75.00 $\pm$ 0.82
& 53.67 $\pm$ 3.30
& 48.00 $\pm$ 2.94 \\

& Gemini3-Flash
& 98.00 $\pm$ 0.82
& 97.33 $\pm$ 1.25
& 92.33 $\pm$ 1.25
& 90.67 $\pm$ 1.70 \\

& Qwen3-VL-235B
& 90.00 $\pm$ 2.83
& 82.33 $\pm$ 3.86
& 44.33 $\pm$ 3.86
& 20.33 $\pm$ 3.86 \\

& Kimi-K3
& 85.67 $\pm$ 1.25
& 73.33 $\pm$ 0.94
& 35.33 $\pm$ 2.49
& 26.67 $\pm$ 2.05 \\

& GLM-4.6V
& 81.33 $\pm$ 0.94
& 71.33 $\pm$ 1.70
& 29.33 $\pm$ 2.05
& 9.33 $\pm$ 1.25 \\

& \textbf{Human}
& \textbf{99.80 $\pm$ 0.60}
& \textbf{99.60 $\pm$ 0.80}
& \textbf{99.60 $\pm$ 0.80}
& \textbf{99.20 $\pm$ 1.33} \\

\bottomrule
\end{tabular}

\end{table*}

\begin{table*}[!htbp]
\ContinuedFloat
\centering
\caption{
\textbf{Cumulative prefix accuracy on Swirl across density levels (continued).}
}
\label{tab:swirl_prefix_all_cont}

\setlength{\tabcolsep}{6pt}
\renewcommand{\arraystretch}{0.95}

\begin{tabular}{llcccc}
\toprule
Density & Model & Prefix-5 & Prefix-6 & Prefix-7 & Prefix-8 \\
\midrule

\multirow{7}{*}{Low}
& Sonnet-4.5
& 44.00 $\pm$ 2.16
& 43.67 $\pm$ 1.70
& 43.67 $\pm$ 1.70
& 43.67 $\pm$ 1.70 \\

& GPT-5.4
& 94.67 $\pm$ 1.25
& 94.67 $\pm$ 1.25
& 94.67 $\pm$ 1.25
& 94.67 $\pm$ 1.25 \\

& Gemini3-Flash
& 96.00 $\pm$ 0.82
& 96.00 $\pm$ 0.82
& 96.00 $\pm$ 0.82
& 96.00 $\pm$ 0.82 \\

& Qwen3-VL-235B
& 81.33 $\pm$ 1.70
& 81.33 $\pm$ 1.70
& 81.33 $\pm$ 1.70
& 81.33 $\pm$ 1.70 \\

& Kimi-K3
& 91.67 $\pm$ 2.87
& 91.33 $\pm$ 2.87
& 91.33 $\pm$ 2.87
& 91.33 $\pm$ 2.87 \\

& GLM-4.6V
& 87.00 $\pm$ 1.41
& 87.00 $\pm$ 1.41
& 87.00 $\pm$ 1.41
& 87.00 $\pm$ 1.41 \\

& \textbf{Human}
& \textbf{100.00 $\pm$ 0.00}
& \textbf{100.00 $\pm$ 0.00}
& \textbf{100.00 $\pm$ 0.00}
& \textbf{100.00 $\pm$ 0.00} \\

\midrule

\multirow{7}{*}{Moderate}
& Sonnet-4.5
& 0.00 $\pm$ 0.00
& 0.00 $\pm$ 0.00
& 0.00 $\pm$ 0.00
& 0.00 $\pm$ 0.00 \\

& GPT-5.4
& 69.67 $\pm$ 3.40
& 69.33 $\pm$ 3.30
& 69.33 $\pm$ 3.30
& 69.33 $\pm$ 3.30 \\

& Gemini3-Flash
& 94.33 $\pm$ 1.25
& 92.67 $\pm$ 1.25
& 91.00 $\pm$ 0.82
& 91.00 $\pm$ 0.82 \\

& Qwen3-VL-235B
& 15.33 $\pm$ 1.70
& 14.33 $\pm$ 2.36
& 14.00 $\pm$ 2.16
& 14.00 $\pm$ 2.16 \\

& Kimi-K3
& 50.67 $\pm$ 4.64
& 50.33 $\pm$ 4.99
& 50.33 $\pm$ 4.99
& 50.33 $\pm$ 4.99 \\

& GLM-4.6V
& 17.00 $\pm$ 0.82
& 14.33 $\pm$ 0.47
& 13.33 $\pm$ 0.47
& 13.33 $\pm$ 0.47 \\

& \textbf{Human}
& \textbf{99.40 $\pm$ 0.92}
& \textbf{99.40 $\pm$ 0.90}
& \textbf{99.40 $\pm$ 0.90}
& \textbf{99.40 $\pm$ 0.90} \\

\midrule

\multirow{7}{*}{High}
& Sonnet-4.5
& 0.00 $\pm$ 0.00
& 0.00 $\pm$ 0.00
& 0.00 $\pm$ 0.00
& 0.00 $\pm$ 0.00 \\

& GPT-5.4
& 27.67 $\pm$ 2.05
& 22.00 $\pm$ 2.16
& 20.33 $\pm$ 2.05
& 20.33 $\pm$ 2.05 \\

& Gemini3-Flash
& 70.67 $\pm$ 4.64
& 61.67 $\pm$ 5.91
& 48.33 $\pm$ 3.30
& 48.33 $\pm$ 3.30 \\

& Qwen3-VL-235B
& 0.00 $\pm$ 0.00
& 0.00 $\pm$ 0.00
& 0.00 $\pm$ 0.00
& 0.00 $\pm$ 0.00 \\

& Kimi-K3
& 8.67 $\pm$ 1.25
& 8.67 $\pm$ 1.25
& 8.67 $\pm$ 1.25
& 8.67 $\pm$ 1.25 \\

& GLM-4.6V
& 1.00 $\pm$ 0.00
& 1.00 $\pm$ 0.00
& 1.00 $\pm$ 0.00
& 1.00 $\pm$ 0.00 \\

& \textbf{Human}
& \textbf{99.00 $\pm$ 1.34}
& \textbf{99.00 $\pm$ 1.34}
& \textbf{98.60 $\pm$ 1.28}
& \textbf{98.60 $\pm$ 1.28} \\

\bottomrule
\end{tabular}

\end{table*}

\clearpage

\subsection{Error-Type Classification for Swirl}
\label{app:swirl_error_analysis}

For incorrect predictions, we classify each sample by the first point at which the prediction diverges from the ground-truth sequence, so that every error is assigned to exactly one mutually exclusive category. Predictions are first normalized using the same procedure as for accuracy evaluation: text after the final ``end'' token is removed, color names are mapped to a canonical vocabulary (purple$\rightarrow$violet, magenta$\rightarrow$pink, cyan$\rightarrow$blue, teal$\rightarrow$green), only palette-color tokens are retained, and the trailing termination token is removed. A prediction is counted as correct only when the resulting color sequence exactly matches the ground-truth traversal.

We first separate \textit{wrong-start} errors, where the first predicted color already differs from the first ground-truth color. Because no correctly traced point precedes this divergence, an adjacent-turn transition is not defined for these cases. For the remaining errors, we identify the first divergence position $i \geq 1$ and classify the error according to the prediction at that point. 

An \textit{adjacent-turn jump} occurs when the substituted dot is the ring-adjacent neighbor of the last correctly traced dot, i.e., a point located one neighboring turn away along the same radial direction. Using the spiral geometry stored in the dataset metadata, this is operationally defined by a Euclidean distance of at most $\tau \cdot \texttt{ring\_spacing}$, with $\tau=1.05$. 

The remaining categories are \textit{Truncation}, where the prediction is a strict prefix of the ground-truth sequence without any substitution, and \textit{Overshoot}, where the complete ground-truth sequence is produced before additional colors are appended. After these categories are checked, all remaining failures are assigned to \textit{Others}. Since every image contains all eight palette colors exactly once, every predicted palette color necessarily corresponds to a dot in the image. Counts are aggregated over three independent runs of 100 images per density level and reported either as the average number of errors per 100 samples or as a proportion of all errors at that density level.

\begin{table*}[htbp]
\centering
\caption{\textbf{First-error type distribution on Swirl across density levels.}
Mean $\pm$ standard deviation denotes the number of samples per 100 evaluations assigned to each mutually exclusive first-error category over three trials. Each incorrect prediction is classified according to the first point of divergence from the ground-truth trace.}
\label{tab:swirl_error_type_all}
\resizebox{\textwidth}{!}{
\begin{tabular}{llccccc}
\toprule
Density & Model & Wrong start & Adjacent-turn jump & Truncation & Overshoot & Others \\
\midrule

\multirow{6}{*}{Low}
& Sonnet-4.5
& 21.3 $\pm$ 0.94
& 13.0 $\pm$ 0.82
& 0.0 $\pm$ 0.00
& 0.7 $\pm$ 0.94
& 22.0 $\pm$ 1.41 \\

& GPT-5.4
& 0.0 $\pm$ 0.00
& 2.7 $\pm$ 0.47
& 0.0 $\pm$ 0.00
& 0.0 $\pm$ 0.00
& 2.7 $\pm$ 1.25 \\

& Gemini3-Flash
& 4.0 $\pm$ 0.82
& 0.0 $\pm$ 0.00
& 0.0 $\pm$ 0.00
& 0.0 $\pm$ 0.00
& 0.0 $\pm$ 0.00 \\

& Qwen3-VL-235B
& 4.3 $\pm$ 1.89
& 8.3 $\pm$ 0.94
& 0.0 $\pm$ 0.00
& 0.0 $\pm$ 0.00
& 6.0 $\pm$ 0.82 \\

& Kimi-K3
& 2.3 $\pm$ 1.70
& 3.7 $\pm$ 1.89
& 0.3 $\pm$ 0.47
& 0.0 $\pm$ 0.00
& 2.3 $\pm$ 2.05 \\

& GLM-4.6V
& 0.3 $\pm$ 0.47
& 3.0 $\pm$ 0.00
& 0.0 $\pm$ 0.00
& 0.0 $\pm$ 0.00
& 9.7 $\pm$ 0.94 \\

\midrule

\multirow{6}{*}{Moderate}
& Sonnet-4.5
& 32.3 $\pm$ 1.70
& 48.0 $\pm$ 1.41
& 0.0 $\pm$ 0.00
& 0.0 $\pm$ 0.00
& 19.7 $\pm$ 2.49 \\

& GPT-5.4
& 1.0 $\pm$ 0.00
& 18.7 $\pm$ 3.40
& 0.0 $\pm$ 0.00
& 0.0 $\pm$ 0.00
& 11.0 $\pm$ 3.74 \\

& Gemini3-Flash
& 3.0 $\pm$ 1.63
& 0.7 $\pm$ 0.47
& 0.0 $\pm$ 0.00
& 0.0 $\pm$ 0.00
& 5.3 $\pm$ 1.89 \\

& Qwen3-VL-235B
& 10.0 $\pm$ 0.00
& 67.7 $\pm$ 3.09
& 0.0 $\pm$ 0.00
& 0.0 $\pm$ 0.00
& 8.3 $\pm$ 0.94 \\

& Kimi-K3
& 9.7 $\pm$ 1.70
& 36.0 $\pm$ 5.66
& 0.0 $\pm$ 0.00
& 0.0 $\pm$ 0.00
& 4.0 $\pm$ 1.41 \\

& GLM-4.6V
& 7.0 $\pm$ 1.63
& 54.7 $\pm$ 1.70
& 0.0 $\pm$ 0.00
& 0.0 $\pm$ 0.00
& 25.0 $\pm$ 1.63 \\

\midrule

\multirow{6}{*}{High}
& Sonnet-4.5
& 36.7 $\pm$ 1.25
& 61.7 $\pm$ 2.36
& 0.0 $\pm$ 0.00
& 0.0 $\pm$ 0.00
& 1.7 $\pm$ 1.25 \\

& GPT-5.4
& 6.7 $\pm$ 0.47
& 53.3 $\pm$ 2.05
& 0.0 $\pm$ 0.00
& 0.0 $\pm$ 0.00
& 19.7 $\pm$ 3.30 \\

& Gemini3-Flash
& 2.0 $\pm$ 0.82
& 3.3 $\pm$ 0.47
& 0.0 $\pm$ 0.00
& 0.0 $\pm$ 0.00
& 46.3 $\pm$ 3.86 \\

& Qwen3-VL-235B
& 10.0 $\pm$ 2.83
& 80.3 $\pm$ 3.86
& 0.7 $\pm$ 0.47
& 0.0 $\pm$ 0.00
& 9.0 $\pm$ 1.63 \\

& Kimi-K3
& 14.3 $\pm$ 1.25
& 60.0 $\pm$ 4.32
& 0.0 $\pm$ 0.00
& 0.0 $\pm$ 0.00
& 17.0 $\pm$ 2.16 \\

& GLM-4.6V
& 18.7 $\pm$ 0.94
& 70.0 $\pm$ 0.82
& 0.0 $\pm$ 0.00
& 0.0 $\pm$ 0.00
& 10.3 $\pm$ 0.47 \\

\bottomrule
\end{tabular}
}
\end{table*}

\subsection{Run-to-Run Consistency on Swirl}
\label{app:swirl_consistency}

To verify that the observed errors are not merely a product of stochastic sampling, we measured how consistently each model behaves across its three runs, aligning the runs image by image (Table~\ref{tab:swirl_consistency}). Mean pairwise Cohen's $\kappa$ on per-image correctness ranges from 0.64 to 0.96, indicating substantial run-to-run agreement. Given the large combinatorial space of possible eight-color output sequences, the repeated occurrence of identical errors is particularly meaningful: when two runs both fail on an image, 41.7--74.3\% of the failures produce the same predicted sequence. These results indicate that the observed failures are largely systematic rather than arising primarily from stochastic variation across runs.

\begin{table}[htbp]
\centering
\caption{
\textbf{Run-to-run consistency on Swirl.}
Statistics are pooled across the Low, Moderate, and High density levels over three runs (300 images per model). $\kappa$ denotes mean pairwise Cohen's $\kappa$ on per-image correctness, and \textit{Same wrong sequence} measures how often two jointly incorrect runs produce the same prediction.
}
\label{tab:swirl_consistency}
\begin{tabular}{lcc}
\toprule
Model & $\kappa$ & Same wrong sequence (\%) \\
\midrule
GPT-5.4        & 0.78 & 46.5 \\
Gemini3-Flash  & 0.65 & 56.6 \\
Sonnet-4.5     & 0.87 & 59.7 \\
Qwen3-VL-235B & 0.89 & 58.3 \\
GLM-4.6V       & 0.96 & 74.3 \\
Kimi-K3        & 0.64 & 41.7 \\
\bottomrule
\end{tabular}
\end{table}

\subsection{Input Resolution Control}
\label{app:swirl_resolution}

\begin{figure}[htbp]
\centering

\begin{minipage}[t]{0.34\textwidth}
    \vspace{0pt}
    \centering
    \includegraphics[width=\linewidth]{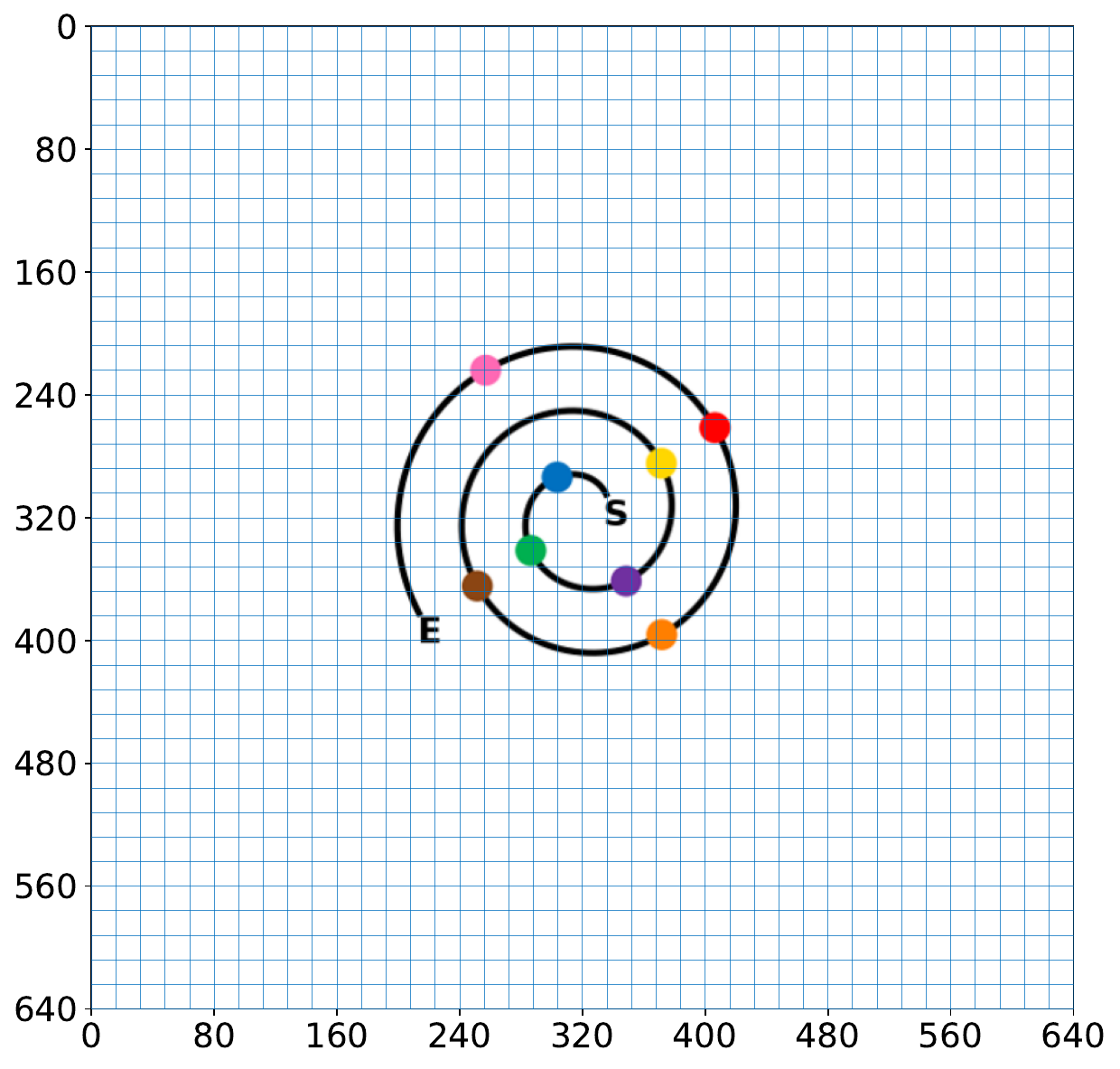}
\end{minipage}
\hfill
\begin{minipage}[t]{0.63\textwidth}
    \vspace{0pt}
    \centering
    \resizebox{\linewidth}{!}{
    \begin{tabular}{lccc}
    \toprule
    Model & \(1\times\) Res. & \(2\times\) Res. & \(\Delta\) \\
    \midrule
    Sonnet-4.5      & 0.0\%  & 0.0\%  & 0.0 pp \\
    GPT-5.4         & 20.3\% & 22.0\% & +1.7 pp \\
    Gemini3-Flash   & 48.3\% & 47.0\% & $-1.3$ pp \\
    Kimi-K3         & 8.7\%  & 16.0\% & +7.3 pp \\
    GLM-4.6V        & 1.0\%  & 4.0\%  & +3.0 pp \\
    Qwen3-VL-235B   & 0.0\%  & 0.0\%  & 0.0 pp \\
    \bottomrule
    \end{tabular}
    }
\end{minipage}

\caption{
\textbf{Input-resolution analysis on High-density Swirl.}
\textbf{Left:} Patch-grid visualization of a $640 \times 640$ Swirl image using the $40 \times 40$ patch grid of the Qwen3-VL vision tokenizer. Each colored dot and its surrounding line span approximately a $3 \times 3$-patch region, and no patch contains more than one dot.
\textbf{Right:} Tracing accuracy when increasing the input resolution from $640 \times 640$ (\(1\times\)) to $1280 \times 1280$ (\(2\times\)).
}
\label{fig:swirl_resolution}
\end{figure}

We examine whether insufficient input resolution could explain the tracing failures. As shown in \Cref{fig:swirl_resolution} (left) for the \textit{High}-density setting, the original $640 \times 640$ images are divided by the Qwen3-VL vision tokenizer into a $40 \times 40$ grid of $16 \times 16$-pixel patches, yielding 1,600 visual patches. No patch contains more than one dot, ruling out unresolved dot merging at the patch level.

Because the patch sizes and visual tokenization schemes of closed-source frontier models are not publicly available, we additionally conduct a model-agnostic resolution control on the \textit{High}-density setting by doubling both spatial dimensions of the input. Specifically, we repeat the evaluation using $1280 \times 1280$ images while preserving the same underlying path, dot placement, and task instructions.

As shown in \Cref{fig:swirl_resolution} (right), doubling the input resolution produces only limited and inconsistent changes across models. Sonnet-4.5 and Qwen3-VL-235B remain unchanged at 0.0\%, while GPT-5.4 increases from 20.3\% to 22.0\% (+1.7 pp) and GLM-4.6V from 1.0\% to 4.0\% (+3.0 pp). Gemini3-Flash slightly decreases from 48.3\% to 47.0\% ($-1.3$ pp). Kimi-K3 shows the largest improvement, from 8.7\% to 16.0\% (+7.3 pp), but its absolute accuracy remains low even at the doubled resolution. Overall, these results suggest that increasing input resolution provides at most limited recovery and that insufficient input resolution is unlikely to be the primary cause of the observed tracing failures.

\clearpage
\subsection{Full-Model Scaling on Swirl}
\label{app:accuracy_variability}

\begin{wrapfigure}{r}{0.45\textwidth}
    \centering
    \vspace{-1.0em}
    \includegraphics[width=0.45\textwidth]{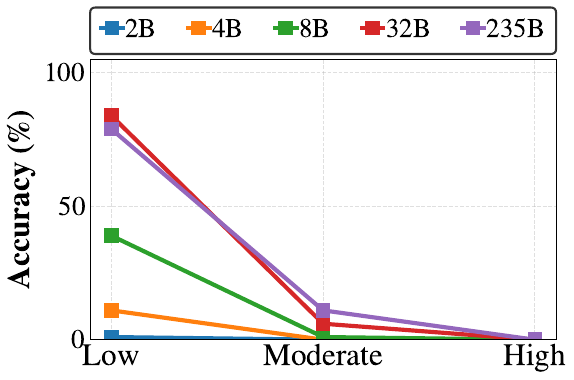}
    \vspace{-0.7em}
    \caption{
    \small
    \textbf{Qwen3-VL model-size scaling on Swirl task.}
    Accuracy across rotation-density levels for models from 2B to 235B parameters.
    }
    \label{fig:qwen_scaling_tracing}
    \vspace{-0.5em}
\end{wrapfigure}

We further examine whether increasing overall model size improves visual tracing across the Qwen3-VL family, comparing the 2B, 4B, 8B, 32B, and 235B variants. The vision encoder scales from 24 layers with a 1024-dimensional hidden size in the 2B and 4B models to 27 layers with a 1152-dimensional hidden size in the 8B, 32B, and 235B models. The language-model component also scales substantially: the 2B, 4B, 8B, and 32B variants use 28/2048, 36/2560, 36/4096, and 64/5120 layers/hidden dimensions, respectively, while the 235B-A22B variant uses a 94-layer mixture-of-experts decoder with 128 experts and 8 activated per token. As shown in \Cref{fig:qwen_scaling_tracing}, larger models provide only modest gains, and these advantages diminish as rotation density increases. In particular, performance remains substantially degraded under the more difficult density conditions even at the largest model scale. This indicates that increasing overall model capacity alone does not resolve the underlying path-preservation failure.


\section{Path Preservation Under Local Competition Details}
\label{app:nearby_exp}

\subsection{Task Construction}
We construct three conditions under otherwise identical visual settings, keeping the queried path fixed for each sample while varying only the number of nearby competing paths (\Cref{fig:examples_3cond}). The \textit{Baseline} condition contains a single straight line with four colored dots and an ``S'' marking the start point. In \textit{Near-1}, one parallel nearby path is added on one side of the queried path. \textit{Near-2} adds a second parallel path on the opposite side at the same distance from the queried path. Across conditions, dot colors are sampled from a fixed eight-color palette, with four non-repeating colors assigned to each path. When colors are reused across paths, they are arranged such that no two paths share the same color at the same tracing step. The first and fourth dot colors are also distinct across paths, allowing the corresponding path to be identified at both the beginning and end of the trace.

\subsection{Prefix Accuracy Under nearby-path Perturbations}
\label{app:nearby_prefix_accuracy}

Across both models, cumulative prefix accuracy declines as nearby paths are introduced (\Cref{tab:nearby_prefix_accuracy}). The \textit{Baseline} remains near-perfect throughout all four tracing steps, while \textit{Near-1} shows only a modest decline. In contrast, \textit{Near-2} exhibits a sharp cumulative drop: Qwen3-VL-8B decreases from 64.3\% at the first dot to 27.7\% at the fourth, while InternVL3.5-8B falls from 94.3\% to 33.0\%. This consistent pattern across two model families shows that path following becomes substantially less reliable when the queried path is surrounded by nearby paths.

\begin{table}[htbp]
\centering
\caption{\textbf{Cumulative prefix accuracy under controlled nearby-path perturbations.}
Mean $\pm$ standard deviation (\%) over three runs is reported for Qwen3-VL-8B and InternVL3.5-8B.}
\label{tab:nearby_prefix_accuracy}
\begin{tabular}{llcccc}
\toprule
Model & Condition & Prefix-1 & Prefix-2 & Prefix-3 & Prefix-4 \\
\midrule

\multirow{3}{*}{Qwen3-VL-8B}
& Baseline
& 100.00 $\pm$ 0.00
& 100.00 $\pm$ 0.00
& 100.00 $\pm$ 0.00
& 100.00 $\pm$ 0.00 \\

& Near-1
& 98.00 $\pm$ 0.00
& 97.67 $\pm$ 0.47
& 97.67 $\pm$ 0.47
& 97.67 $\pm$ 0.47 \\

& Near-2
& 64.33 $\pm$ 0.47
& 47.67 $\pm$ 0.47
& 36.67 $\pm$ 0.47
& 27.67 $\pm$ 0.47 \\

\midrule

\multirow{3}{*}{InternVL3.5-8B}
& Baseline
& 99.00 $\pm$ 1.00
& 99.00 $\pm$ 1.00
& 99.00 $\pm$ 1.00
& 99.00 $\pm$ 1.00 \\

& Near-1
& 99.33 $\pm$ 0.58
& 96.67 $\pm$ 0.58
& 91.67 $\pm$ 1.53
& 86.67 $\pm$ 2.31 \\

& Near-2
& 94.33 $\pm$ 0.58
& 77.67 $\pm$ 6.03
& 48.00 $\pm$ 3.61
& 33.00 $\pm$ 2.65 \\

\bottomrule
\end{tabular}
\end{table}

\subsection{Linear Probing Details}
\label{app:probing_details}

To examine the corresponding internal representations, we extract queried-path dot representations from both the vision encoder and LLM image tokens and train single-layer linear probes for two targets. \textit{Probe~A} predicts the color of the probed dot itself, measuring preservation of local visual identity. \textit{Probe~B} predicts the color of the next dot along the queried path, measuring whether the representation contains information required for path continuation. The same probing procedure is applied across all three conditions.

\paragraph{Probe training.}
For each condition and layer independently, we train a multinomial linear probe using
\[
\texttt{StandardScaler}\rightarrow\texttt{PCA}\rightarrow\texttt{LogisticRegression},
\]
with $C=1.0$, L2 regularization, the L-BFGS optimizer, and a maximum of 2,000 iterations. Each queried-path dot contributes one training example: the averaged $3\times3$ patch embedding for the vision encoder or the corresponding merged image token for the LLM. The scaler and PCA are fitted only on the training fold.

\paragraph{Probe Input Construction and Cross-Validation.}

\begin{wrapfigure}{r}{0.20\textwidth}
\centering
\vspace{-10pt}
\includegraphics[width=\linewidth]{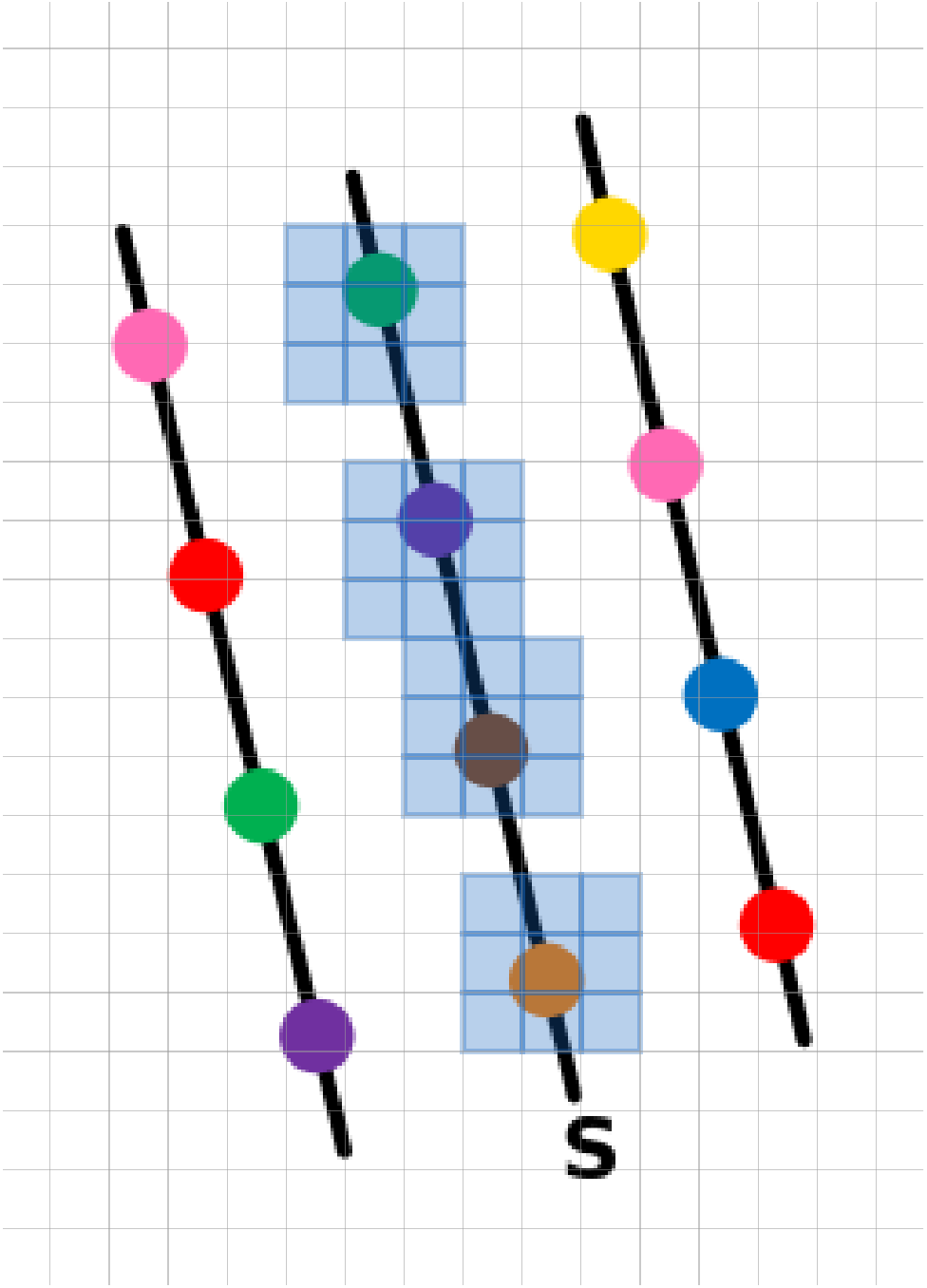}
\caption{\textbf{Patch assignment for linear probing.}
Example of the $3\times3$ patch region used to extract each queried-path dot representation.}
\label{fig:probe_patches}
\vspace{-10pt}
\end{wrapfigure}

We conduct the probing analysis using Qwen3-VL-8B~\citep{qwen3_vl} and InternVL3.5-8B~\citep{internvl3.5} with 400 images per condition, probing only representations associated with dots on the queried path. For the vision encoder, we extract a $3\times3$ pre-merge patch region centered on each queried-path dot (\Cref{fig:probe_patches}) at blocks $\{0,4,8,12,16,20,24,26\}$ for Qwen3-VL-8B (27 blocks) and $\{0,4,8,12,16,20,23\}$ for InternVL3.5-8B (24 blocks), sampling every fourth block and including the final one. Each region covers the full colored dot without overlapping regions assigned to other dots. For the LLM, we probe layers $\{0,6,12,18,24,30,35\}$; both models use a 36-layer Qwen3 language model.

For a queried-path dot $q_j$ at vision block $\ell$, the corresponding patch features are
\[
H_{i,j}^{(\ell)} \in \mathbb{R}^{9 \times d_v},
\]
where $d_v$ is the vision-encoder hidden dimension. We average the nine patch vectors to obtain a single dot representation,
\[
x_{i,j}^{(\ell)}
=
\frac{1}{9}\sum_{p=1}^{9} H_{i,j,p}^{(\ell)}
\in \mathbb{R}^{d_v}.
\]
For the LLM, each dot is represented by its corresponding merged image token,
\[
z_{i,j}^{(k)} \in \mathbb{R}^{d_l},
\]
where $d_l$ is the LLM hidden dimension at layer $k$.

Thus, for each layer and condition, the full probe inputs are
\[
X_A \in \mathbb{R}^{1600 \times d},
\qquad
X_B \in \mathbb{R}^{1200 \times d},
\]
where $d=d_v$ for vision representations and $d=d_l$ for LLM representations. \textit{Probe~A} uses all four queried-path dots ($400\times4=1{,}600$ examples), whereas \textit{Probe~B} uses the first three dots because the fourth has no successor ($400\times3=1{,}200$ examples).

We evaluate the probes using five-fold grouped cross-validation with images as groups. The 400 images in each condition are divided into five disjoint folds of 80 images. In each split, one fold is used for testing and the remaining four for training, ensuring that representations from the same image never appear in both sets. This yields 1,280/320 train/test examples for \textit{Probe~A} and 960/240 for \textit{Probe~B} in each split. The number of examples is matched across conditions, and probe accuracy is averaged over the five test folds.


\paragraph{PCA dimension sensitivity.}
PCA is applied to reduce the high-dimensional vision and LLM representations before fitting the linear probe, limiting overfitting while retaining information relevant for decoding. We repeat the analysis across PCA dimensions to verify that the observed Probe~A and Probe~B results are not specific to the dimensionality choice. The main experiments use $n_{\mathrm{PCA}}=512$.

For \textit{Probe~A}, \Cref{tab:probe_pca_sweep} (A) reports results at vision block 24 and LLM layer 6, the respective peak-performing locations. Accuracy saturates early and remains similarly high across \textit{Baseline}, \textit{Near-1}, and \textit{Near-2}, indicating that local color identity is preserved regardless of nearby paths. We nevertheless use $n_{\mathrm{PCA}}=512$ to keep the setup consistent with Probe~B.

For \textit{Probe~B}, \Cref{tab:probe_pca_sweep} (B) reports vision block 24, where vision accuracy peaks, and LLM layer 0, the earliest probed LLM representation. Vision accuracy saturates at 512 dimensions, while LLM accuracy increases only slightly beyond 512. Across informative PCA dimensions, the ordering \textit{Baseline} $>$ \textit{Near-1} $>$ \textit{Near-2} remains stable, showing that the observed degradation is not driven by the dimensionality choice.

\begin{table*}[htbp]
\centering
\caption{
\textbf{Probe accuracy across PCA dimensions.}
Held-out linear-probe accuracy across PCA dimensions for \textbf{(A)} Probe~A and \textbf{(B)} Probe~B.
The main setting, $n_{\mathrm{PCA}}=512$, is highlighted in bold.
Explained variance in Panel B refers to the vision features.
}
\label{tab:probe_pca_sweep}
\small
\setlength{\tabcolsep}{5pt}
\renewcommand{\arraystretch}{1.12}

\textbf{(A) Probe~A: Current-dot color prediction}

\vspace{0.4em}

\resizebox{0.78\textwidth}{!}{
\begin{tabular}{ccccccc}
\toprule
& \multicolumn{3}{c}{\textbf{Vision block 24}}
& \multicolumn{3}{c}{\textbf{LLM layer 6}} \\
\cmidrule(lr){2-4}
\cmidrule(lr){5-7}
$n_{\mathrm{PCA}}$
& Baseline & Near-1 & Near-2
& Baseline & Near-1 & Near-2 \\
\midrule
8   & 0.930 & 0.988 & 0.998 & 0.151 & 0.741 & 0.913 \\
16  & 1.000 & 1.000 & 1.000 & 0.922 & 0.994 & 0.989 \\
32  & 1.000 & 1.000 & 1.000 & 0.983 & 0.992 & 0.989 \\
64  & 1.000 & 1.000 & 1.000 & 0.984 & 0.989 & 0.989 \\
128 & 1.000 & 1.000 & 1.000 & 0.986 & 0.989 & 0.989 \\
256 & 1.000 & 1.000 & 1.000 & 0.990 & 0.991 & 0.990 \\
\textbf{512}
& \textbf{1.000} & \textbf{1.000} & \textbf{1.000}
& \textbf{0.991} & \textbf{0.993} & \textbf{0.991} \\
768 & 1.000 & 1.000 & 1.000 & 0.991 & 0.993 & 0.991 \\
Raw (960) & 1.000 & 1.000 & 1.000 & 0.991 & 0.993 & 0.991 \\
\bottomrule
\end{tabular}
}

\vspace{1.2em}

\textbf{(B) Probe~B: Next-dot color prediction}

\vspace{0.4em}

\resizebox{\textwidth}{!}{
\begin{tabular}{cccccccc}
\toprule
& & \multicolumn{3}{c}{\textbf{Vision block 24}}
& \multicolumn{3}{c}{\textbf{LLM layer 0}} \\
\cmidrule(lr){3-5}
\cmidrule(lr){6-8}
$n_{\mathrm{PCA}}$
& Explained var.
& Baseline & Near-1 & Near-2
& Baseline & Near-1 & Near-2 \\
\midrule
8   & 65.4\%  & 0.163 & 0.142 & 0.156 & 0.156 & 0.156 & 0.163 \\
16  & 83.2\%  & 0.302 & 0.205 & 0.163 & 0.214 & 0.173 & 0.166 \\
32  & 91.1\%  & 0.347 & 0.220 & 0.183 & 0.336 & 0.234 & 0.175 \\
64  & 95.5\%  & 0.452 & 0.368 & 0.260 & 0.356 & 0.245 & 0.193 \\
128 & 98.0\%  & 0.485 & 0.386 & 0.318 & 0.370 & 0.293 & 0.218 \\
256 & 99.3\%  & 0.543 & 0.460 & 0.355 & 0.395 & 0.329 & 0.240 \\
\textbf{512} & \textbf{99.9\%}
& \textbf{0.570} & \textbf{0.484} & \textbf{0.366}
& \textbf{0.411} & \textbf{0.340} & \textbf{0.253} \\
768 & 100.0\% & 0.569 & 0.493 & 0.374 & 0.423 & 0.345 & 0.260 \\
Raw (960) & 100.0\% & 0.570 & 0.493 & 0.373 & 0.427 & 0.343 & 0.267 \\
\bottomrule
\end{tabular}
}

\end{table*}

\begin{figure}[htbp!]
\centering
\includegraphics[width=0.78\textwidth]{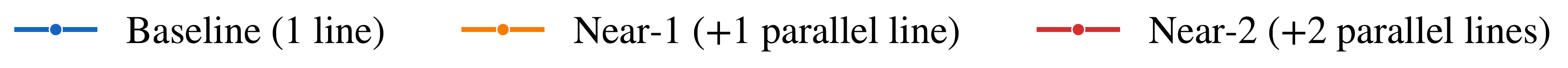}
\vspace{0.4em}
\includegraphics[width=\textwidth]{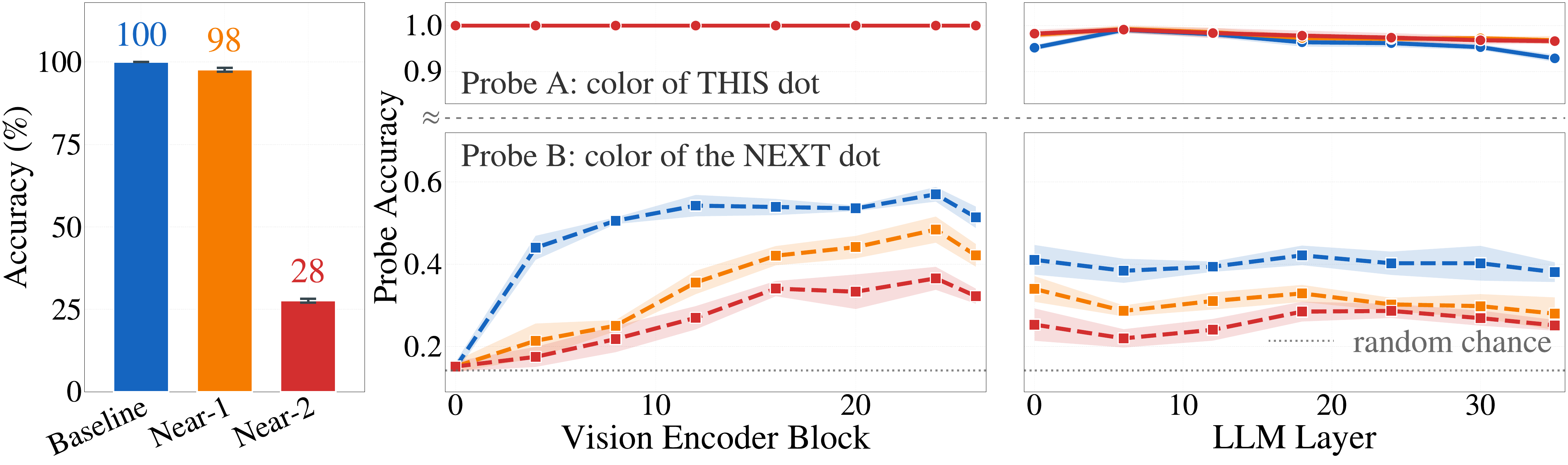}
\vskip -0.1em
\caption{
\textbf{Task performance and probing accuracy of Qwen3-VL-8B under nearby-path competition.}
Left: Task accuracy of Qwen3-VL-8B across the \textit{Baseline}, \textit{Near-1}, and \textit{Near-2} conditions, showing the effect of nearby paths.
Right: Linear probing accuracy of Qwen3-VL-8B vision-encoder and LLM representations. \textit{Probe~A} measures whether the representation of a dot patch preserves its own color identity, while \textit{Probe~B} measures whether it contains information about the next dot along the queried path. The dashed line marks random-choice accuracy ($1/7$) among the seven remaining palette colors for next-dot prediction.
}
\label{fig:acc_probe}
\vspace{-1.0em}
\end{figure}

\begin{figure*}[t]
    \centering

    \begin{subfigure}[t]{\textwidth}
        \centering
        \includegraphics[width=\linewidth]{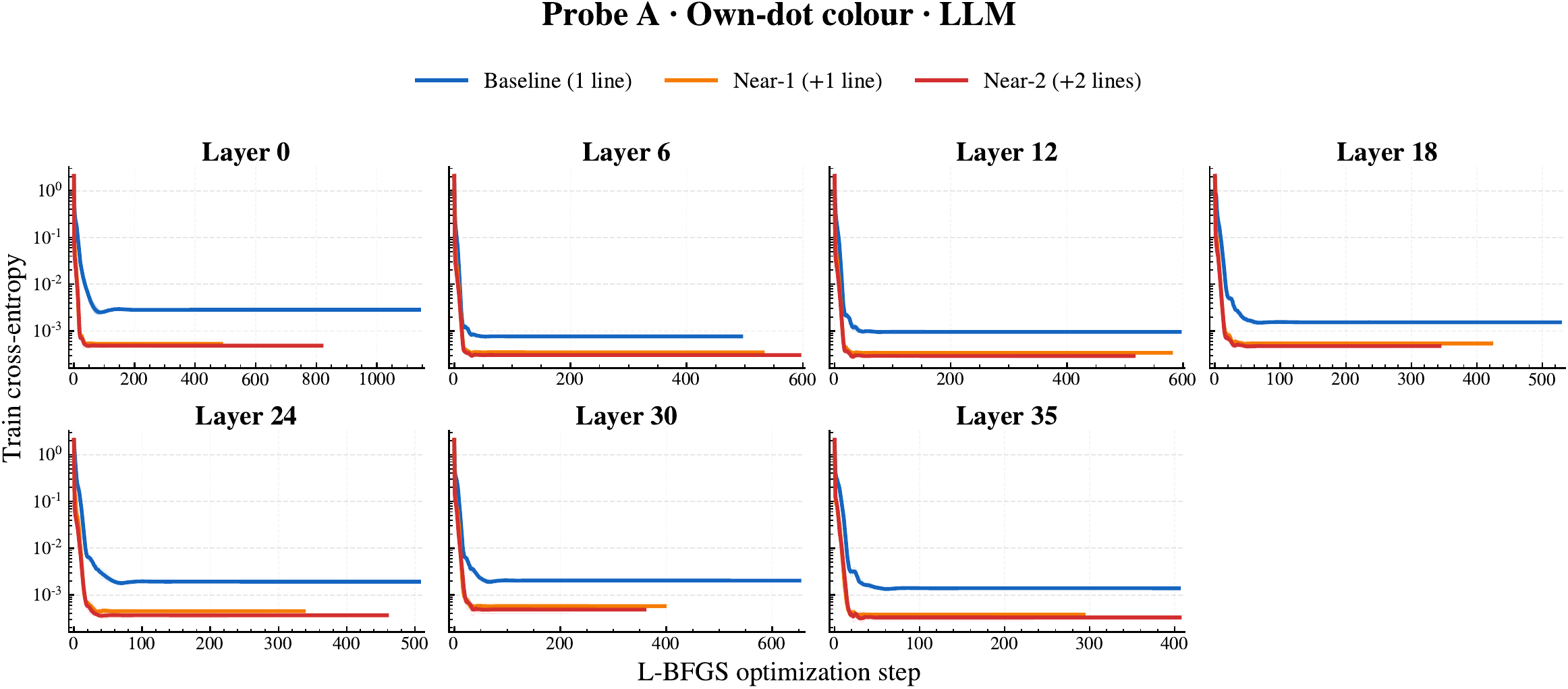}
        \caption{LLM layers.}
        \label{fig:probe_A_train_steps_llm}
    \end{subfigure}

    \vspace{0.6em}

    \begin{subfigure}[t]{\textwidth}
        \centering
        \includegraphics[width=\linewidth]{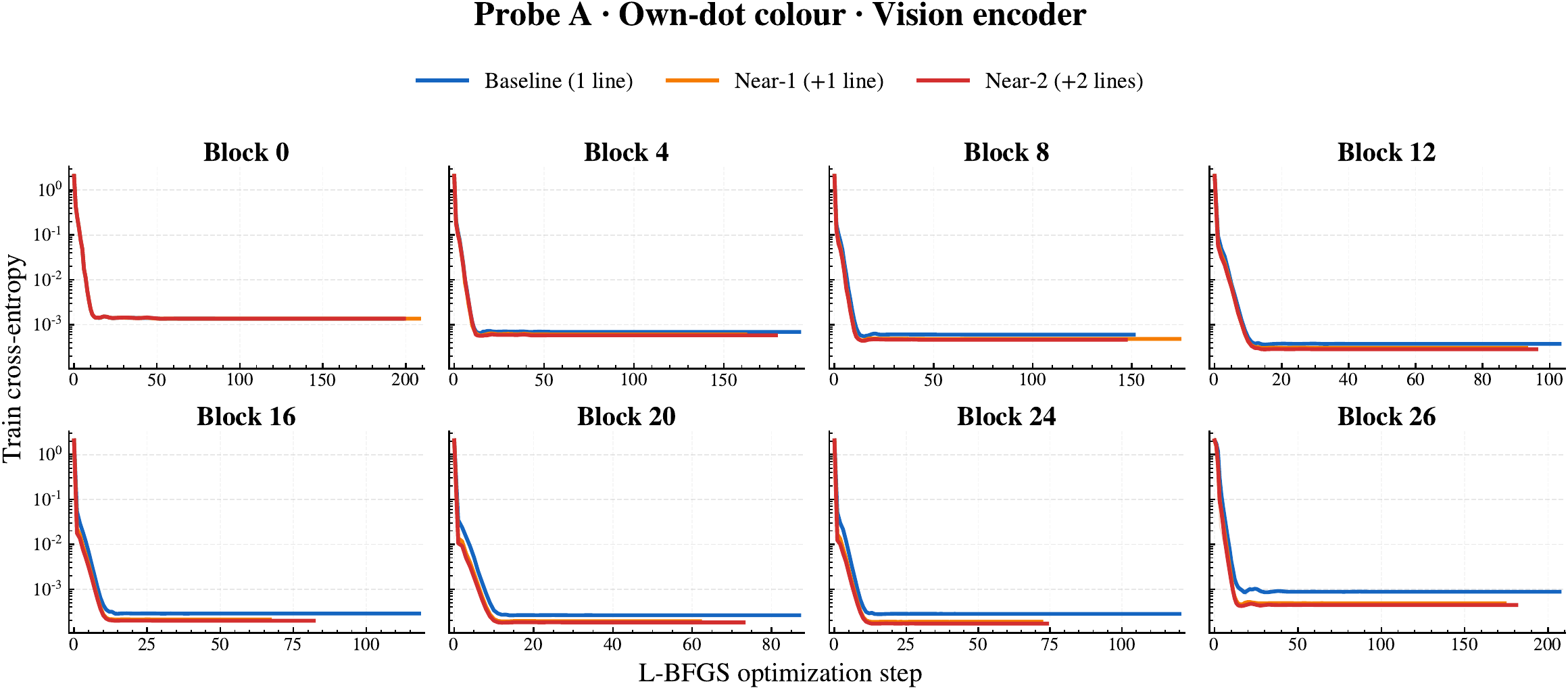}
        \caption{Vision-encoder blocks.}
        \label{fig:probe_A_train_steps_vis}
    \end{subfigure}

    \caption{
    \textbf{Probe~A optimization across model representations.}
    Training cross-entropy across L-BFGS optimization steps for \textit{Baseline}, \textit{Near-1}, and \textit{Near-2} in the LLM (top) and vision encoder (bottom).
    }
    \label{fig:probe_A_train_steps}
\end{figure*}

\begin{figure*}[t]
    \centering

    \begin{subfigure}[t]{\textwidth}
        \centering
        \includegraphics[width=\linewidth]{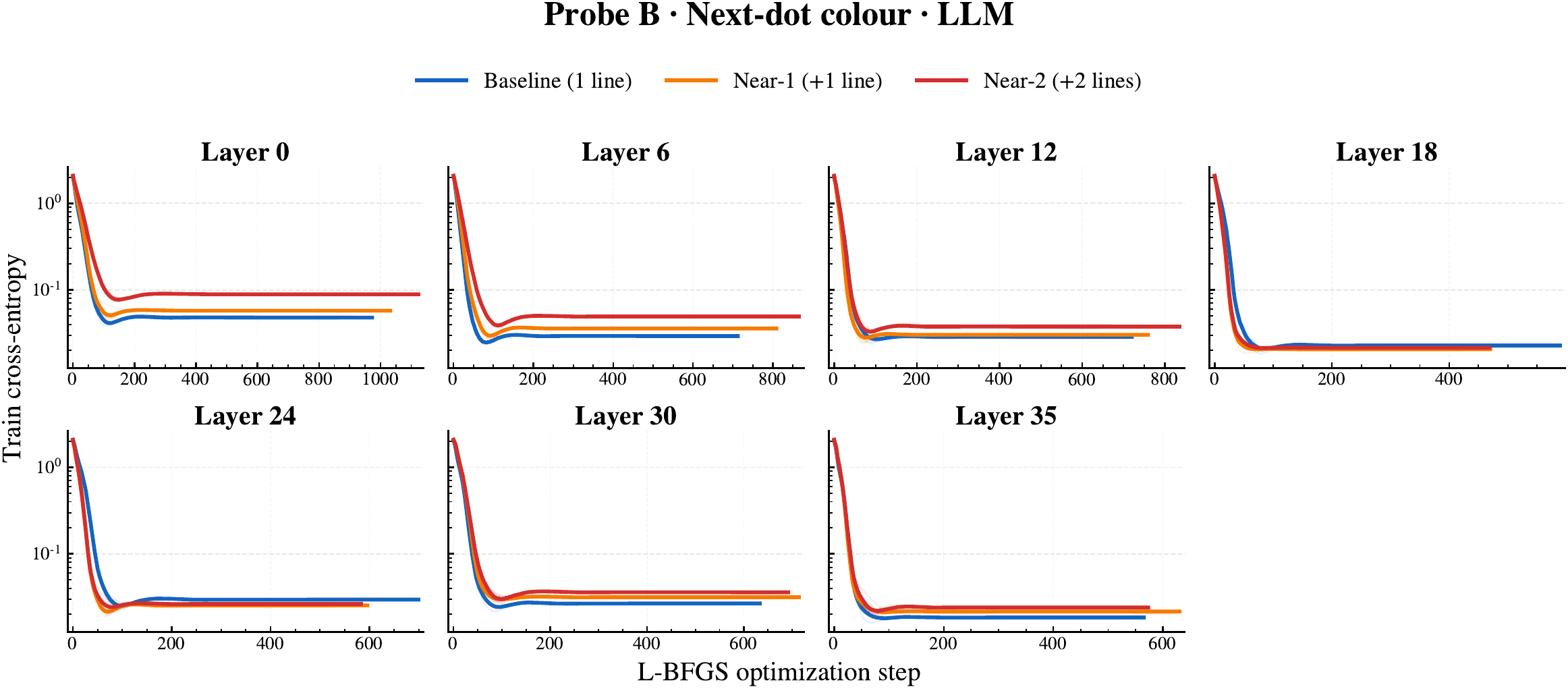}
        \caption{LLM layers.}
        \label{fig:probe_B_train_steps_llm}
    \end{subfigure}

    \vspace{0.6em}

    \begin{subfigure}[t]{\textwidth}
        \centering
        \includegraphics[width=\linewidth]{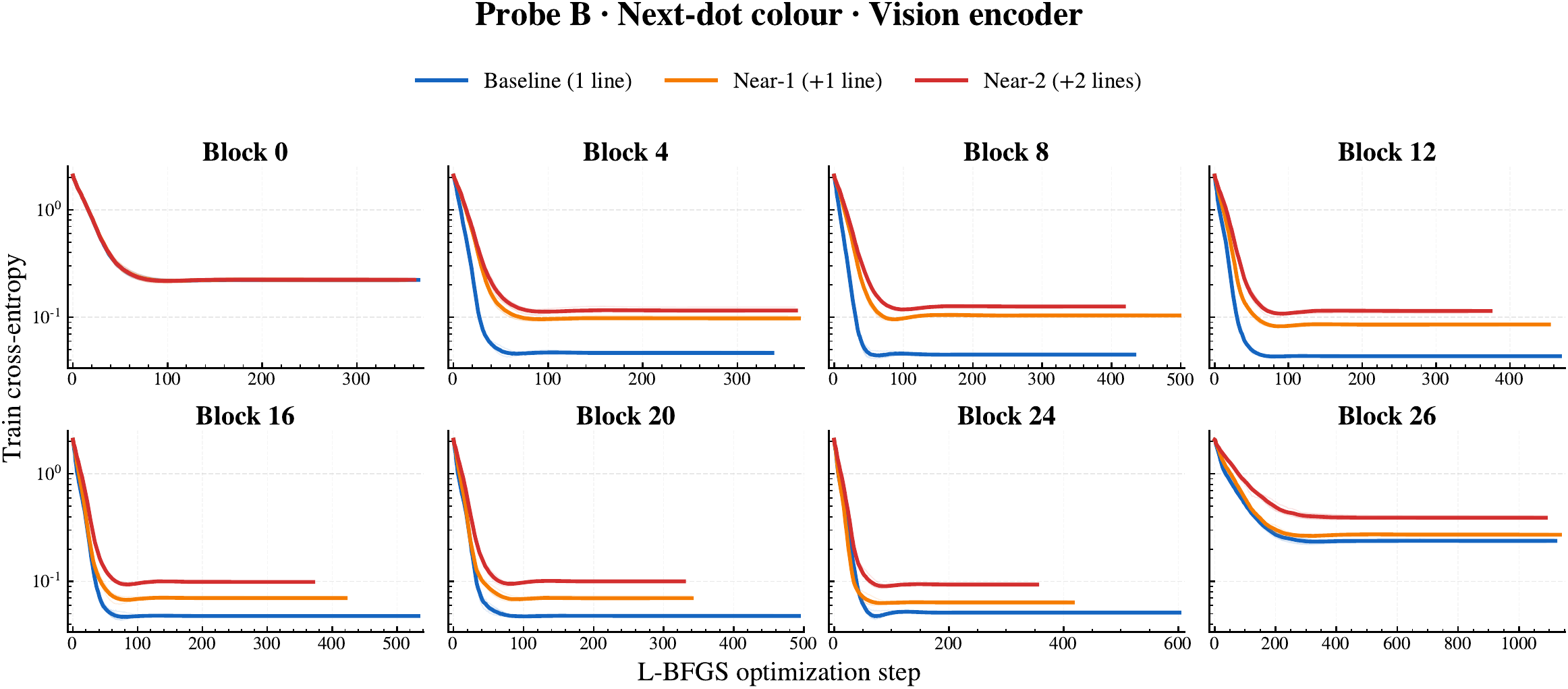}
        \caption{Vision-encoder blocks.}
        \label{fig:probe_B_train_steps_vis}
    \end{subfigure}

    \caption{
    \textbf{Probe~B optimization across model representations.}
    Training cross-entropy across L-BFGS optimization steps for \textit{Baseline}, \textit{Near-1}, and \textit{Near-2} in the LLM (top) and vision encoder (bottom).
    }
    \label{fig:probe_B_train_steps}
\end{figure*}

\paragraph{Optimization and training loss.}
We additionally track the training cross-entropy throughout L-BFGS optimization to ensure that differences between conditions are not caused by incomplete probe training. We use an optimization tolerance of $10^{-4}$, and allow up to 2,000 iterations for convergence. For each condition, layer, and cross-validation fold, we record the cross-entropy after each accepted optimization step. The plotted loss excludes the L2 penalty and therefore reflects how well the probe fits the training labels directly. The individual fold trajectories and their mean $\pm$ SEM are shown in \Cref{fig:probe_A_train_steps_llm,fig:probe_A_train_steps_vis,fig:probe_B_train_steps_llm,fig:probe_B_train_steps_vis}. Across Probe~A and Probe~B, the optimization curves reach stable plateaus under this convergence criterion for all three conditions. We therefore compare held-out probe accuracy only after the probes are sufficiently optimized, reducing the possibility that the lower accuracy under \textit{Near-1} or \textit{Near-2} is simply due to slower or incomplete optimization.

\subsection{Replication with InternVL3.5-8B.}

\begin{figure}[htbp!]
\centering
\includegraphics[width=\textwidth]{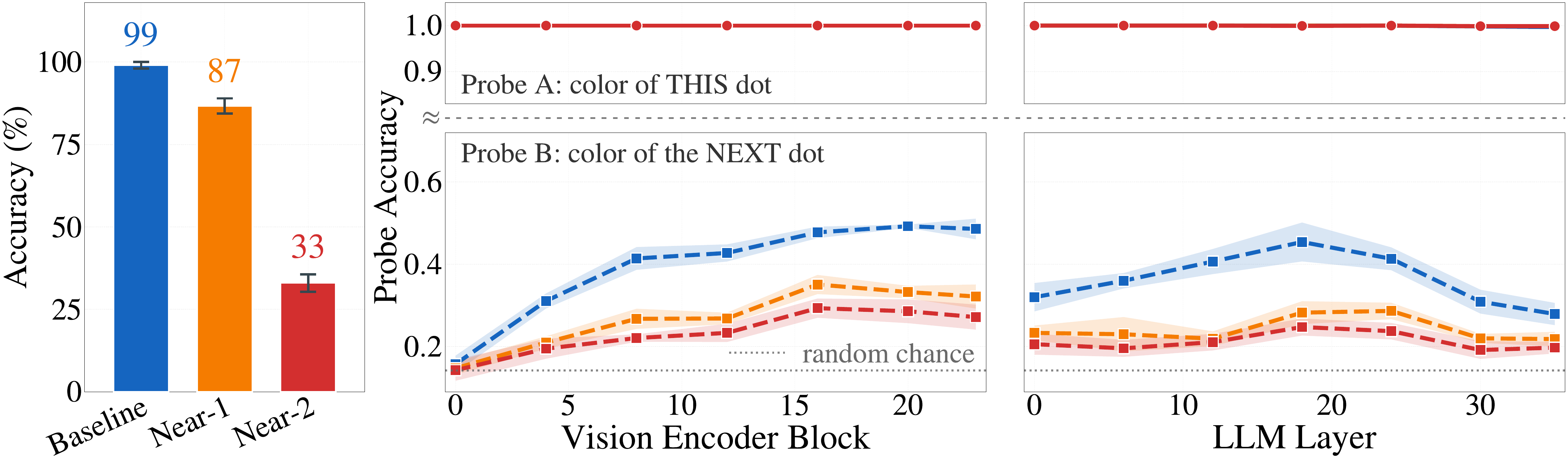}
\vskip -0.1em
\caption{
\textbf{Task performance and probing results for InternVL3.5-8B under nearby-path competition.}
\textbf{(a)} Tracing accuracy across the \textit{Baseline}, \textit{Near-1}, and \textit{Near-2} conditions.
\textbf{(b)} Linear probing of vision-encoder and LLM representations. Probe~A predicts the probed dot's own color, while Probe~B predicts the next dot color along the queried path. The dashed baseline indicates random-choice accuracy ($1/7$).
}
\label{fig:internvl_acc_probe}
\vspace{-0.5em}
\end{figure}

InternVL3.5-8B shows a similar overall trend (\Cref{fig:internvl_acc_probe}). Tracing accuracy decreases from 99\% in \textit{Baseline} to 87\% in \textit{Near-1} and 33\% in \textit{Near-2}. Probe~A remains high across conditions, while Probe~B weakens as nearby paths are introduced, indicating reduced information about which dot comes next along the queried path. This replication suggests that the sensitivity of continuation information to nearby paths is not specific to Qwen3-VL-8B.

\clearpage
\section{Control Experiments with Non-Line Visual Elements}
\label{app:distractor_type_controls}

\begin{figure*}[htbp!]
    \centering
    \includegraphics[width=\textwidth]{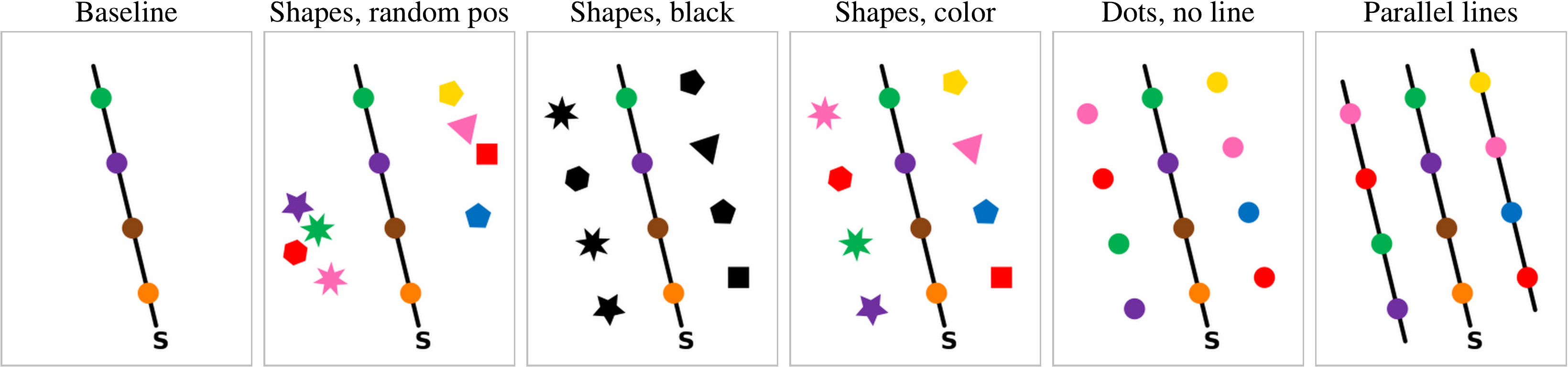}
    \caption{
    \textbf{Representative examples of non-line controls.}
    The controls vary shape, color, position, and connectivity while keeping the queried path fixed.
    }
    \label{fig:distractor_type_examples}
\end{figure*}

To test whether tracing failure is caused simply by nearby visual content or specifically by competing traceable structure, we conduct this control analysis with Qwen3-VL-8B~\citep{qwen3_vl}, replacing the two parallel lines in \textit{Near-2} with several non-line controls while keeping the queried path unchanged. Representative examples are shown in \Cref{fig:distractor_type_examples}.

\paragraph{Control Construction.}
We compare six conditions while keeping the queried path unchanged. \textit{Baseline} contains only the queried path with no nearby visual elements. \textit{Shapes, Random Position} places colored shapes at random nearby positions, breaking their aligned arrangement while ensuring that they are not closer to the queried path than the original \textit{Near-2} elements. \textit{Shapes, Black} places black shapes at the aligned \textit{Near-2} positions, while \textit{Shapes, Color} uses randomly selected colored shapes at the same positions. \textit{Dots, No Line} preserves the colors and positions of all eight nearby dots from \textit{Near-2} but removes the connecting lines. Finally, \textit{Parallel Lines} corresponds to the original \textit{Near-2} condition, where the aligned colored dots are connected into two parallel paths. For the shape-based controls, the rendered shape area is matched to that of a colored dot. Across conditions, the queried path, its rotation, and its color sequence remain fixed.

\paragraph{Competing Traceable Structure Drives the Behavioral Failure.}
As shown in \Cref{tab:distractor_type_prefix}, nearby non-line visual elements have almost no effect on tracing. Randomly positioned colored shapes preserve 100\% prefix accuracy across all four steps, while aligned black shapes, colored shapes, and colored dots without connecting lines remain between 98--100\%. In contrast, connecting the aligned colored dots into two continuous parallel lines causes cumulative accuracy to fall from 64.3\% at the first dot to 27.7\% after all four dots. The non-line controls produce at most a 0--2 pp decline across the trace, whereas \textit{Near-2} already loses 35.7 pp at the first dot and continues to accumulate errors at subsequent steps. These results indicate that nearby visual content, color, or aligned placement alone does not reproduce the failure; substantial path-preservation failure emerges when the nearby structure forms an alternative continuous path that can itself be traced.

\begin{table*}[htbp!]
\centering
\caption{
\textbf{Cumulative prefix accuracy across control types.}
Accuracy (\%) after correctly tracing the first $k$ dots. Values report mean $\pm$ standard deviation across repeated evaluations.
}
\label{tab:distractor_type_prefix}
\small
\setlength{\tabcolsep}{9pt}
\renewcommand{\arraystretch}{1.12}
\begin{tabular}{lcccc}
\toprule
\textbf{Control Type} & $\mathbf{k=1}$ & $\mathbf{k=2}$ & $\mathbf{k=3}$ & $\mathbf{k=4}$ \\
\midrule

None
& $100.0 \pm 0.0$
& $100.0 \pm 0.0$
& $100.0 \pm 0.0$
& $100.0 \pm 0.0$ \\

Shapes, Random Position
& $100.0 \pm 0.0$
& $100.0 \pm 0.0$
& $100.0 \pm 0.0$
& $100.0 \pm 0.0$ \\

Shapes, Black
& $100.0 \pm 0.0$
& $100.0 \pm 0.0$
& $99.0 \pm 1.0$
& $99.0 \pm 1.0$ \\

Shapes, Color
& $99.0 \pm 0.0$
& $99.0 \pm 0.0$
& $98.0 \pm 0.0$
& $98.0 \pm 0.0$ \\

Dots, No Line
& $100.0 \pm 0.0$
& $99.3 \pm 0.6$
& $99.3 \pm 0.6$
& $99.3 \pm 0.6$ \\

\textbf{Parallel Lines (Near-2)}
& $\mathbf{64.3 \pm 0.6}$
& $\mathbf{47.7 \pm 0.6}$
& $\mathbf{36.7 \pm 0.6}$
& $\mathbf{27.7 \pm 0.6}$ \\

\bottomrule
\end{tabular}
\end{table*}

\begin{figure*}[t]
    \centering
    \includegraphics[width=0.78\textwidth]{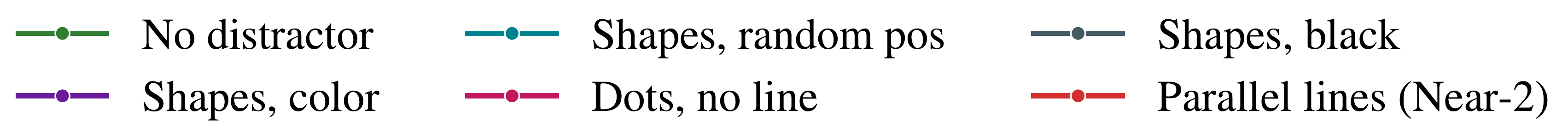}
    \vspace{0.4em}
    \includegraphics[width=\textwidth]{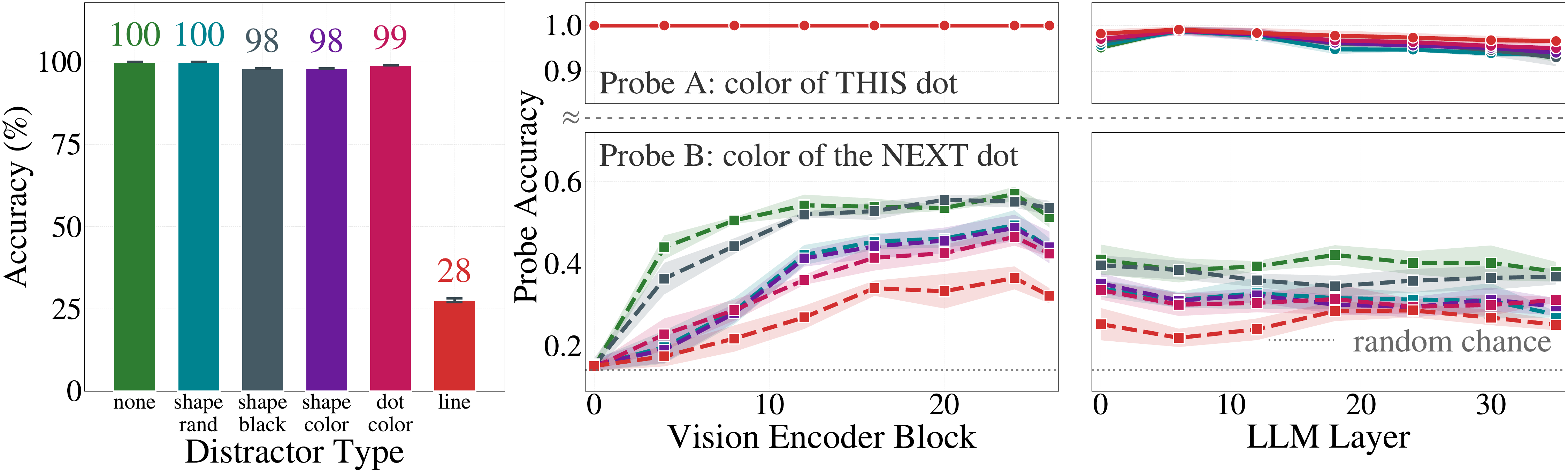}
    \caption{
    \textbf{Behavioral performance and internal signals across control types.}
    Task accuracy is compared with Probe~A, which measures information about each dot's own color, and Probe~B, which measures information about the next dot along the queried path. Non-line visual elements largely preserve task performance, whereas the connected parallel-line condition produces substantial behavioral degradation.
    }
    \label{fig:distractor_type_probe}
\end{figure*}

\paragraph{Path-Continuation Information Is More Sensitive Than Local Dot Identity.}
The two probes show different sensitivity to nearby visual elements (\Cref{fig:distractor_type_probe}). Probe~A remains perfect throughout the vision encoder across all control conditions and above 0.93 in the LLM representations, indicating that the color identity of individual dots remains well represented. Probe~B is more sensitive: at vision block 24, accuracy decreases from 0.570 with no nearby element to 0.556 with aligned black shapes, 0.494 with randomly positioned colored shapes, 0.487 with aligned colored shapes, 0.466 with colored dots without connecting lines, and 0.366 with parallel lines. Because Probe~B explicitly predicts the next-dot color, some reduction in the presence of additional colored elements is expected and may reflect interference from the verifier colors themselves. Consistent with this interpretation, these colored non-line controls retain near-perfect tracing accuracy despite lower Probe~B accuracy.

The key difference appears when the same nearby colored elements are connected into continuous parallel paths: Probe~B drops further to 0.366, accompanied by a sharp decline in tracing performance. Thus, the main degradation emerges not from nearby colors alone, but when they form competing traceable paths. Together with our appearance-similarity analysis (Appendix~\ref{app:distractor_analysis}), these results further show that path preservation becomes increasingly fragile as nearby structures more closely resemble competing continuations.


\section{Effect of Appearance Similarity on Path Preservation}
\label{app:distractor_analysis}

We further examine how similarity between the paths themselves affects path following under local competition. Specifically, we vary the visual and structural similarity of nearby paths to the queried path and measure how tracing performance changes. We conduct the additional analysis on InternVL3.5-8B~\citep{internvl3.5}.

\subsection{Similar Nearby Paths Break Path Preservation}

As shown in \Cref{tab:condition_acc}, accuracy drops sharply in the \textit{Shared segment} condition and is highest in the \textit{Different all} condition. If the model were failing because of a generic inability to read colored dots or follow the task, performance should be similarly poor across conditions. Instead, the substantial recovery as similarity is reduced indicates that path preservation becomes increasingly fragile when nearby paths more closely resemble the queried path.

\begin{table}[htbp!]
\centering
\caption{
\textbf{Accuracy across local similarity conditions.}
Tracing accuracy of Qwen3-VL-8B and InternVL3.5-8B under varying appearance similarity between nearby paths and the queried path. Results are reported as mean $\pm$ standard deviation over three runs.
}
\label{tab:condition_acc}
\label{tab:internvl_condition_acc}
\begin{tabular}{lcc}
\toprule
Condition & Qwen3-VL-8B & InternVL3.5-8B \\
\midrule
Shared segment & $76.67 \pm 0.76$ & $50.50 \pm 2.50$ \\
Different segment & $95.17 \pm 0.29$ & $80.50 \pm 1.32$ \\
Different angle & $94.67 \pm 0.29$ & $65.50 \pm 2.29$ \\
Different all & $99.67 \pm 0.29$ & $86.83 \pm 2.08$ \\
\bottomrule
\end{tabular}
\end{table}

\subsection{Internal States Become Less Selective for the True Continuation}
\label{app:internal}


We examine how \textit{attention preference} in the vision encoder and \textit{representational preference} across the vision encoder and LLM decoder change with the appearance similarity of nearby paths. Because the four-dot paths yield fewer valid next-step samples and make the main probing setup in the previous section less directly comparable, we use these preference-based measures instead. If tracing is robust, both attention and representations should remain selective for the true continuation across similarity conditions.

\paragraph{Attention Preference Extraction.}
We compute attention preference using Qwen3-VL-8B and InternVL3.5-8B by mapping the queried-path and nearby-path dot coordinates to their corresponding vision-token regions. Each dot is represented by a \(3{\times}3\) pre-merge patch neighborhood centered on its coordinate to reduce sensitivity to patch-alignment errors. Specifically, we examine the transition from \(q_2\) to \(q_3\) on the queried path, where local appearance similarity is controlled across conditions. We compare attention from \(q_2\) to the correct continuation \(q_3\) against attention from \(q_2\) to the corresponding second-position dots on the two nearby paths, \(d_{1,2}\) and \(d_{2,2}\). At each vision-encoder block, we average attention over heads and source tokens.

\paragraph{Representational Preference Extraction.}
For the representation analysis, we register hooks on the output of each vision-encoder block and each LLM decoder layer. For each dot region, we average the hidden states over its \(3{\times}3\) image-token neighborhood to obtain one region representation. Before computing cosine similarity, hidden states are mean-centered within the layer. We then compute the cosine similarity between the queried-path representation at \(q_2\) and its true continuation \(q_3\), and compare it with the larger similarity between \(q_2\) and the corresponding second dots on the two nearby paths, \(d_{1,2}\) and \(d_{2,2}\).

\begin{figure}[htbp!]
\centering

\includegraphics[width=0.80\textwidth]{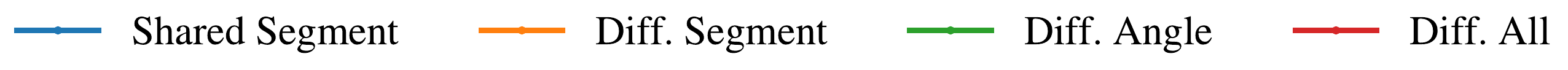}

\vspace{0.4em}

\includegraphics[width=\textwidth]{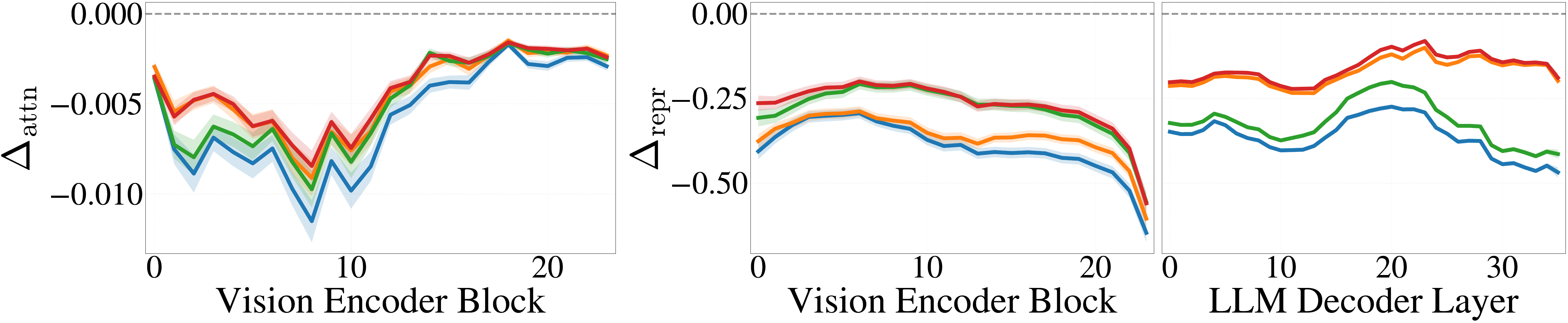}

\vskip -0.1em
\caption{
\textbf{Internal preference in InternVL3.5-8B under varying appearance similarity.}
Left: Block-wise attention preference ($\Delta_{\mathrm{attn}}$) in the vision encoder, comparing attention from the second queried-path dot to the correct third dot against attention to the corresponding second-position dots on nearby paths. Right: Representational preference ($\Delta_{\mathrm{repr}}$) across the vision encoder and LLM decoder, comparing similarity to the true continuation against similarity to the nearby-path alternatives. Positive values indicate greater preference for the correct continuation.
}
\label{fig:internal_selectivity}
\vspace{-0.5em}
\end{figure}

\paragraph{InternVL3.5-8B Result.}
We observe the same overall pattern in InternVL3.5-8B. Tracing accuracy decreases as nearby paths become more similar to the queried path, with the lowest performance in the \textit{Shared segment} condition and higher accuracy as local similarity is reduced (\Cref{tab:condition_acc}). The internal analysis shows a corresponding trend: attention preference for the true continuation is lowest when nearby paths share the strongest local similarity with the queried path and increases as this similarity is reduced, while representational preference likewise decreases under greater nearby-path similarity across the vision encoder and LLM decoder (\Cref{fig:internal_selectivity}). These results replicate the same sensitivity to nearby-path appearance observed in Qwen3-VL-8B, indicating that the effect is not specific to a single model.

\clearpage
\subsection{Additional Masking Intervention Results}
\label{app:masking_results}

We perform a masking intervention to test whether locally similar nearby-path cues directly contribute to tracing failures across all similarity conditions. For each intervention, we mask a $3{\times}3$ image-token patch centered on the specified target: the corresponding dot on the nearby path, the queried-path dot, or a randomly selected location. Random masks are sampled only from object-free regions, excluding a $5{\times}5$ area around every dot center to avoid overlap with task-relevant elements. We use the same prompt as in the Swirl task construction (Appendix~\ref{app:swirl_task_construction}) and evaluate performance using exact-match scoring.

\begin{figure}[htbp!]
\centering
\includegraphics[width=0.65\textwidth]{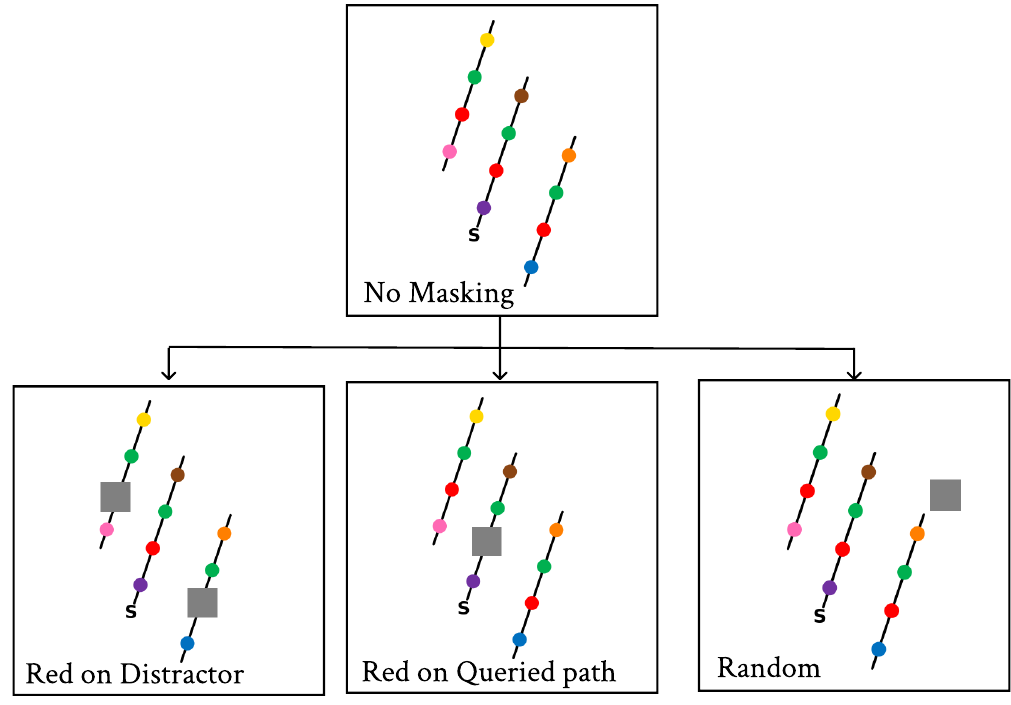}
\caption{
\textbf{Causal masking intervention.}
Example shown for the \textit{Shared segment} condition. Masking is applied around the local transition where the model must preserve the queried path against a nearby competing continuation. We compare masking the corresponding dot on the nearby path, the queried-path dot, and a random image region.
}
\label{fig:causal_masking}
\vspace{-0.5em}
\end{figure}

\paragraph{Causal Masking Confirms Nearby-Path Interference.}
The intervention setup is illustrated in \Cref{fig:causal_masking}, and the full results across all conditions are reported in \Cref{tab:masking_all_conditions}. Across all conditions, masking a random patch has little effect, while masking the queried-path region sharply reduces accuracy, confirming that this region is necessary for correct tracing. In contrast, masking the corresponding nearby-path region improves performance, with the largest recovery in the \textit{Shared segment} condition, where accuracy increases from 76.5\% to 90.0\% (+13.5 points). The gain becomes smaller as local similarity is reduced and is negligible in \textit{Different all}, providing causal evidence that locally similar nearby-path cues interfere most strongly with preservation of the queried path.

\begin{table}[htbp]
\centering
\small
\caption{
\textbf{Patch masking accuracy across all path conditions.}
All results use a $3{\times}3$ masking patch. Values in parentheses indicate the change relative to the \textit{No Masking} baseline within the same condition.
}
\vspace{1.0em}
\label{tab:masking_all_conditions}
\begin{tabular}{lcccc}
\toprule
Masking Setup 
& Shared Segment
& Different Segment 
& Different Angle 
& Different All \\
\midrule
No Masking 
& 76.5 
& 95.0 
& 94.5 
& 99.5 \\

Nearby-Path Dot 
& 90.0 \; (\textbf{+13.5}) 
& 98.0 \; (+3.0) 
& 99.0 \; (+4.5) 
& 99.5 \; (+0.0) \\

Queried-Path Dot 
& 2.0 \; (-74.5) 
& 0.5 \; (-94.5) 
& 0.5 \; (-94.0) 
& 0.0 \; (-99.5) \\

Random 
& 76.5 \; (+0.0) 
& 94.5 \; (-0.5) 
& 94.0 \; (-0.5) 
& 99.0 \; (-0.5) \\
\bottomrule
\end{tabular}
\end{table}

\clearpage

\section{Additional Analysis of Reasoning Traces}
\label{app:reasoning_trace_analysis}

\subsection{Self-Correction Frequency Estimation}
\label{app:self_correction_rate}

We estimate self-correction frequency from the generated reasoning traces rather than from the final answers alone. Because the reasoning outputs differ substantially across models, we use model-specific textual markers that indicate revision, rechecking, or reconsideration. For Sonnet-4.5, GPT-5.4, and Gemini3-Flash, which do not expose the full reasoning trace, we apply the same analysis to their provided reasoning summaries. The goal is to quantify self-correction frequency by detecting explicit markers that the model revises or re-evaluates its current solution.

\begin{table}[htbp]
\centering
\small
\setlength{\tabcolsep}{6pt}
\renewcommand{\arraystretch}{1.2}
\caption{
\textbf{Self-correction markers used to estimate self-correction frequency.}
}
\label{tab:self_correction_signals}
\vspace{1.0em}
\begin{tabular}{p{0.28\linewidth} p{0.62\linewidth}}
\toprule
\textbf{Model} & \textbf{Self-correction markers} \\
\midrule
GPT-5.4 &
\textit{wait}, \textit{actually}, \textit{revising}, \textit{rechecking}, \textit{reconsidering} \\
\midrule
Sonnet-4.5 &
\textit{wait}, \textit{actually}, \textit{let me check}, \textit{let me recheck}, \textit{re-examine} \\
\midrule
Gemini3-Flash &
Bold revision headers containing correction-related terms, including \textit{revise/revising}, \textit{revisit}, \textit{re-examine}, \textit{reconsider}, \textit{recheck}, \textit{correct}, \textit{fix}, \textit{adjust}, \textit{amend}, \textit{backtrack}, \textit{wrong}, \textit{error}, or \textit{mistake} \\
\midrule
Qwen3-VL-235B &
\textit{wait}, \textit{actually}, \textit{let me check}, \textit{let me recheck}, \textit{re-examine}, and \textit{reconsider} \\
\midrule
GLM-4.6V &
\textit{wait}, \textit{let's check}, \textit{let me confirm}, and \textit{let's look again} \\
\midrule
Kimi-K3 &
\textit{wait}, \textit{actually}, \textit{hmm}, \textit{need verify}, \textit{need check}, and \textit{need re-check} \\
\bottomrule
\end{tabular}
\end{table}

For model $m$ at rotation density level $\ell$, we define \textit{self-correction frequency} as the average number of detected self-correction markers per reasoning trace:

\begin{equation}
F(m,\ell)
=
\frac{1}{\left|\mathcal{D}_{\ell}^{(m)}\right|}
\sum_{i \in \mathcal{D}_{\ell}^{(m)}}
c\!\left(t_i^{(m)}\right),
\label{eq:sc_frequency}
\end{equation}

where $\mathcal{D}_{\ell}^{(m)}$ denotes the set of evaluated samples for model $m$ at density level $\ell$, $t_i^{(m)}$ is the corresponding reasoning trace for sample $i$, and $c(t_i^{(m)})$ denotes the number of detected self-correction markers in that trace. Higher self-correction frequency therefore indicates more frequent revisions or reconsiderations during reasoning.

\subsection{Keyword-Based Analysis of Reasoning-Style Shifts}
\label{app:reasoning_keyword_analysis}

To quantify how visible reasoning patterns change with rotation density, we use the brief sequential tracing observed at \textit{Low} density as a reference and identify lexical cues associated with alternative, non-tracing strategies. Each reasoning trace is matched against a fixed set of keyword and regular-expression patterns grouped into three categories: angular matching, ring decomposition, and radial estimation. \textit{Angular matching} refers to reasoning that orders dots using their angular positions rather than visible line connectivity. \textit{Ring decomposition} refers to partitioning the spiral into inner, middle, or outer loops and reasoning over these groups separately. \textit{Radial estimation} refers to using estimated coordinates or distances from the spiral center to determine the dot sequence. These categories were manually defined by the authors as a coarse quantitative proxy for changes in visible reasoning style, rather than an exhaustive or absolute classification of model reasoning. The complete mapping is provided in \Cref{tab:reasoning_keyword_mapping}. We count all matched occurrences within each trace and report their average frequency per sample at \textit{Low}, \textit{Moderate}, and \textit{High} density.

\begin{table}[htbp!]
\centering
\caption{\textbf{Keyword mappings for non-tracing reasoning patterns.}
Exact lexical and regular-expression patterns used to count each reasoning category.}
\label{tab:reasoning_keyword_mapping}
\small
\setlength{\tabcolsep}{5pt}
\renewcommand{\arraystretch}{1.15}
\begin{tabular}{p{0.27\linewidth}p{0.65\linewidth}}
\toprule
\textbf{Category} & \textbf{Keywords / patterns used for counting} \\
\midrule

Angular matching
& Numerical angle expressions ending in \texttt{\textdegree}, \texttt{degree}, or \texttt{degrees};
\texttt{top-right}, \texttt{top-left}, \texttt{bottom-right}, \texttt{bottom-left}
(and their space-separated forms). \\

Ring decomposition
& \texttt{inner ring}, \texttt{inner loop}, \texttt{inner circle}, \texttt{inner spiral},
\texttt{outer ring}, \texttt{outer loop}, \texttt{outer circle}, \texttt{outer spiral},
\texttt{innermost}, \texttt{outermost}. \\

Radial estimation
& Pixel-coordinate expressions of the form \texttt{(x, y)};
\texttt{radius}, \texttt{radii}, and expressions of the form \texttt{r = [number]}. \\

\bottomrule
\end{tabular}
\end{table}

\begin{figure}[t]
    \centering
    \includegraphics[width=0.82\linewidth]{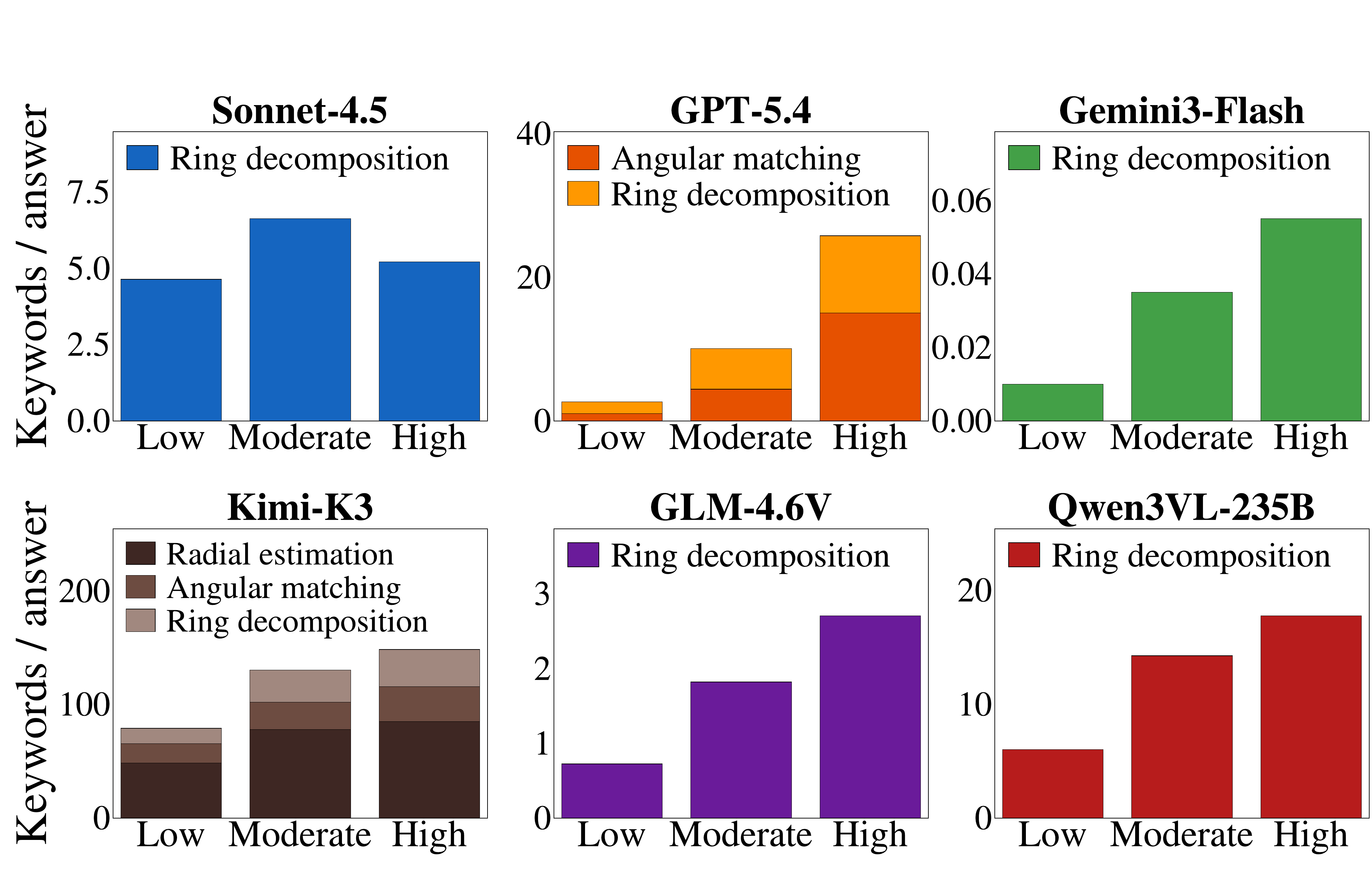}
    \caption{
    \textbf{Non-tracing reasoning-pattern intensity across Swirl density.}
    Average keyword-match frequency for non-tracing reasoning patterns at \textit{Low}, \textit{Moderate}, and \textit{High} rotation density. Higher values indicate more frequent lexical cues associated with global geometric reasoning.
    }
    \label{fig:substitution_intensity}
\end{figure}

As shown in \Cref{fig:substitution_intensity}, non-tracing reasoning patterns generally become more prominent as rotation density increases. With the exception of Sonnet-4.5, all models show higher keyword-match frequency at higher density, indicating more frequent use of global geometric constructs such as rings, angles, radii, and coordinates. Gemini3-Flash exhibits substantially lower overall intensity than the other models, likely because it provides only a limited reasoning summary rather than the full reasoning trace, reducing the amount of detectable strategy-related vocabulary. Because this analysis relies on keyword-based matching, it should be interpreted only as an indirect indicator of trends in visible reasoning style rather than an absolute measure of strategy use or internal reasoning.

\subsection{Additional Reasoning Results for Qwen3-VL-235B and GLM-4.6V}
\label{app:qwen_glm_reasoning}

\begin{figure}[htbp!]
\centering
\includegraphics[width=0.50\linewidth]{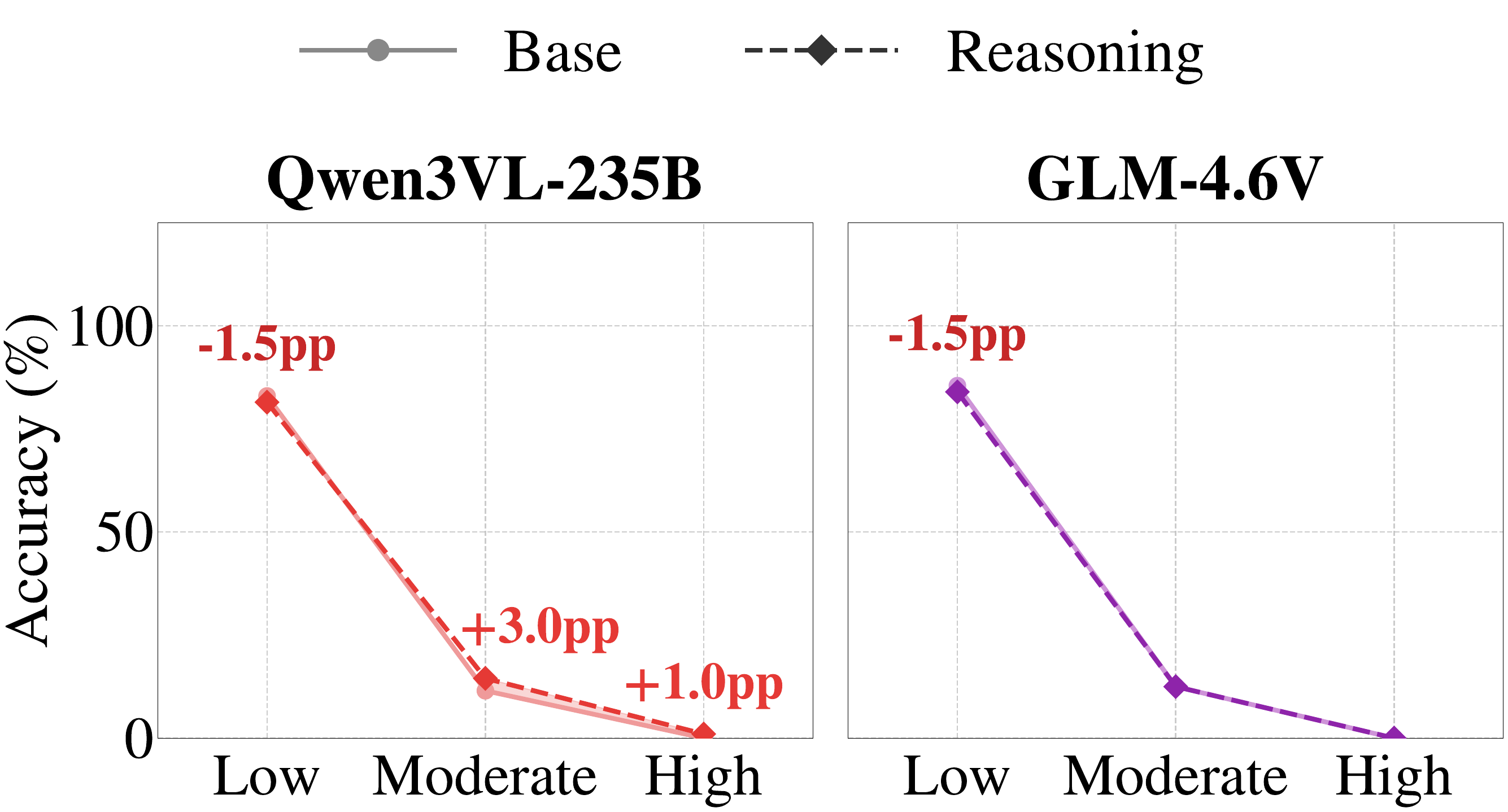}
\caption{
\small
\textbf{Additional reasoning results for Qwen3-VL-235B and GLM-4.6V.}
Accuracy of \textit{Base} and \textit{Reasoning} models on Swirl across rotation-density levels.
}
\label{fig:qwen_glm_reasoning}
\end{figure}

We further examine Qwen3-VL-235B and GLM-4.6V, which are omitted from the main-text reasoning analysis for brevity. As shown in \Cref{fig:qwen_glm_reasoning}, neither model exhibits a substantial improvement in tracing performance when reasoning is enabled. These results provide additional evidence that extended reasoning does not consistently compensate for the lack of robustness in visual tracing.

\subsection{Reasoning Trace Examples}
\label{app:reasoning-trace-examples}

We provide qualitative examples of how reasoning changes with rotation density across models. At \textit{Low} density, models generally describe the path through brief sequential tracing, proceeding from one connected dot to the next; representative examples for each model are shown in \Cref{tab:brief_tracing_low}. At \textit{Moderate} and \textit{High} density, however, reasoning increasingly includes alternative strategies that depart from this sequential procedure, such as ring decomposition, angular or directional matching, coordinate estimation, and repeated direction reconsideration. Representative examples of these density-dependent reasoning patterns are provided in \Cref{tab:substitution_reasoning}.

\begin{table}[htbp]
    \caption{
    \small
    \textbf{Representative reasoning traces at the Low difficulty level.}
    }
    \vspace{1.5em}
    \centering
    \small
    \renewcommand{\arraystretch}{1.35}
    \begin{tabularx}{\linewidth}{@{} l >{\raggedright\arraybackslash}p{2.8cm} X @{}}
    \toprule
    \textbf{Model} & \textbf{Reasoning Type} & \textbf{Generated Reasoning} \\
    \midrule
    Sonnet-4.5 &
    Brief spatial tracing &
    ``Let me \textbf{trace the path from `S'} in the image. Starting at `S',
    I need to \textbf{follow the line} and identify the colored dots
    \textbf{in sequence}. Looking at the image: S is at a purple dot;
    \textbf{following the line from S}, it goes to a yellow dot, then to an
    orange dot, then to a green dot, then to a red dot, then to a blue dot,
    then to a brown dot, then to a pink dot, \textbf{then to E}.'' \\
    \midrule
    GPT-5.4 &
    Brief spatial tracing &
    ``\textbf{Listing colors in order.} I'll note the colors \textbf{directly
    as I trace} the spiral path, \textbf{starting from `S' and going outward}.
    The colors I find in order are: red, green, pink, purple, orange, yellow,
    brown, blue, and then `E' to wrap it up with ``end'' at the finish.'' \\
    \midrule
    Gemini3-Flash &
    Brief spatial tracing &
    ``\textbf{Mapping the Sequence.} I've pinpointed the `S' and, as
    anticipated, it's \textbf{directly adjacent to an orange dot}. Now, the
    tracing begins! I'm ready to \textbf{follow the line, dot by dot}, from the
    orange starting point. \textbf{Defining the Path.} I have finalized the
    color sequence: orange, red, yellow, blue, purple, green, brown, pink, and
    end. The spiral from `S' to `E' is now \textbf{completely mapped out}.'' \\
    \midrule
    Qwen3-VL-235B &
    Brief spatial tracing &
    ``The path starts at S and \textbf{follows the line}. \textbf{Let's trace
    it.} First, \textbf{from S}, the line goes to green, then brown, then blue,
    then purple, then orange, then red, then pink, then yellow, \textbf{and then
    to E}. ... Starting at S, the first dot is green, then brown, blue, purple,
    orange, red, pink, yellow.'' \\
    \midrule
    Kimi-K3 &
    Brief spatial tracing &
    ``\textbf{Need inspect image.} ... Image shows black line forming spiral
    from S at inner near right of small loop.
    Dots colored along path: near S \textbf{inner} yellow at top of
    \textbf{inner arc}, orange left \textbf{inner}, green lower-left
    \textbf{inner}, blue bottom \textbf{outer}, purple right \textbf{outer} ...'' \\
    
    \midrule
    GLM-4.6V &
    Brief spatial tracing &
    ``The path starts at S, then \textbf{we trace the line}. Let's list the dots
    \textbf{in order}. \textbf{Starting from S}, the first dot is orange, then
    yellow, then pink, then green, then red, then brown, then blue, then purple.
    ... So the sequence is orange, yellow, pink, green, red, brown, blue,
    purple.'' \\

    \bottomrule
    \end{tabularx}
    \label{tab:brief_tracing_low}
\end{table}

\begin{table*}[!htbp]
\caption{
\small
\textbf{Representative reasoning traces exhibiting global-heuristic substitution.}
Examples illustrate global-heuristic reasoning patterns observed across models. Gemini3-Flash does not expose full reasoning traces, making such patterns less directly observable; its excerpt should therefore be viewed as illustrative rather than fully representative.
}
\centering
\small
\renewcommand{\arraystretch}{1.35}
\begin{tabularx}{\textwidth}{@{} l >{\raggedright\arraybackslash}p{2.8cm} X @{}}
\toprule
\textbf{Model} & \textbf{Substitution Type} & \textbf{Generated Reasoning} \\
\midrule

Sonnet-4.5 &
Ring decomposition &
``There are \textbf{two circles}: an \textbf{outer one} and an
\textbf{inner one}. S is marked on the inner circle near the purple dot.
Tracing from S along the line:
\textbf{purple $\rightarrow$ red $\rightarrow$ pink $\rightarrow$ blue
$\rightarrow$ back to purple}. Since it is a \textbf{circular path},
I should list the sequence of dots until returning to the starting point,
then append `end'.'' \\

\midrule

\multirow{2}{*}{GPT-5.4}
& Ring decomposition
& ``\textbf{I need to document the order of the colored dots:} red, purple,
orange, and yellow \textbf{on the inner loop}; then brown, pink, blue, and green
\textbf{on the outer loop}.'' The two per-ring lists are concatenated into the
final answer. \\

& Angular matching
& ``I am considering loops that start with the inner section at purple
(\textbf{southeast}) and progress to red (\textbf{northeast}), then pink
(\textbf{northwest}), yellow (\textbf{southwest}), orange
(\textbf{southeast}), and blue. \ldots\ This arrangement follows a
counterclockwise spiral and increases in radius, with points at every
\textbf{quadrant} roughly every 90 degrees as it turns.'' \\

\midrule

Gemini3-Flash &
Ring decomposition &
``There are \textbf{two loops}. The \textbf{inner loop} holds orange,
brown, purple, and blue, while the \textbf{outer loop} holds pink,
yellow, green, and red. \ldots\ So the sequence is orange, brown,
purple, blue, then pink, yellow, green, red.'' \\

\midrule

Qwen3-VL-235B &
Ring decomposition &
``\textbf{Inner circle (smallest loop)}: purple (top-left inner),
yellow (bottom inner), brown (right inner) ---
\textbf{Middle circle (next loop)}: red (left middle), blue (bottom middle),
green (right middle) ---
\textbf{Outer circle (E loop)}: orange (top-right outer), pink (top outer).'' \\

\midrule

\multirow{3}{*}{Kimi-K3}
& Coordinate estimation
& ``Dots along it at \textbf{increasing radii and angles}.
\textbf{If we list by radius from center (\textasciitilde325,330)}:
yellow \textbf{(301,338) r\textasciitilde25}; green \textbf{(347,356) r\textasciitilde34};
purple \textbf{(365,295) r\textasciitilde52}; orange \textbf{(289,270) r\textasciitilde68};
red \textbf{(375,395) r\textasciitilde82}; pink \textbf{(408,266) r\textasciitilde105};
blue \textbf{(263,230) r\textasciitilde117}; brown \textbf{(220,391) r\textasciitilde122}.
\textbf{Radii not monotonic perfectly due center estimate}. Sequence by angle maybe.'' \\

& Angular matching
& ``This order (yellow, red, orange, pink, green, brown, blue, purple) is
\textbf{not monotonic angle around center} maybe. Let's \textbf{compute angles
relative to center S (\textasciitilde300,320)}: yellow angle
\textbf{\textasciitilde90\textdegree} (down), red
\textbf{\textasciitilde135\textdegree} (down-right) \ldots\
orange (39,--67) angle \textbf{--60\textdegree}, pink (--46,--42) angle
\textbf{--137\textdegree} or 223, green (--50,67) angle
\textbf{127\textdegree}, brown (67,105) angle \textbf{57\textdegree},
blue (55,--132) angle \textbf{--67\textdegree}, purple (--107,--74) angle
\textbf{--145\textdegree}. \textbf{Not simple}.'' \\

& Ring decomposition
& ``Two black curved paths: \textbf{inner small nearly circular arc} around S,
and \textbf{outer larger arc/circle} with E at upper left. Colored dots placed
on paths: \textbf{Inner circle dots}: brown at upper-left, purple at upper-right,
red at right-lower, orange at bottom near S. \textbf{Outer large circle}:
yellow at top-right, pink at right, blue at bottom, green at left-lower?
Hmm green dot at lower-left \textbf{may lie on outer circle}? \ldots\
It might be on outer circle, \textbf{not inner}.'' \\

\midrule

GLM-4.6V &
Ring decomposition &
``Wait, the spiral: the \textbf{innermost} is S, then the \textbf{first loop}
has orange, blue, pink, yellow? So the first dot after S is orange, then blue,
then pink, then yellow (that's \textbf{the inner loop}), then
\textbf{the next loop}: purple, green, brown, red?'' \\

\bottomrule
\end{tabularx}
\label{tab:substitution_reasoning}
\end{table*}




\clearpage
\section{Prompting Intervention Details}
\label{app:prompting_intervention_details}

\paragraph{Prompt Construction.}
We compare three prompting conditions. \textit{Standard} uses the original tracing prompt without procedural guidance. \textit{Human-instructed} adds a tracing procedure explicitly specified by a human, instructing the model to follow visible line connectivity from S while rejecting choices based on proximity or visual similarity. For \textit{Model-generated}, we provide each model with the same representative \textit{Low}-density Swirl image and ask it not to solve the image, but instead to generate three reusable and meaningfully distinct tracing procedures that could be applied to other instances of the task. Each procedure must consist of explicit numbered steps describing how to trace from S, without assuming any particular color sequence, path shape, number of dots, or answer from the example image. The task statement and final answer prefix are held fixed across all versions, while only the generated instruction and hypothetical example may vary. \Cref{tab:model_generated_instructions} reports only the instruction component of all three versions generated by each model.

\begin{tcolorbox}[
    colback=gray!5,
    colframe=gray!45,
    title=\textbf{Prompt for Generating Model-Specific Tracing Instructions},
    fonttitle=\bfseries,
    breakable
]
\small

You are asked to write explicit instructions for solving a visual line-tracing task.

Your goal is \textbf{NOT} to solve the image shown here. Instead, generate three reusable tracing instructions that you yourself would find effective for reliably solving this type of task.

For each version, the instruction should explicitly describe, in sequential order, the visual tracing procedure that should be followed from the starting point ``S'' along the target line. Each numbered step should describe one concrete action or decision that can be directly followed when solving a new image.

Do not assume any particular color sequence, path shape, number of dots, or answer from the current image. The instructions must generalize to other images of the same task.

Each version must include a simple hypothetical example demonstrating how its tracing procedure should be applied. The example must not use or reveal the solution to the current image.

Generate three meaningfully distinct tracing procedures. Each version should represent a different approach to reliably tracing the target line rather than a superficial rewording of the same procedure.

Keep each instruction concise, but include every procedural step that you consider necessary for reliable tracing.

\vspace{0.5em}
For all three versions:
\begin{itemize}
    \item Copy the Task sentence exactly as provided below.
    \item Copy the final sentence beginning with ``Visually tracing...'' exactly as provided below.
    \item Do not modify, paraphrase, or omit either of these two fixed components.
    \item Only the Instruction and Example should differ across versions.
    \item The instruction must use numbered sequential steps.
    \item Use as many numbered steps as you consider necessary; the number of steps is not fixed.
    \item Do not include any text outside the three specified blocks.
    \item Do not use code fences.
\end{itemize}

\vspace{0.5em}
Return exactly three blocks in the following structure:

\vspace{0.5em}
\texttt{<version\_1>}

\textbf{Task:} Complete the sentence by listing the colors of the dots on the target line in sequential order, followed by ``end''. No explanations are allowed.

\textbf{Instruction:}

1. [First procedural step]

2. [Next procedural step]

3. [Continue with additional numbered steps as necessary]

[$\cdots$]

\textbf{Example:}

[Give one simple hypothetical example demonstrating how this tracing procedure should be applied. The example must not use or reveal the solution to the current image.]

Visually tracing the path of the line from ``S'' in the image, the sequence of colored dots on its path is strictly

\texttt{</version\_1>}

...

\texttt{</version\_3>}

\end{tcolorbox}

\begin{table*}[htbp]
\centering
\caption{
\textbf{Human-instructed tracing procedure.}
The tracing procedure was explicitly specified by a human and provided identically to all evaluated models.
}
\label{tab:human_instructed}
\small
\setlength{\tabcolsep}{6pt}
\renewcommand{\arraystretch}{1.15}

\begin{tabularx}{\textwidth}{
@{}p{0.07\textwidth}>{\raggedright\arraybackslash}X@{}}
\toprule
\textbf{Step} & \textbf{Instruction} \\
\midrule

1 &
Start from the point labeled ``S'' and treat it as the current position. \\

2 &
Select the dot that is visibly connected to the current position by a continuous line; use this line connection as the sole criterion for selection. \\

3 &
Do not select a dot based on proximity or visual similarity to the path. For example, do not select a dot simply because it is close to the current position, or because it lies on a nearby parallel segment. \\

4 &
Move to the selected dot, treat it as the new current position, and repeat until the path ends. \\

\bottomrule
\end{tabularx}
\end{table*}

\begin{table*}[htbp]
\centering
\caption{
\textbf{Model-generated tracing instructions.}
Three reusable tracing procedures generated independently by each model. Only the generated \textit{Instruction} component is shown; the fixed task statement, hypothetical example, and final answer prefix are omitted.
}
\label{tab:model_generated_instructions}
\small
\setlength{\tabcolsep}{5pt}
\renewcommand{\arraystretch}{1.12}
\begin{tabularx}{\textwidth}{
@{}p{0.13\textwidth}p{0.06\textwidth}>{\raggedright\arraybackslash}X@{}}
\toprule
\textbf{Model} & \textbf{Ver.} & \textbf{Generated Instruction} \\
\midrule

\multirow{3}{*}{Sonnet-4.5}
& V1 &
1.~Locate the starting point labeled 'S' in the image. \\
&& 2.~Identify the line segment that begins directly at or connects to 'S'. \\
&& 3.~Follow this line continuously with your eyes in the direction it extends from 'S'. \\
&& 4.~Each time the line passes through or touches a colored dot, record that dot's color in sequence. \\
&& 5.~Continue tracing the line without deviation until it terminates (reaches an endpoint, a label like 'E', or closes back on itself). \\
&& 6.~List all recorded colors in the order encountered, followed by 'end'. \\

\cmidrule(lr){2-3}

& V2 &
1.~Locate the starting point 'S' and identify which line segment is connected to it. \\
&& 2.~From 'S', move along the connected line to the first colored dot you encounter on that line. \\
&& 3.~Record the color of this dot. \\
&& 4.~At each dot, verify that the line continues beyond it by identifying the line segment exiting the dot. \\
&& 5.~Follow the exiting line segment to the next colored dot on the same continuous path. \\
&& 6.~Repeat steps 3--5, ensuring you stay on the same continuous line (ignore any other lines that may be nearby but not connected). \\
&& 7.~Stop when the line reaches a termination point or label (such as 'E') with no further continuation. \\
&& 8.~Output the list of colors in the sequence recorded, followed by 'end'. \\

\cmidrule(lr){2-3}

& V3 &
1.~Find the point marked 'S' in the image. \\
&& 2.~Observe the direction of the line emanating from 'S' and identify the next dot that lies directly on this line path. \\
&& 3.~Record the color of this next dot. \\
&& 4.~Position yourself mentally at this dot and determine which direction the line continues from here. \\
&& 5.~Look ahead along the continuing line to find the next dot on the path. \\
&& 6.~Record this new dot's color and repeat steps 4--5. \\
&& 7.~Continue this forward progression until no more dots are encountered on the line or the line reaches a terminal point. \\
&& 8.~State the complete sequence of colors in order, concluding with 'end'. \\

\bottomrule
\end{tabularx}
\end{table*}

\begin{table*}[htbp]
\ContinuedFloat
\caption{
\textbf{Model-generated tracing instructions (continued).}
}
\small
\setlength{\tabcolsep}{5pt}
\renewcommand{\arraystretch}{1.12}
\begin{tabularx}{\textwidth}{
@{}p{0.13\textwidth}p{0.06\textwidth}>{\raggedright\arraybackslash}X@{}}
\toprule
\textbf{Model} & \textbf{Ver.} & \textbf{Generated Instruction} \\
\midrule

\multirow{3}{*}{GPT-5.4}
& V1 &
1.~Locate the letter 'S' and identify the single line that touches it. \\
&& 2.~Start at 'S' and trace that exact line continuously in the direction it extends, without switching to any nearby path. \\
&& 3.~Each time the traced line passes through a colored dot, write down that dot's color immediately. \\
&& 4.~At curves, loops, or close approaches, stay on the same unbroken stroke; ignore dots on lines that are adjacent but not connected to your traced path. \\
&& 5.~Continue tracing until the line clearly terminates. \\
&& 6.~Output the recorded colors in the order encountered, then add end. \\

\cmidrule(lr){2-3}

& V2 &
1.~Find 'S' and note the first visible direction the target line leaves it. \\
&& 2.~Follow the line only a short segment at a time, always asking which stroke is the direct continuation of the one you were just on. \\
&& 3.~When you reach a colored dot that lies on that stroke, add its color to your list. \\
&& 4.~After each dot, resume tracing from just beyond that dot along the same line. \\
&& 5.~If the path passes near another part of the drawing, ignore proximity and follow continuity of the stroke itself. \\
&& 6.~Repeat until no more line remains. \\
&& 7.~Give only the color sequence you collected, followed by end. \\

\cmidrule(lr){2-3}

& V3 &
1.~Begin at 'S' and mentally mark your current position on the line. \\
&& 2.~Trace forward until you hit the next colored dot on that same line; record its color as the next checkpoint. \\
&& 3.~From that dot, continue to the next dot on the same connected path and record it. \\
&& 4.~Keep using each found dot as the new checkpoint so you never skip ahead or jump across gaps. \\
&& 5.~If the line bends around or passes close to earlier sections, confirm that the path remains physically connected before recording any dot. \\
&& 6.~Stop when the checkpoint-to-checkpoint tracing reaches the line's endpoint. \\
&& 7.~Return the recorded colors in order, followed by end. \\

\bottomrule
\end{tabularx}
\end{table*}

\begin{table*}[p]
\ContinuedFloat
\caption{
\textbf{Model-generated tracing instructions (continued).}
}
\small
\setlength{\tabcolsep}{5pt}
\renewcommand{\arraystretch}{1.12}
\begin{tabularx}{\textwidth}{
@{}p{0.13\textwidth}p{0.06\textwidth}>{\raggedright\arraybackslash}X@{}}
\toprule
\textbf{Model} & \textbf{Ver.} & \textbf{Generated Instruction} \\
\midrule

\multirow{3}{*}{Gemini3-Flash}
& V1 &
1.~Locate the 'S' marker which indicates the start of the line. \\
&& 2.~Identify the specific line segment that is physically connected to the 'S' marker. \\
&& 3.~Move your focus along this line away from the start point. \\
&& 4.~When you encounter a colored dot directly on the line, note its color. \\
&& 5.~Continue following the same line path, ensuring you do not cross over to adjacent or intersecting lines. \\
&& 6.~Record the color of every dot in the exact order you pass them until you reach the 'E' marker. \\

\cmidrule(lr){2-3}

& V2 &
1.~Position your gaze at the 'S' symbol at the center of the image. \\
&& 2.~Follow the continuous path of the line as it spirals outward from the center. \\
&& 3.~Identify every colored circle that sits directly on the line you are tracing. \\
&& 4.~Keep track of the colors in a list according to their distance from the 'S' marker along the path. \\
&& 5.~Be careful at points where the line loops close to itself; only record dots that are on the specific path originating from 'S'. \\
&& 6.~Stop the process once the path reaches the 'E' terminal marker. \\

\cmidrule(lr){2-3}

& V3 &
1.~Identify the 'S' start point and the 'E' end point on the diagram. \\
&& 2.~Trace the single, unbroken line that connects these two points, starting your journey at 'S'. \\
&& 3.~Scan forward along the line and list the color of the first dot you pass. \\
&& 4.~Continue scanning forward along the line's specific trajectory, listing each new dot you encounter. \\
&& 5.~Ensure you do not skip any dots or include dots from other parts of the spiral. \\
&& 6.~Once the path ends at the 'E' marker, finalize your list. \\

\bottomrule
\end{tabularx}
\end{table*}

\begin{table*}[htbp]
\ContinuedFloat
\caption{
\textbf{Model-generated tracing instructions (continued).}
}
\small
\setlength{\tabcolsep}{5pt}
\renewcommand{\arraystretch}{1.12}
\begin{tabularx}{\textwidth}{
@{}p{0.13\textwidth}p{0.06\textwidth}>{\raggedright\arraybackslash}X@{}}
\toprule
\textbf{Model} & \textbf{Ver.} & \textbf{Generated Instruction} \\
\midrule

\multirow{3}{*}{Kimi-K3}
& V1 &
1.~Locate 'S' and identify the single visible line segment that leaves it; ignore all other labels and nearby dots. \\
&& 2.~Place an imaginary tracer at 'S' and move only along the center of that stroke in the direction the stroke leaves 'S'. \\
&& 3.~Continue smoothly along curves, keeping the tracer on the same continuous stroke and never jumping across blank space. \\
&& 4.~Whenever the stroke passes through or touches a dot, append that dot's color to the sequence in the order encountered. \\
&& 5.~At any crossing or junction, choose the branch with the smallest change in direction that keeps the same stroke width/style; do not turn onto a merely touching or crossing line. \\
&& 6.~If a chosen branch dead-ends, backtrack to the last junction and take the next smoothest untried continuation. \\
&& 7.~Stop when the stroke ends, reaches a labeled endpoint such as 'E', returns to the start with no new dots, or no untraced continuation remains. \\
&& 8.~Output the recorded colors separated by spaces, then 'end'. \\

\cmidrule(lr){2-3}

& V2 &
1.~Mark 'S' as the current position and estimate the outgoing tangent of the target line at 'S'. \\
&& 2.~Advance a short fixed step along that tangent; after each step, re-estimate the local tangent from the most recent stroke direction. \\
&& 3.~Use a look-ahead window: before crossing any dot or junction, confirm the next piece of stroke continues the same tangent, thickness, and spacing pattern. \\
&& 4.~Treat a dot as on-path only if the stroke center intersects the dot or clearly connects to its edge along the current direction; record its color once. \\
&& 5.~At a junction, rank candidate exits by tangent alignment and continuity, try the best-ranked exit, and mentally mark it as used. \\
&& 6.~If the trail becomes ambiguous, compare the immediate backward direction and reject any candidate that requires a sharp reversal unless the visible stroke truly makes that U-turn. \\
&& 7.~Repeat step-advance, dot recording, and junction ranking until an endpoint label, open end, completed loop, or exhausted stroke is reached. \\
&& 8.~Read the recorded list from first to last and finish with 'end'. \\

\cmidrule(lr){2-3}

& V3 &
1.~Convert the picture mentally into a graph: nodes are 'S', endpoints/labels, junctions, and dots; edges are stroke pieces between adjacent nodes. \\
&& 2.~Keep only edges belonging to the target stroke connected to 'S'; discard isolated decorations, labels, and crossing strokes that do not share a continuous junction. \\
&& 3.~Starting at the 'S' node, choose the incident edge that is the actual outgoing stroke and mark it visited. \\
&& 4.~Traverse from node to node across unvisited connected edges; when a node is a dot, append its color to the sequence. \\
&& 5.~At a junction node with multiple unvisited edges, select the edge that continues the same stroke with the least curvature change; mark rejected edges so they are not revisited unless backtracking is required. \\
&& 6.~If traversal reaches a dead end with unvisited edges elsewhere, backtrack along visited edges to the nearest junction with an untried plausible edge and resume. \\
&& 7.~For loops, traverse the cycle once from the entry direction and do not recount dots when the path closes or repeats a node. \\
&& 8.~Stop after all reachable target-stroke edges from 'S' have been considered or an explicit end label is reached; then report colors in visit order followed by 'end'. \\

\bottomrule
\end{tabularx}
\end{table*}

\begin{table*}[htbp]
\ContinuedFloat
\caption{
\textbf{Model-generated tracing instructions (continued).}
}
\small
\setlength{\tabcolsep}{5pt}
\renewcommand{\arraystretch}{1.12}
\begin{tabularx}{\textwidth}{
@{}p{0.13\textwidth}p{0.06\textwidth}>{\raggedright\arraybackslash}X@{}}
\toprule
\textbf{Model} & \textbf{Ver.} & \textbf{Generated Instruction} \\
\midrule

\multirow{3}{*}{GLM-4.6V}
& V1 &
1.~Locate the starting point marked 'S' in the image. \\
&& 2.~Begin tracing the line starting from 'S', moving in the direction the line extends from this point. \\
&& 3.~As you trace, identify and record the color of each dot you encounter in the exact order they appear along the line. \\
&& 4.~Continue tracing until you reach the end of the visible line path. \\
&& 5.~After tracing the complete path, list the recorded colors in sequence, followed by 'end'. \\

\cmidrule(lr){2-3}

& V2 &
1.~Identify the starting point 'S' and the initial direction of the line extending from it. \\
&& 2.~Trace the first segment of the line from 'S' until you reach the first colored dot. \\
&& 3.~Record the color of this first dot. \\
&& 4.~Continue tracing from this dot to the next colored dot along the line, recording its color. \\
&& 5.~Repeat step 4 for each subsequent dot encountered along the line path. \\
&& 6.~Stop when no more dots appear on the line and list all recorded colors in order, followed by 'end'. \\

\cmidrule(lr){2-3}

& V3 &
1.~Find the starting point 'S' and note the immediate direction the line takes from this point. \\
&& 2.~Follow the line continuously, mentally marking each time you pass a colored dot. \\
&& 3.~Create a list in your mind of the colors of dots as you encounter them along the tracing path. \\
&& 4.~After completing the full trace from 'S' to the end of the line, verify your mental list by retracing quickly. \\
&& 5.~Convert your verified list of colors into the required format, ending with 'end'. \\

\bottomrule
\end{tabularx}
\end{table*}

\begin{table*}[htbp]
\ContinuedFloat
\caption{
\textbf{Model-generated tracing instructions (continued).}
}
\small
\setlength{\tabcolsep}{5pt}
\renewcommand{\arraystretch}{1.12}
\begin{tabularx}{\textwidth}{
@{}p{0.13\textwidth}p{0.06\textwidth}>{\raggedright\arraybackslash}X@{}}
\toprule
\textbf{Model} & \textbf{Ver.} & \textbf{Generated Instruction} \\
\midrule

\multirow{3}{*}{Qwen3-VL-235B}
& V1 &
1.~Locate the label 'S' marking the starting point on the line. \\
&& 2.~Begin moving along the line in the direction it initially curves away from 'S'. \\
&& 3.~Note the color of each dot encountered in the order they appear along the path. \\
&& 4.~Continue moving forward along the line without reversing direction until all dots on the path are recorded. \\
&& 5.~List the colors in the recorded order, followed by 'end'. \\

\cmidrule(lr){2-3}

& V2 &
1.~Identify 'S' as the starting position on the line. \\
&& 2.~Determine the overall direction of the line's path (e.g., clockwise spiral or counterclockwise spiral) from 'S'. \\
&& 3.~Follow the line consistently in that direction, noting each colored dot in sequence. \\
&& 4.~If the line loops inward or outward, maintain the established direction to avoid missing dots. \\
&& 5.~Stop when the line terminates or no more dots remain on the path. \\

\cmidrule(lr){2-3}

& V3 &
1.~Start at 'S' and visually follow the line to the nearest adjacent dot. \\
&& 2.~Record the color of this first dot. \\
&& 3.~From the current dot, proceed along the line to the next nearest dot, ensuring the line segment connects directly without crossing other segments. \\
&& 4.~Repeat step 3 for each subsequent dot until the end of the line is reached. \\
&& 5.~Combine all recorded colors in order, appending 'end' at the conclusion. \\

\bottomrule
\end{tabularx}
\end{table*}

\clearpage


\subsection{Prompt Engineering Does Not Stabilize Reasoning}
\label{app:prompting_reasoning_stability}

We examine whether explicit tracing instructions stabilize both tracing performance and the reasoning process as density increases. If prompt engineering successfully induced a more robust tracing procedure, we would expect not only improved accuracy but also more stable performance and reasoning cost across density levels.

\paragraph{Explicit Instructions Do Not Stabilize Performance or Reasoning.}
As shown in Table~\ref{tab:prompting_full_accuracy}, neither human- nor model-generated instructions consistently improve tracing accuracy or stabilize performance from \textit{Low} to \textit{High} density. This instability is also reflected in the reasoning process: reasoning length and self-correction frequency both continue to increase with density across models and prompt conditions (\Cref{tab:reasoning_cost_selfcorrection_app}). For example, GPT-5.4 increases from 617 to 4,178 tokens under the \textit{Human-instructed} prompt and from 716 to 4,031 under \textit{Model-generated} prompts, while Qwen3-VL-235B rises from 13.30 to 37.55 self-corrections per sample under the \textit{Human-instructed} prompt. In several cases, prompting therefore fails to suppress, and can even amplify, the increase in reasoning cost and repeated reconsideration as density rises.

\begin{table*}[htbp]
\centering
\caption{
\textbf{Full accuracy results for the prompting intervention.}
Tracing accuracy (\%) under the \textit{Standard}, \textit{Human-instructed}, and three \textit{Model-generated} prompting conditions across Swirl density levels. Each condition contains 200 samples per density level.
}
\label{tab:prompting_full_accuracy}
\small
\setlength{\tabcolsep}{7pt}
\renewcommand{\arraystretch}{1.12}

\begin{tabular}{llccc}
\toprule
\textbf{Model} & \textbf{Prompt} & \textbf{Low} & \textbf{Moderate} & \textbf{High} \\
\midrule

\multirow{5}{*}{Sonnet-4.5}
& Standard             & 65.5 & 9.5  & 0.0 \\
& Human-instructed     & 55.0 & 7.0  & 0.0 \\
& Model-generated V1   & 51.5 & 3.0  & 0.5 \\
& Model-generated V2   & 60.0 & 8.0  & 0.0 \\
& Model-generated V3   & 32.5 & 2.5  & 0.0 \\
\midrule

\multirow{5}{*}{GPT-5.4}
& Standard             & 99.5 & 95.0 & 72.5 \\
& Human-instructed     & 94.0 & 91.5 & 71.0 \\
& Model-generated V1   & 81.5 & 66.5 & 51.0 \\
& Model-generated V2   & 98.0 & 95.0 & 75.0 \\
& Model-generated V3   & 93.5 & 89.5 & 74.0 \\
\midrule

\multirow{5}{*}{Gemini3-Flash}
& Standard             & 100.0 & 97.5 & 68.0 \\
& Human-instructed     & 91.0  & 90.5 & 61.5 \\
& Model-generated V1   & 99.0  & 97.5 & 67.0 \\
& Model-generated V2   & 99.0  & 96.5 & 74.0 \\
& Model-generated V3   & 94.5  & 93.5 & 73.0 \\
\midrule

\multirow{5}{*}{Kimi-K3}
& Standard             & 87.5 & 67.0 & 19.5 \\
& Human-instructed     & 96.5 & 80.5 & 29.0 \\
& Model-generated V1   & 91.5 & 66.5 & 23.0 \\
& Model-generated V2   & 96.5 & 76.5 & 21.5 \\
& Model-generated V3   & 88.0 & 61.0 & 16.0 \\
\midrule

\multirow{5}{*}{Qwen3-VL-235B}
& Standard             & 81.5 & 14.5 & 1.0 \\
& Human-instructed     & 75.5 & 11.0 & 1.0 \\
& Model-generated V1   & 84.0 & 11.5 & 1.5 \\
& Model-generated V2   & 74.5 & 14.5 & 1.0 \\
& Model-generated V3   & 78.5 & 12.0 & 1.5 \\
\midrule

\multirow{5}{*}{GLM-4.6V}
& Standard             & 84.0 & 12.5 & 0.0 \\
& Human-instructed     & 77.5 & 8.5  & 0.0 \\
& Model-generated V1   & 81.5 & 7.5  & 0.0 \\
& Model-generated V2   & 77.0 & 10.0 & 0.5 \\
& Model-generated V3   & 82.0 & 10.5 & 0.0 \\

\bottomrule
\end{tabular}
\end{table*}

\begin{table*}[t!]
\centering
\small
\caption{
\small
\textbf{Reasoning length and self-correction frequency across Swirl difficulty levels and prompt conditions.}
Average reasoning-token usage and number of self-correction markers per sample are reported for \textit{Low}, \textit{Moderate}, and \textit{High} density conditions.
\textit{Standard} uses no tracing instruction; \textit{Human-instructed} uses a single hand-written instruction; \textit{Model-generated} averages the three instructions each model wrote for itself.
}
\label{tab:reasoning_cost_selfcorrection_app}
\small
\setlength{\tabcolsep}{5pt}
\renewcommand{\arraystretch}{1.15}
\begin{tabular}{llcccccc}
\toprule
& & \multicolumn{3}{c}{\textbf{Reasoning tokens}} & \multicolumn{3}{c}{\textbf{Self-corrections / sample}} \\
\cmidrule(lr){3-5}
\cmidrule(lr){6-8}
\textbf{Reasoning Model} & \textbf{Prompt} & \textbf{Low} & \textbf{Moderate} & \textbf{High} & \textbf{Low} & \textbf{Moderate} & \textbf{High} \\
\midrule
\multirow{3}{*}{Sonnet-4.5}
& Standard         & 390 & 510 & 537 & 4.20 & 6.14 & 6.39 \\
& Human-instructed & 448 & 521 & 534 & 4.44 & 6.15 & 6.58 \\
& Model-generated  & 477 & 544 & 525 & 4.99 & 6.17 & 6.16 \\
\midrule
\multirow{3}{*}{GPT-5.4}
& Standard         & 521 & 1,392 & 3,460 & 0.12 & 0.54 & 1.50 \\
& Human-instructed & 617 & 1,659 & 4,178 & 0.27 & 0.73 & 1.84 \\
& Model-generated  & 716 & 1,812 & 4,031 & 0.33 & 0.84 & 1.81 \\
\midrule
\multirow{3}{*}{Gemini3-Flash}
& Standard         & 557 & 800 & 1,337 & 0.47 & 0.88 & 1.66 \\
& Human-instructed & 878 & 1,238 & 2,463 & 0.56 & 1.14 & 3.00 \\
& Model-generated  & 452 & 567 & 1,087 & 0.05 & 0.18 & 0.88 \\
\midrule
\multirow{3}{*}{GLM-4.6V}
& Standard         & 477 & 706 & 854 & 9.25 & 15.64 & 19.49 \\
& Human-instructed & 690 & 1,111 & 1,434 & 11.20 & 22.79 & 30.98 \\
& Model-generated  & 607 & 909 & 1,091 & 11.64 & 20.55 & 25.89 \\
\midrule
\multirow{3}{*}{Kimi-K3}
& Standard         & 2,565 & 2,984 & 4,631 & 9.44 & 10.07 & 16.48 \\
& Human-instructed & 1,486 & 2,612 & 4,512 & 3.81 & 6.55 & 10.29 \\
& Model-generated  & 1,830 & 3,093 & 4,498 & 6.01 & 8.80 & 10.70 \\
\midrule
\multirow{3}{*}{Qwen3-VL-235B}
& Standard         & 657 & 997 & 1,310 & 10.23 & 16.70 & 24.23 \\
& Human-instructed & 867 & 1,612 & 2,119 & 13.30 & 29.34 & 37.55 \\
& Model-generated  & 959 & 1,610 & 2,034 & 15.37 & 27.76 & 36.90 \\
\bottomrule
\end{tabular}
\end{table*}

\clearpage

\paragraph{Explicit Instructions Do Not Stabilize Reasoning Strategies.}

\begin{figure}[!htbp]
    \centering
    \includegraphics[width=\linewidth]{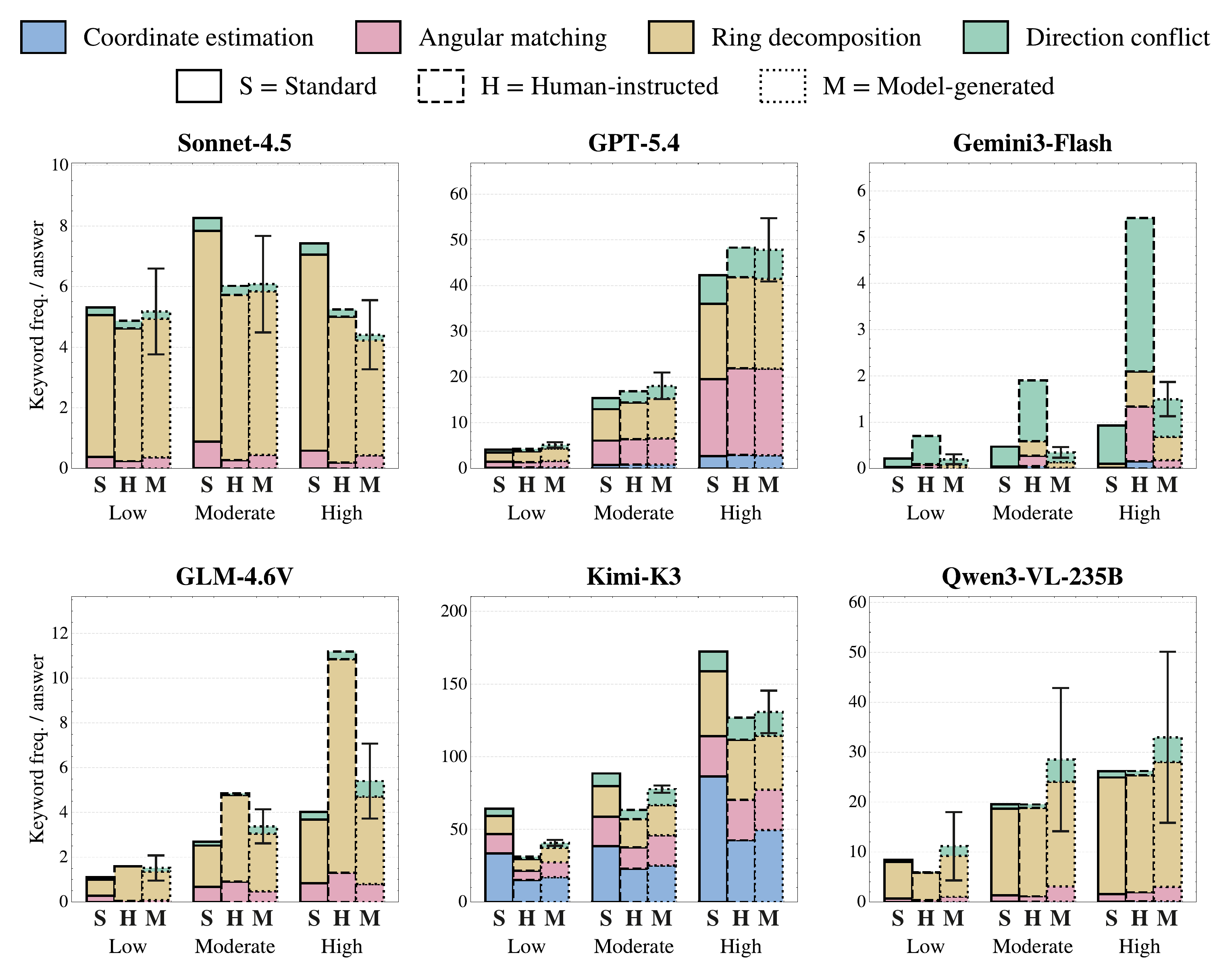}
    \caption{
    \small
    \textbf{Non-tracing strategy intensity across prompting conditions.}
    Keyword-based strategy intensity across \textit{Low}, \textit{Moderate}, and \textit{High} density for \textit{Standard}, \textit{Human-instructed}, and \textit{Model-generated} prompting. \textit{Model-generated} reports the mean across three generated instructions, with error bars showing the standard deviation. The measure is intended as an indirect indicator of reasoning-pattern shifts rather than an absolute measure of strategy use.
    }
    \label{fig:substitution_intensity_conditions}
\end{figure}

Reasoning strategies remain sensitive to density across prompting conditions. As shown in \Cref{fig:substitution_intensity_conditions}, the intensity of non-tracing strategies generally increases from \textit{Low} to \textit{High} density under \textit{Standard}, \textit{Human-instructed}, and \textit{Model-generated} prompting, indicating that explicit tracing instructions do not consistently suppress the shift away from sequential tracing. Strategy categories are identified through keyword-based matching using the mappings in Table~\ref{tab:reasoning_keyword_mapping}; accordingly, this intensity should be interpreted as an indirect indicator of reasoning-pattern change rather than an absolute measure of strategy use.

The qualitative reasoning traces show the same pattern more directly. As shown in Table~\ref{tab:reasoning_pattern_examples_modelgen}, even when models are given their own tracing procedures, \textit{Low}-density reasoning typically advances sequentially from S along connected dots, whereas \textit{High}-density reasoning again shifts toward global reconstruction using rings, angles, radii, or other global criteria rather than line connectivity. Thus, even self-generated tracing instructions do not prevent the strategy shift observed as density increases.

\begin{table*}[t!]
\centering
\caption{
\small
\textbf{Reasoning pattern examples under \textit{model-generated} instructions, at \textit{Low} vs.\ \textit{High} rotation density.}
Excerpts are verbatim from the \textit{Model-generated} condition.
Even when the model is given its own tracing instruction, \textit{Low} density is handled by advancing from S to the next connected dot, whereas \textit{High} density again induces ordering by a global criterion rather than by line connectivity.
}
\label{tab:reasoning_pattern_examples_modelgen}
\footnotesize
\setlength{\tabcolsep}{4pt}
\renewcommand{\arraystretch}{1.25}

\begin{tabular}{@{}lcp{0.76\textwidth}@{}}
\toprule
\textbf{Model} & \textbf{Density} & \textbf{Reasoning excerpt} \\
\midrule

\multirow{2}{*}{Sonnet-4.5}
& Low
& ``From S, there appears to be a line that goes to a purple dot, then \textbf{follows the spiral outward} \ldots\ The line goes to purple dot -- Then to red dot -- Then to green dot\ldots'' \\
& High
& ``Red sits near the \textbf{9--10 o'clock position on the inner ring}, while yellow is around \textbf{5--6 o'clock on that same inner ring}. Moving outward, green appears at roughly \textbf{3 o'clock on the middle ring}, blue at 6--7 o'clock\ldots'' \\
\midrule

\multirow{2}{*}{GPT-5.4}
& Low
& ``if I start from yellow, I can \textbf{follow along line one} to reach red, and then orange. After orange, I wonder if the line continues \ldots\ how these colors \textbf{connect} and what follows next.'' \\
& High
& ``I need to \textbf{assign each dot an angle along the spiral corresponding to their increasing radius} and determine their sequence. \ldots\ purple is roughly at $330^\circ$ and brown around $390^\circ$, both at inner radii; green at about $410^\circ$\ldots'' \\
\midrule

\multirow{2}{*}{Gemini3-Flash}
& Low
& ``\textbf{Finding Start Point}: I've successfully identified the starting point `S' on the spiral. My next step is to \textbf{trace the path} from there.'' (137 characters, single block) \\
& High
& ``\textbf{radial and angular positions} of the outermost points \ldots\ Orange is at \textbf{6 o'clock} and Pink at \textbf{9 o'clock} on the outer loop. \ldots\ I'm now \textbf{working backwards from E}, tracing counter-clockwise from E ($\sim$2 o'clock) past Yellow ($\sim$12 o'clock)\ldots'' \\
\midrule

\multirow{2}{*}{GLM-4.6V}
& Low
& ``locate the `S' point. Then, \textbf{follow the line} from S. The first dot after S is green \ldots\ then brown, then blue, then purple, then orange, then red, then pink, then yellow.'' \\
& High
& ``the \textbf{innermost loop} has S, green \ldots\ then orange (\textbf{on the middle loop}), then yellow (\textbf{middle loop}), then purple (\textbf{outer loop}), then red (\textbf{outer loop}), then pink (\textbf{middle loop}), then blue (middle loop)?'' \\
\midrule

\multirow{2}{*}{Kimi-K3}
& Low
& ``The line starts at S, \textbf{goes to} blue dot, \textbf{then curves down to} green dot, \textbf{then to} yellow dot, then to brown dot, then up to red dot, then orange dot at top\ldots'' \\
& High
& ``pink \textbf{(287,330)} lower-left inner; green \textbf{(309,255)} \ldots\ \textbf{radius $\sim$70}; blue (301,214) top outer \textbf{radius $\sim$108}. \textbf{Colors positions by angle around center (320,320)}: red angle $-80$ r23; yellow r60 angle $-5$; green r65 angle $-105$\ldots'' \\
\midrule

\multirow{2}{*}{Qwen3-VL-235B}
& Low
& ``The \textbf{first adjacent dot from S} is the green one. Then \textbf{from green, the next dot along the line} is purple. Then from purple, the next is brown. Then blue, then pink, then yellow\ldots'' \\
& High
& ``\textbf{Inner circle}: purple at \textbf{3 o'clock}, orange at \textbf{9 o'clock}. \textbf{Counterclockwise order}: purple (3) $\rightarrow$ orange (9). \textbf{Middle circle}: yellow (4) $\rightarrow$ pink (8). \textbf{Outer circle}: brown (2), blue (10), green (8), red (4).'' \\
\bottomrule
\end{tabular}
\end{table*}

\clearpage

\section{Full Results for Richer Visual Ambiguity Settings}
\label{app:real_world}

To assess whether the path-preservation failures observed in our controlled experiments extend to visually richer settings, we evaluate tracing in tangled cables from HANDLOOM~\citep{viswanath2023handloom} and metro maps. These settings introduce crossings, overlaps, clutter, and realistic diagram structure, allowing us to test whether VLMs remain sensitive to nearby competing paths and exhibit the same path-preservation failures beyond simplified synthetic tasks.

\subsection{HANDLOOM Cable Tracing}
\label{app:cable_failure_examples}

\begin{wrapfigure}{r}{0.40\textwidth}
\centering
\vspace{-1.0em}
\includegraphics[width=0.40\textwidth]{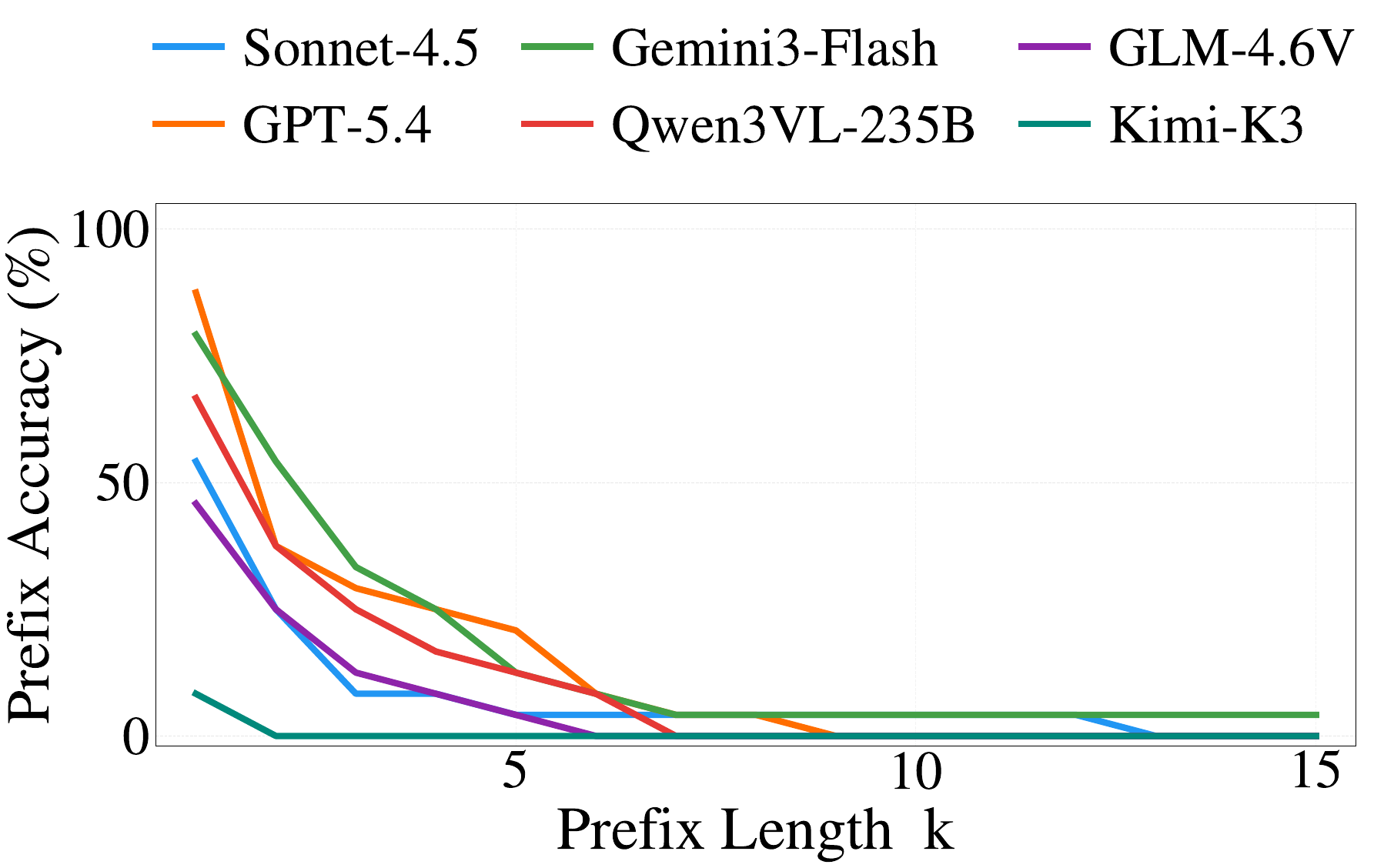}
\vspace{-0.6em}
\caption{
\small
\textbf{Progressive sequence breakdown on HANDLOOM cable tracing.}
Prefix accuracy on 24 cable images, plotted over the first 15 of 30 colored-dot steps.
}
\label{fig:cable_prefix}
\vspace{-1.5em}
\end{wrapfigure}

We evaluate reasoning-enabled models on tangled cable images from HANDLOOM~\citep{viswanath2023handloom}. In this setting, the target cable must be traced through scenes containing self-overlap, crossings, and nearby cable segments. We place 30 colored dots along each target cable, and models are asked to output these dots in endpoint-to-endpoint order.

As shown in \Cref{fig:cable_prefix}, cumulative prefix accuracy drops progressively across sequence steps. Models often begin on the correct cable, but their predictions become less reliable as the traced sequence grows longer and local ambiguities accumulate. This pattern is consistent with a gradual loss of the selected cable rather than a purely formatting-related error.

To separate nearby-cable distraction from explicit intersection handling, we inspect local regions that are away from crossings. For each region, we align the model prediction with the local ground-truth colored-dot sequence. We treat a prediction as corresponding to that region only when it contains at least three consecutive colors that match the ground-truth sequence in either forward or reverse order. We then examine whether the model continues along the target cable or jumps to a nearby cable segment. Red text in \Cref{tab:cable_failure_examples} marks deviations from the ground truth. Across the aligned examples, models often produce a short correct subsequence before leaving the target cable and continuing along an adjacent segment, consistent with the path-switching pattern observed in the controlled tasks. We do not include additional Qwen3-VL-235B examples because no further outputs satisfied this matching criterion in either direction; Kimi-K3 is also excluded because it exceeded the 5,000-token generation limit without producing a final answer.

\vspace{-0.6em}
\begin{table*}[htbp]
\centering
\small
\setlength{\tabcolsep}{6pt}
\renewcommand{\arraystretch}{1.15}
\caption{
\textbf{HANDLOOM cable-tracing failure examples.}
}
\label{tab:cable_failure_examples}

\begin{minipage}[t]{0.33\textwidth}
    \vspace{0.5em}
    \centering
    \includegraphics[width=\linewidth]{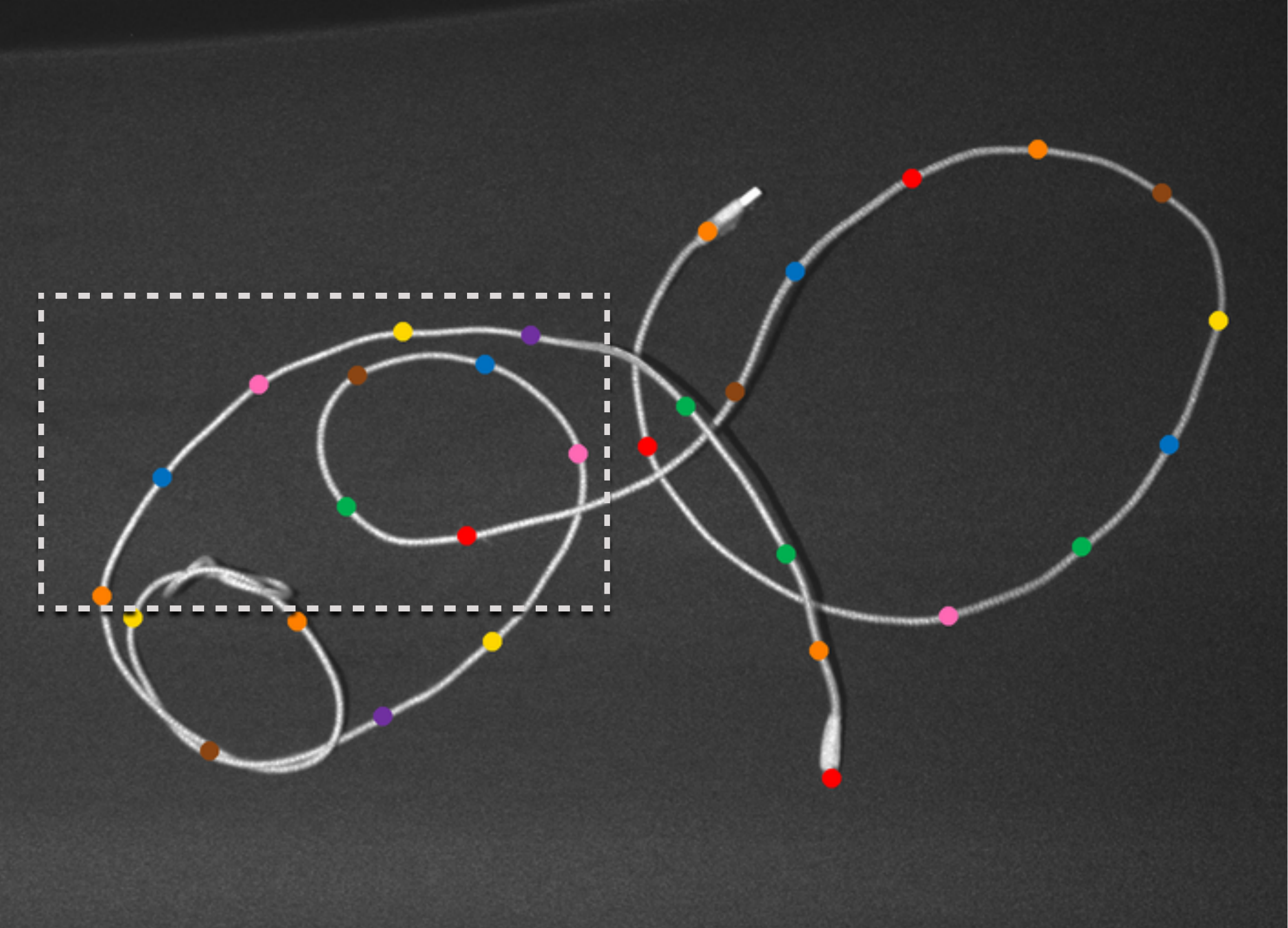}
\end{minipage}
\hfill
\begin{minipage}[t]{0.66\textwidth}
    \vspace{0pt}
    \centering
    \footnotesize
    \begin{tabular}{p{0.22\linewidth} p{0.72\linewidth}}
    \toprule
    \textbf{Source} & \textbf{Colored dot sequence} \\
    \midrule
    \rowcolor{gray!12}
    Ground truth &
    \texttt{... green brown blue pink yellow ...} \\
    Gemini3-Flash &
    \texttt{... green brown blue } \textcolor{red}{\texttt{yellow purple}} \texttt{...} \\
    \midrule
    \rowcolor{gray!12}
    Ground truth &
    \texttt{... orange blue pink yellow purple green ...} \\
    Sonnet-4.5 &
    \texttt{... orange blue pink yellow } \textcolor{red}{\texttt{blue purple}} \texttt{...} \\
    \midrule
    \rowcolor{gray!12}
    Ground truth &
    \texttt{... red green brown pink yellow ...} \\
    GPT-5.4 &
    \texttt{... red green brown} \textcolor{red}{\texttt{blue yellow pink blue}} \texttt{...} \\
    \midrule
    \rowcolor{gray!12}
    Ground truth &
    \texttt{... yellow pink blue brown green red ...} \\
    GLM-4.6V &
    \texttt{... yellow pink blue red } \textcolor{red}{\texttt{yellow purple green}} \\
    \bottomrule
    \end{tabular}
\end{minipage}

\vspace{1.0em}

\begin{minipage}[t]{0.33\textwidth}
    \vspace{0.5em}
    \centering
    \includegraphics[width=\linewidth]{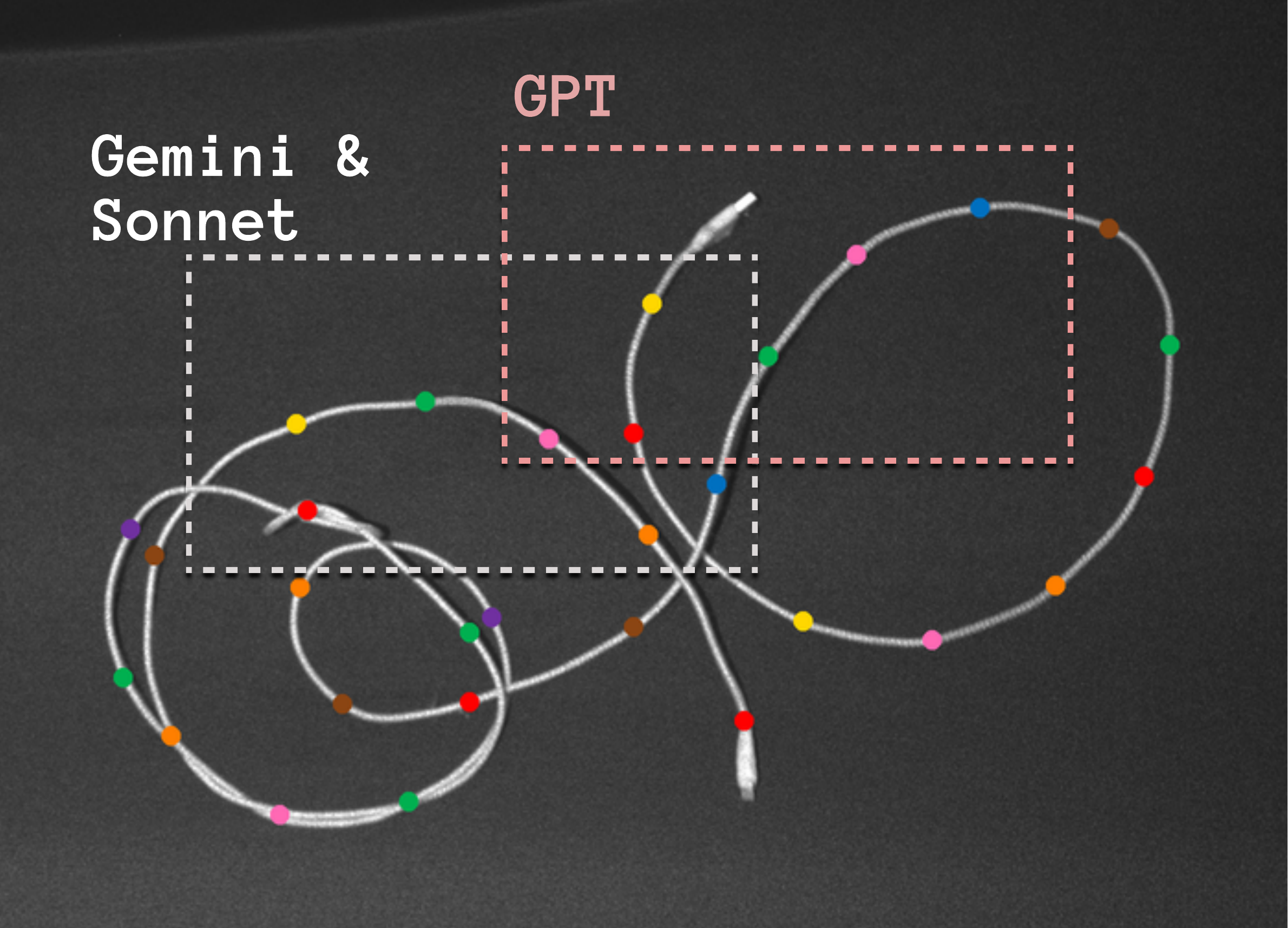}
\end{minipage}
\hfill
\begin{minipage}[t]{0.66\textwidth}
    \vspace{0pt}
    \centering
    \footnotesize
    \begin{tabular}{p{0.22\linewidth} p{0.72\linewidth}}
    \toprule
    \textbf{Source} & \textbf{Colored dot sequence} \\
    \midrule
    \rowcolor{gray!12}
    Ground truth &
    \texttt{... yellow green pink orange red ...} \\
    Gemini3-Flash &
    \texttt{... yellow green pink } \textcolor{red}{\texttt{red blue green}} \texttt{...} \\
    \midrule
    \rowcolor{gray!12}
    Ground truth &
    \texttt{... yellow green pink orange red ...} \\
    Sonnet-4.5 &
    \texttt{... yellow green pink } \textcolor{red}{\texttt{red blue}} \texttt{...} \\
    \midrule
    \rowcolor{gray!12}
    Ground truth &
    \texttt{yellow red yellow pink ...} \\
    GPT-5.4 &
    \texttt{yellow } \textcolor{red}{\texttt{green pink blue}} \texttt{...} \\
    \bottomrule
    \end{tabular}
\end{minipage}

\vspace{1.0em}

\begin{minipage}[t]{0.33\textwidth}
    \vspace{0.5em}
    \centering
    \includegraphics[width=\linewidth]{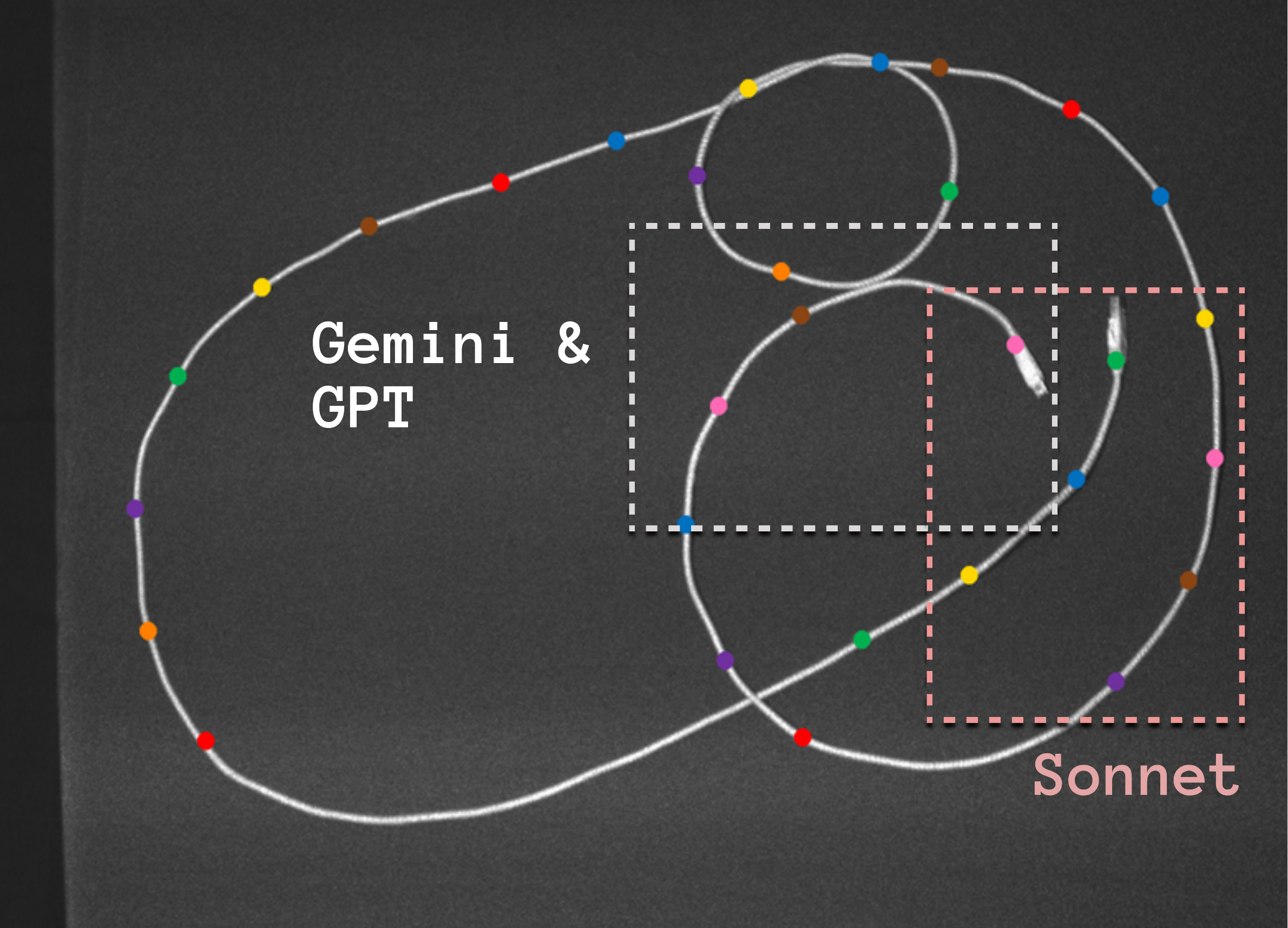}
\end{minipage}
\hfill
\begin{minipage}[t]{0.66\textwidth}
    \vspace{0pt}
    \centering
    \footnotesize
    \begin{tabular}{p{0.22\linewidth} p{0.72\linewidth}}
    \toprule
    \textbf{Source} & \textbf{Colored dot sequence} \\
    \midrule
    \rowcolor{gray!12}
    Ground truth &
    \texttt{... blue pink brown pink blue ...} \\
    Gemini3-Flash &
    \texttt{... blue pink brown} \textcolor{red}{\texttt{orange purple}} \texttt{...} \\
    \midrule
    \rowcolor{gray!12}
    Ground truth &
    \texttt{... green yellow blue green ...} \\
    Sonnet-4.5 &
    \texttt{... green yellow blue} \textcolor{red}{\texttt{pink yellow blue}} \texttt{...} \\
    \midrule
    \rowcolor{gray!12}
    Ground truth &
    \texttt{... yellow purple orange green blue ...} \\
    GPT-5.4 &
    \texttt{... yellow purple orange} \textcolor{red}{\texttt{brown pink blue}} \texttt{...} \\
    \bottomrule
    \end{tabular}
\end{minipage}

\end{table*}

\clearpage

\subsubsection{Cable attention analysis.}

\begin{wrapfigure}{r}{0.50\textwidth}
\vspace{-2.0em}
    \centering
    \vspace{-8pt}
\includegraphics[width=\linewidth]{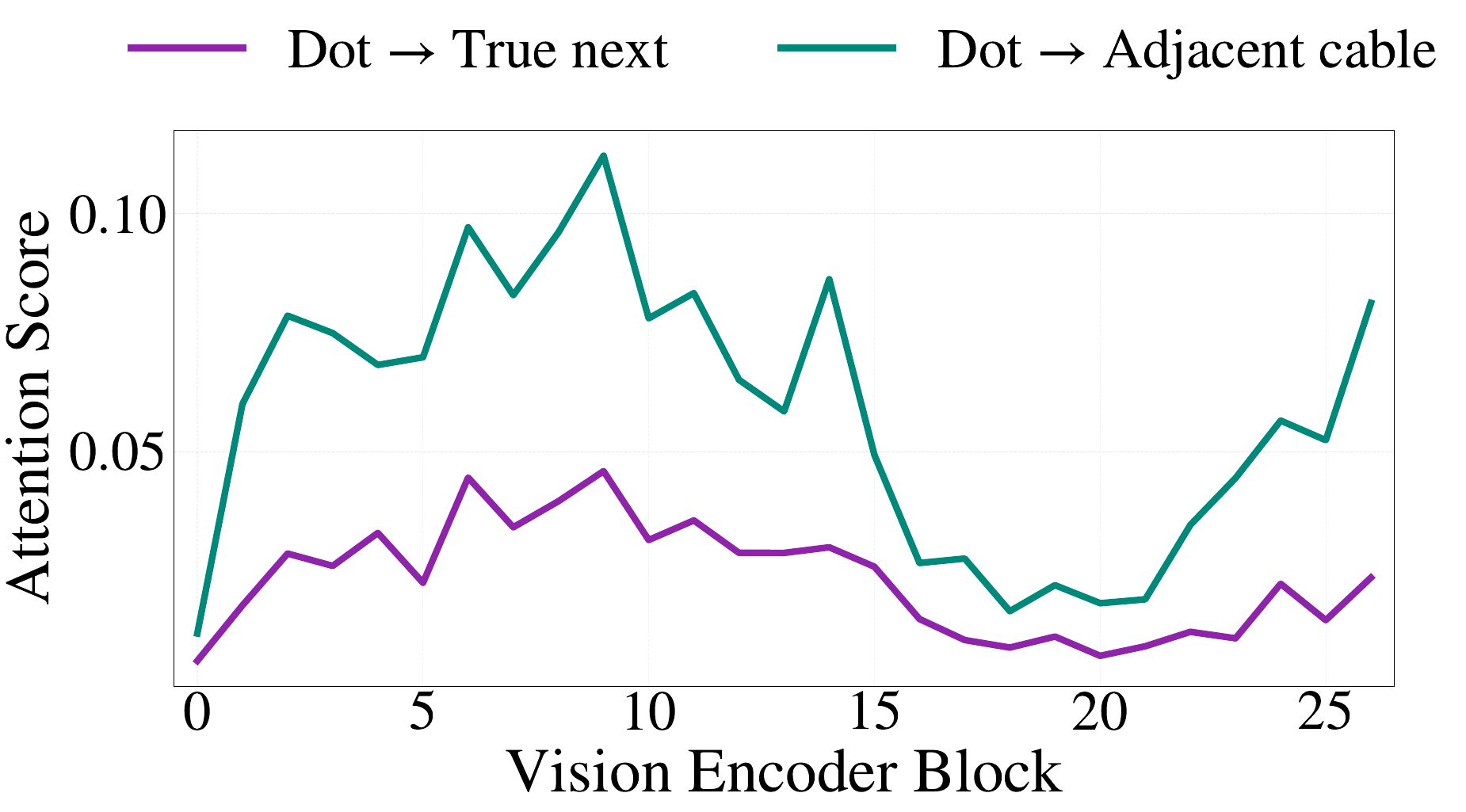}
\caption{
\textbf{Attention to true next dots and nearby cable competitors.}
Qwen3-VL-8B-Thinking vision-encoder attention from the source dot to the true next dot versus a nearby cable competitor, averaged over HANDLOOM regions away from crossings and self-overlaps.
}
\label{fig:cable_attention}
\vspace{-20pt}
\end{wrapfigure}

We apply the same internal-attention analysis described in Appendix~\ref{app:internal} to the tangled-cable setting using Qwen3-VL-8B-Thinking~\citep{qwen3_vl}. To focus on local competition rather than explicit crossing resolution, we sample regions away from crossings and self-overlap points. For each source dot $s$ on the target cable, we manually label nearby dots on adjacent cables based on their spatial proximity to $s$, excluding the immediately preceding and following dots in the ground-truth sequence. Let $t$ denote the true sequential next dot and $\{c_j\}_{j=1}^{N_C}$ the labeled nearby competitors. At each vision-encoder layer $\ell$, we select the three nearby competitors receiving the highest attention from $s$ and measure the attention preference as

\[
\Delta_{\mathrm{attn,cable}}^{(\ell)}
=
a_{s\rightarrow t}^{(\ell)}
-
\frac{1}{3}
\sum_{j \in \mathcal{T}_3^{(\ell)}(s)}
a_{s\rightarrow c_j}^{(\ell)},
\]

where $\mathcal{T}_3^{(\ell)}(s)$ denotes the three adjacent-cable competitors with the highest attention from $s$ at layer $\ell$. Thus, the second term represents the mean attention assigned to the three strongest nearby competing dots.

As shown in \Cref{fig:cable_attention}, attention is frequently distributed toward nearby competitors rather than cleanly favoring the true next dot. This pattern is consistent with the behavioral failures above: even when the local region does not require resolving an explicit crossing, nearby cable structure can compete with the target continuation and weaken path preservation.

\begin{figure*}[htbp]
\centering

\includegraphics[width=\textwidth]{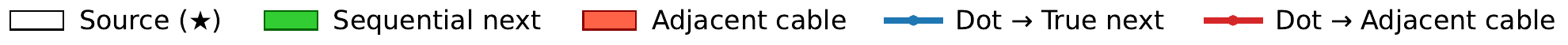}

\vspace{0.4em}

\begin{minipage}[t]{0.32\textwidth}
    \centering
    \includegraphics[width=\linewidth]{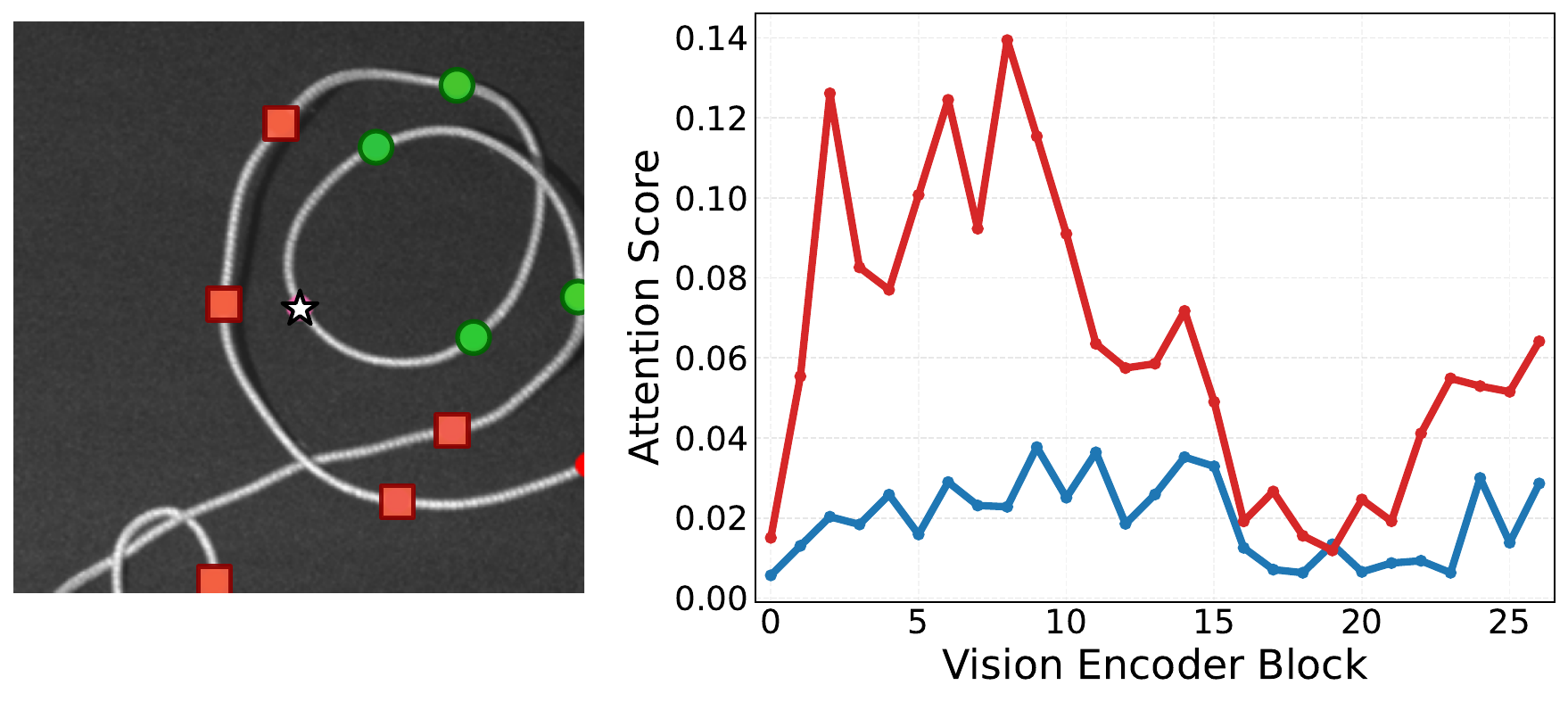}
\end{minipage}
\hfill
\begin{minipage}[t]{0.32\textwidth}
    \centering
    \includegraphics[width=\linewidth]{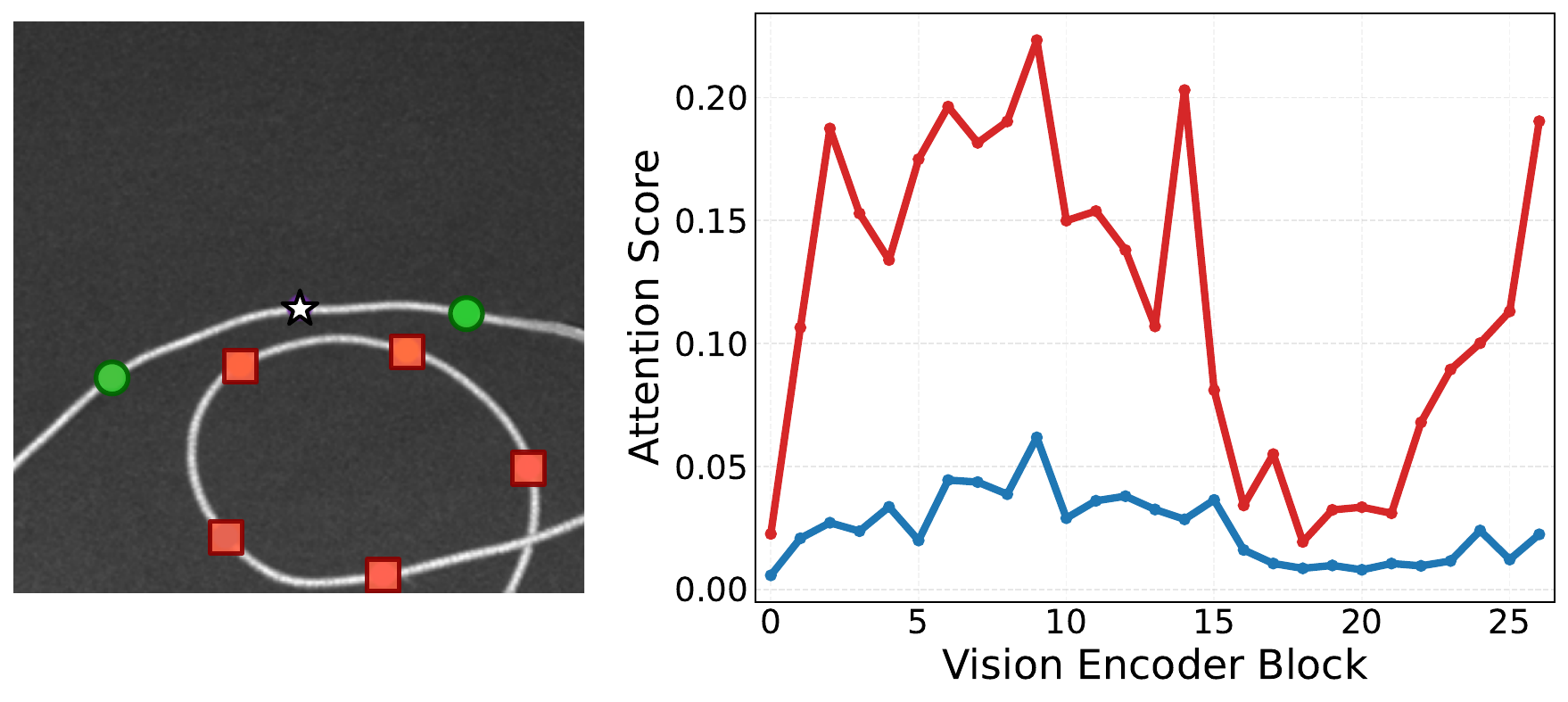}
\end{minipage}
\hfill
\begin{minipage}[t]{0.32\textwidth}
    \centering
    \includegraphics[width=\linewidth]{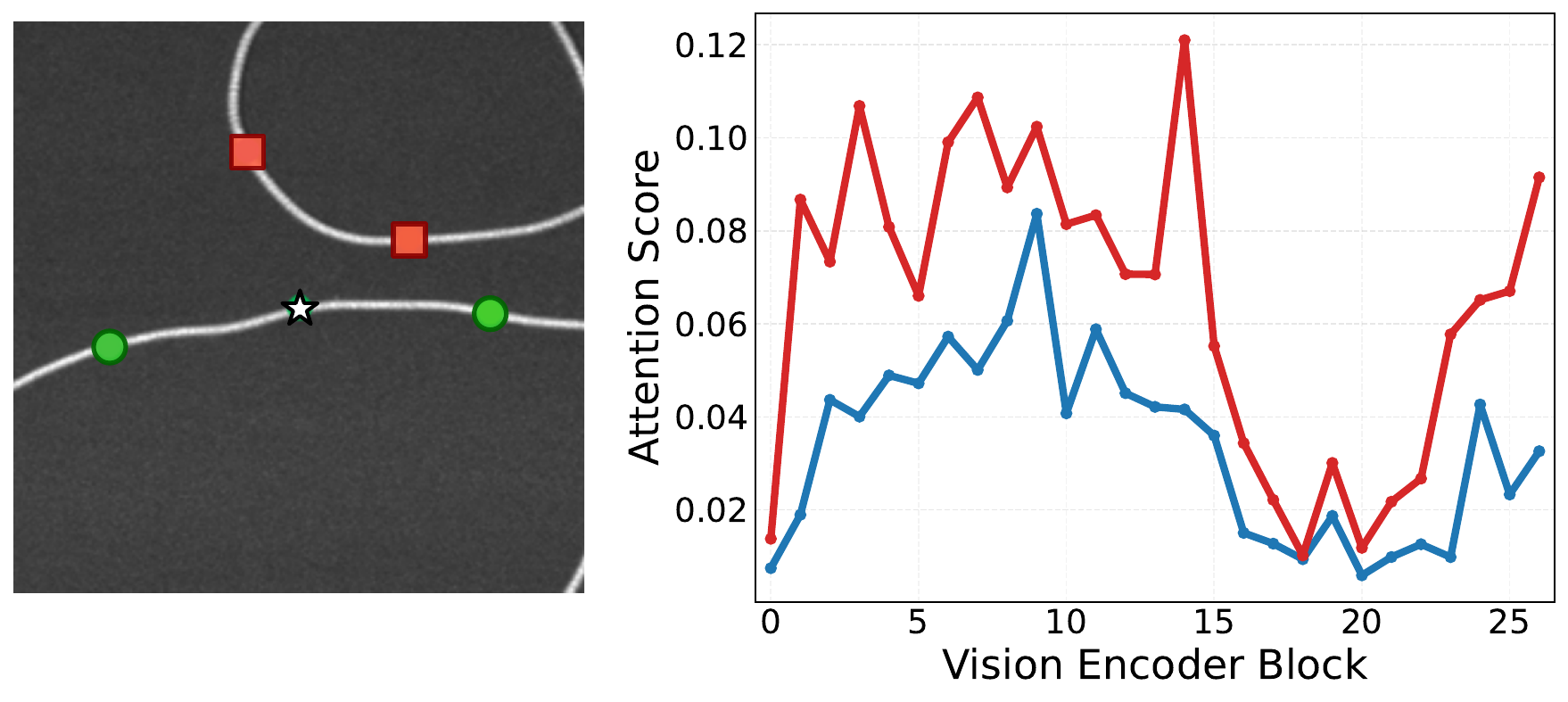}
\end{minipage}

\vspace{0.4em}

\begin{minipage}[t]{0.32\textwidth}
    \centering
    \includegraphics[width=\linewidth]{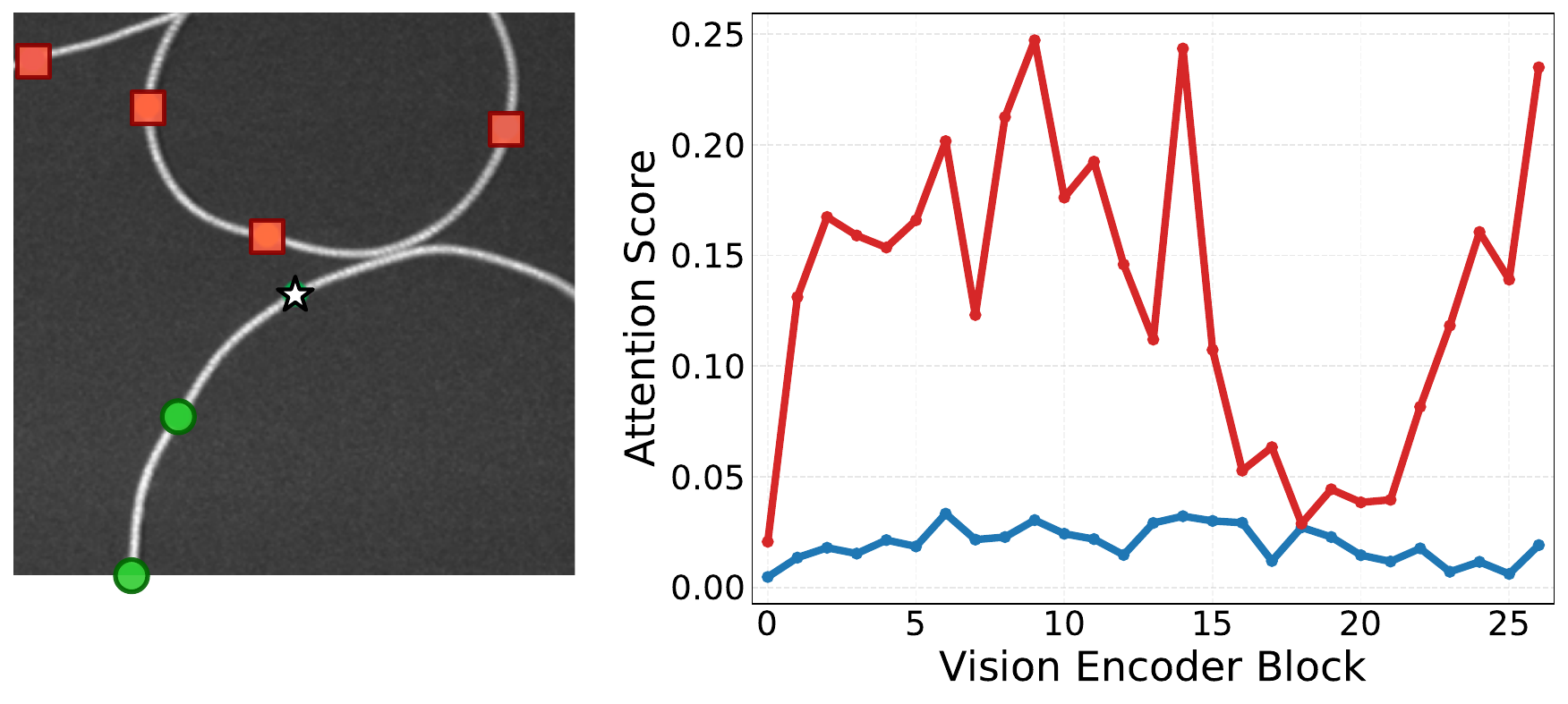}
\end{minipage}
\hfill
\begin{minipage}[t]{0.32\textwidth}
    \centering
    \includegraphics[width=\linewidth]{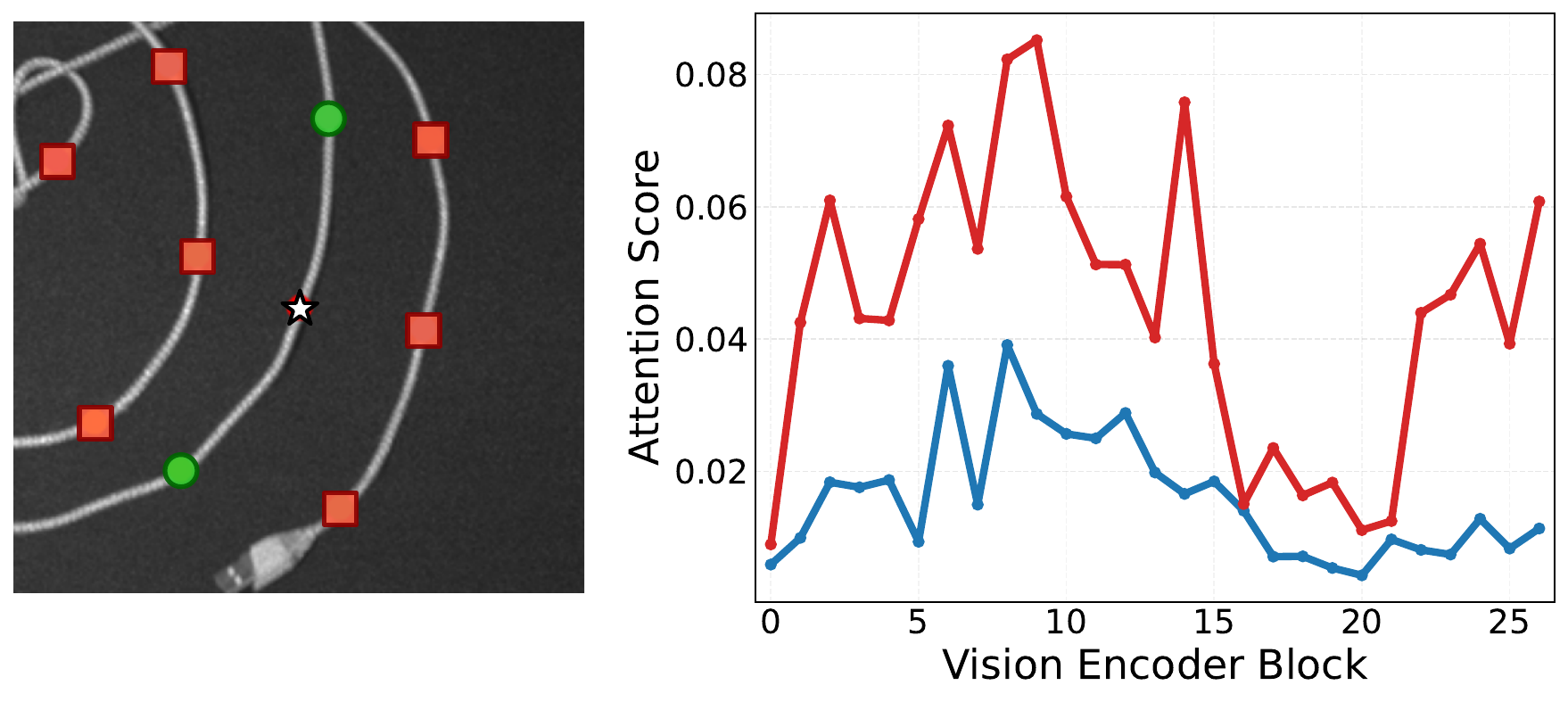}
\end{minipage}
\hfill
\begin{minipage}[t]{0.32\textwidth}
    \centering
    \includegraphics[width=\linewidth]{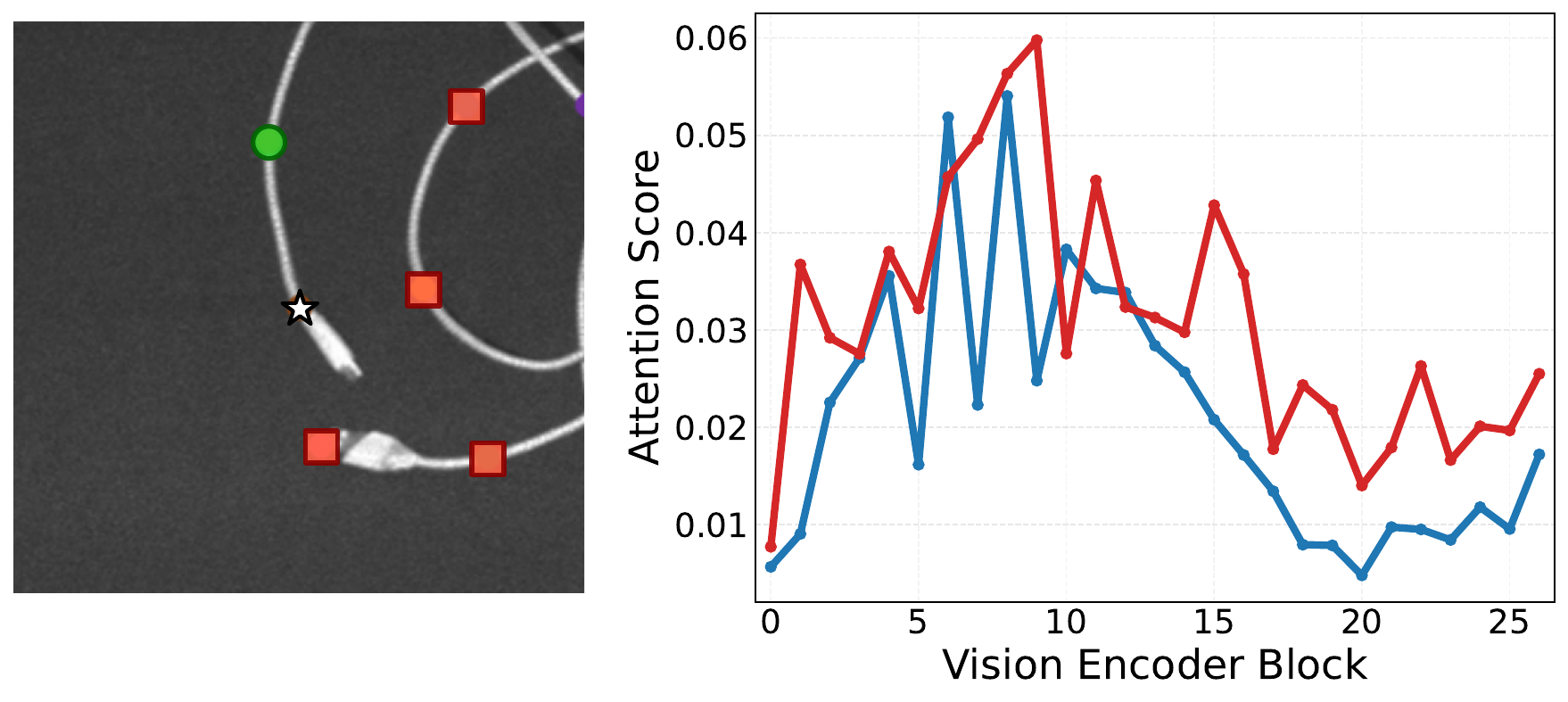}
\end{minipage}
\caption{
\textbf{Cropped HANDLOOM cable attention examples.}
Each panel shows a selected crop and vision-encoder attention from the source dot to the true next dot versus a nearby adjacent-cable competitor.
}
\label{fig:cable_attention_app}
\end{figure*}

\paragraph{Cropped Cable Attention Examples.}
To visualize this effect at the example level, we further inspect cropped HANDLOOM regions. 
Each crop contains a source dot on the currently traced cable, the true sequential next dot, and nearby dots on spatially adjacent cable segments. 
For each selected source dot, we compare attention from the source region to the true next dot with attention from the same source region to the adjacent cable competitor. 
The visualizations in \Cref{fig:cable_attention_app} show the selected crop at early, middle, and late vision-encoder blocks, together with the corresponding block-wise attention scores.

Together, the prefix-accuracy trend, qualitative sequence deviations, average attention analysis, and cropped attention examples show that the HANDLOOM failures are consistent with the path-switching pattern observed in the controlled tasks.

\clearpage
\subsection{Metro Maps}
\label{app:metro}

Following the HANDLOOM cable probe, we also evaluate reasoning-enabled models in a visually richer practical diagrammatic setting: metro maps. Metro-map reading depends directly on the tracing operation studied in this paper: a reader must maintain the identity of a selected route as it bends, runs near other colored lines, shares stations or segments, and competes with dense station labels. We use this setting as an additional practical probe of whether extended reasoning traces maintain stable local continuation under realistic diagram conventions.

\subsubsection{Dataset and Prompt}
\label{app:metro_dataset}

\begin{figure*}[t]
    \centering
    \includegraphics[width=\textwidth]{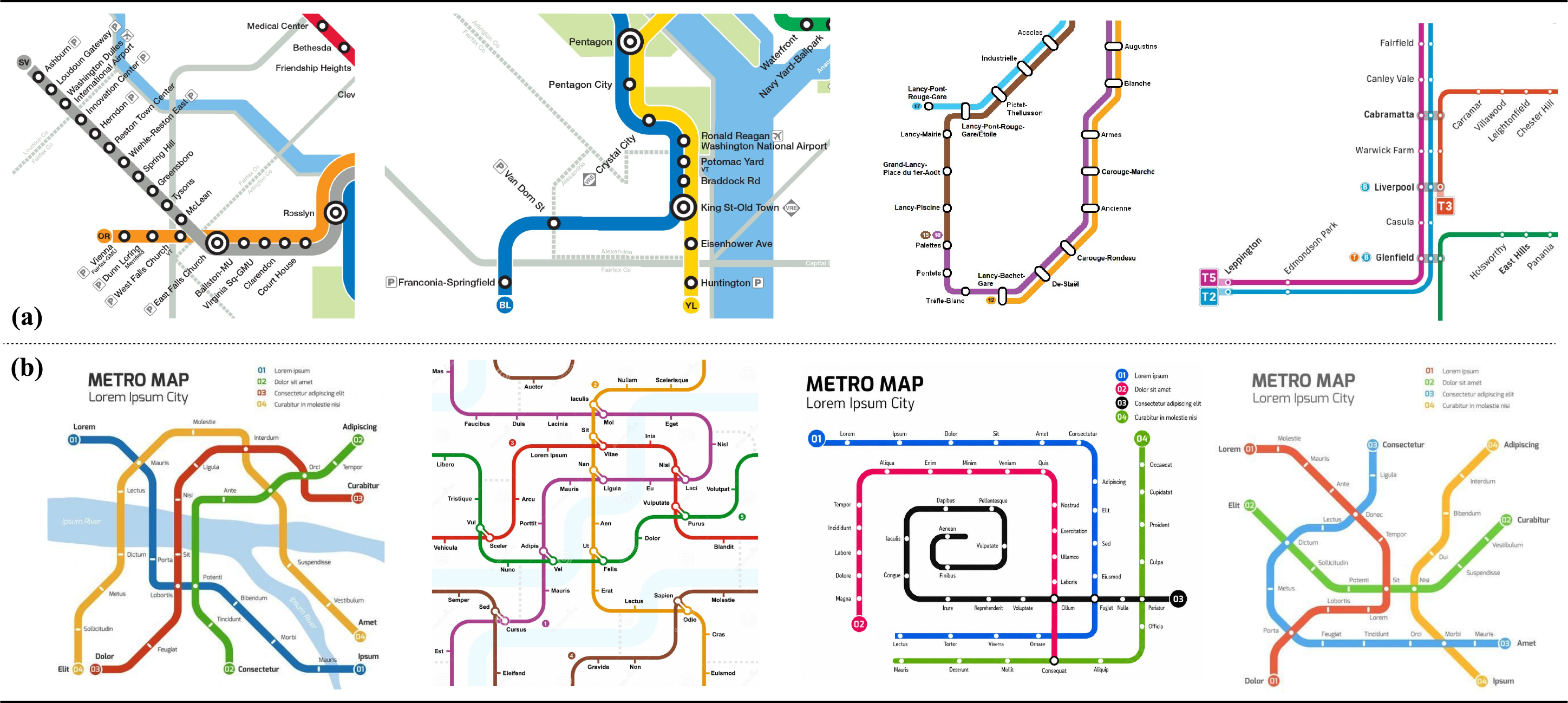}
    \caption{
    \textbf{Examples from the metro-map set.}
    The set contains (a) cropped regions from real transit maps and (b) synthetic images rendered in a metro-map style. Crops are selected to preserve local route ambiguity while avoiding regions dominated by extreme clutter or unreadable text.
    }
    \label{fig:metro_examples}
\end{figure*}

To cover both real map conventions and controlled metro-like geometries, we construct a small metro-map set from cropped public-transit maps and synthetic images. Full city transit maps are often too large and visually dense for a diagnostic tracing evaluation, so we crop local regions with moderate complexity: enough nearby routes, junctions, and labels to make line identity nontrivial, while avoiding regions dominated by extreme clutter or unreadable text. The final set contains 20 images and 50 line-tracing queries. Examples are shown in \Cref{fig:metro_examples}. In each query, the model is given a line color and a starting station, and must output the station names on that line exactly as written in the image, in traversal order.

We use the following prompt template for each metro-map query. The final answer is constrained to a station-name list; for reasoning-enabled models, we separately inspect the model-emitted reasoning trace.

\begin{tcolorbox}[
    colback=gray!5,
    colframe=black!60,
    boxrule=0.5pt,
    arc=2pt,
    left=6pt,
    right=6pt,
    top=6pt,
    bottom=6pt,
    title={Prompt for Metro Maps}
]
\small
\textbf{Task:} You must complete the sentence by listing only station names as written in the image, in sequential order, and append `end' at the very end. Do not include any non-station text.

\medskip

Visually tracing the path of the \texttt{\{line\_color\}} line from \texttt{\{start\_point\}} in the image, the sequence of stations on its path is strictly
\end{tcolorbox}

\subsubsection{Quantitative Results}
\label{app:metro_quantitative}

\begin{table}[htbp]
\centering
\small
\setlength{\tabcolsep}{4pt}
\caption{
\textbf{Reasoning accuracy and reasoning length on the metro-map set.}
Only reasoning-enabled variants are shown. Exact-match accuracy is reported over 50 queries, with raw counts in parentheses. Token counts report the average length of model-emitted reasoning traces, both overall and separately for correct and incorrect responses.
}
\label{tab:metro_reasoning}
\vspace{4pt}
\begin{tabular}{lcccc}
\toprule
Model & Accuracy (\%) & Avg. Tokens & Correct Avg. & Wrong Avg. \\
\midrule
Sonnet-4.5 & 32.0 (16/50) & 1533 & 998 & 1784 \\
GPT-5.4 & 58.0 (29/50) & 1522 & 793 & 2528 \\
Gemini3-Flash & 60.0 (30/50) & 1518 & 673 & 2785 \\
Qwen3-VL-235B & 18.0 (9/50) & 1807 & 917 & 2002 \\
GLM-4.6V & 12.0 (6/50) & 2372 & 621 & 2610 \\
Kimi-K3 & 56.0 (28/50) & 1709 & 1219 & 2333 \\
\bottomrule
\end{tabular}
\end{table}

\begin{table}[htbp]
\centering
\small
\setlength{\tabcolsep}{5pt}
\caption{
\textbf{Self-correction behavior on the metro-map set.}
Self-correction rate is reported over all 50 queries and separately over correct and incorrect responses. We use the model-specific lexical signals defined in \Cref{tab:self_correction_signals}. For Gemini3-Flash, we additionally count revision-style headers containing terms such as \textit{Revisiting}, \textit{Revising}, or \textit{Re-examining}.
}
\label{tab:metro_self_correction}
\vspace{4pt}
\begin{tabular}{lccc}
\toprule
Model & Overall Self-Corr. (\%) & Correct Self-Corr. (\%) & Wrong Self-Corr. (\%) \\
\midrule
Sonnet-4.5 & 86.0 (43/50) & 81.2 (13/16) & 88.2 (30/34) \\
GPT-5.4 & 28.0 (14/50) & 10.3 (3/29) & 52.4 (11/21) \\
Gemini3-Flash & 18.0 (9/50) & 6.7 (2/30) & 35.0 (7/20) \\
Qwen3-VL-235B & 100.0 (50/50) & 100.0 (9/9) & 100.0 (41/41) \\
GLM-4.6V & 100.0 (50/50) & 100.0 (6/6) & 100.0 (44/44) \\
Kimi-K3 & 94.0 (47/50) & 89.3 (25/28) & 100.0 (22/22) \\
\bottomrule
\end{tabular}
\end{table}



As shown in \Cref{tab:metro_reasoning}, reasoning-enabled models remain unreliable on this practical tracing task. Gemini3-Flash, GPT-5.4, and Kimi-K3 show relatively stronger performance, solving 30/50, 29/50, and 28/50 queries, respectively. Performance then drops substantially: Sonnet-4.5 solves 16/50, Qwen3-VL-235B solves 9/50, and GLM-4.6V solves only 6/50. No model exceeds 60\% exact-match accuracy. These results are consistent with the controlled and HANDLOOM evaluations: richer visual context and extended reasoning traces do not make line tracing uniformly reliable.

The reasoning length statistics suggest that long deliberations are not a reliable indicator of successful tracing. All models produce more than 1.5K reasoning tokens per query on average, and incorrect responses are consistently longer than correct responses. GLM-4.6V produces the longest traces overall while attaining the lowest accuracy, with incorrect responses averaging 2610 tokens compared to 621 for correct responses. Kimi-K3 exhibits the same direction of association, averaging 2333 tokens on incorrect responses and 1219 on correct responses. We do not interpret this as length causing failure. Rather, the pattern suggests that failures often coincide with unresolved ambiguity, repeated checking, or route-level reconstruction attempts that do not recover the correct local continuation. 

The prevalence of self-correction markers in \Cref{tab:metro_self_correction} does not show a consistent relationship with tracing accuracy. For GPT-5.4 and Gemini3-Flash, markers appear more often in incorrect responses, while GLM-4.6V and Qwen3-VL-235B use them pervasively despite their low accuracy. Kimi-K3 provides a complementary case: markers appear in 47/50 traces even though its accuracy is close to that of the best-performing models, occurring in all 22 incorrect traces as well as in 25 of 28 correct traces. Taken together, these patterns suggest that marker frequency may reflect a model-specific deliberation style rather than successful recovery of a lost path state.

\subsubsection{Qualitative Examples}
\label{app:metro_examples}

\begin{figure*}[p]
\centering
\includegraphics[width=\textwidth,height=0.82\textheight,keepaspectratio]{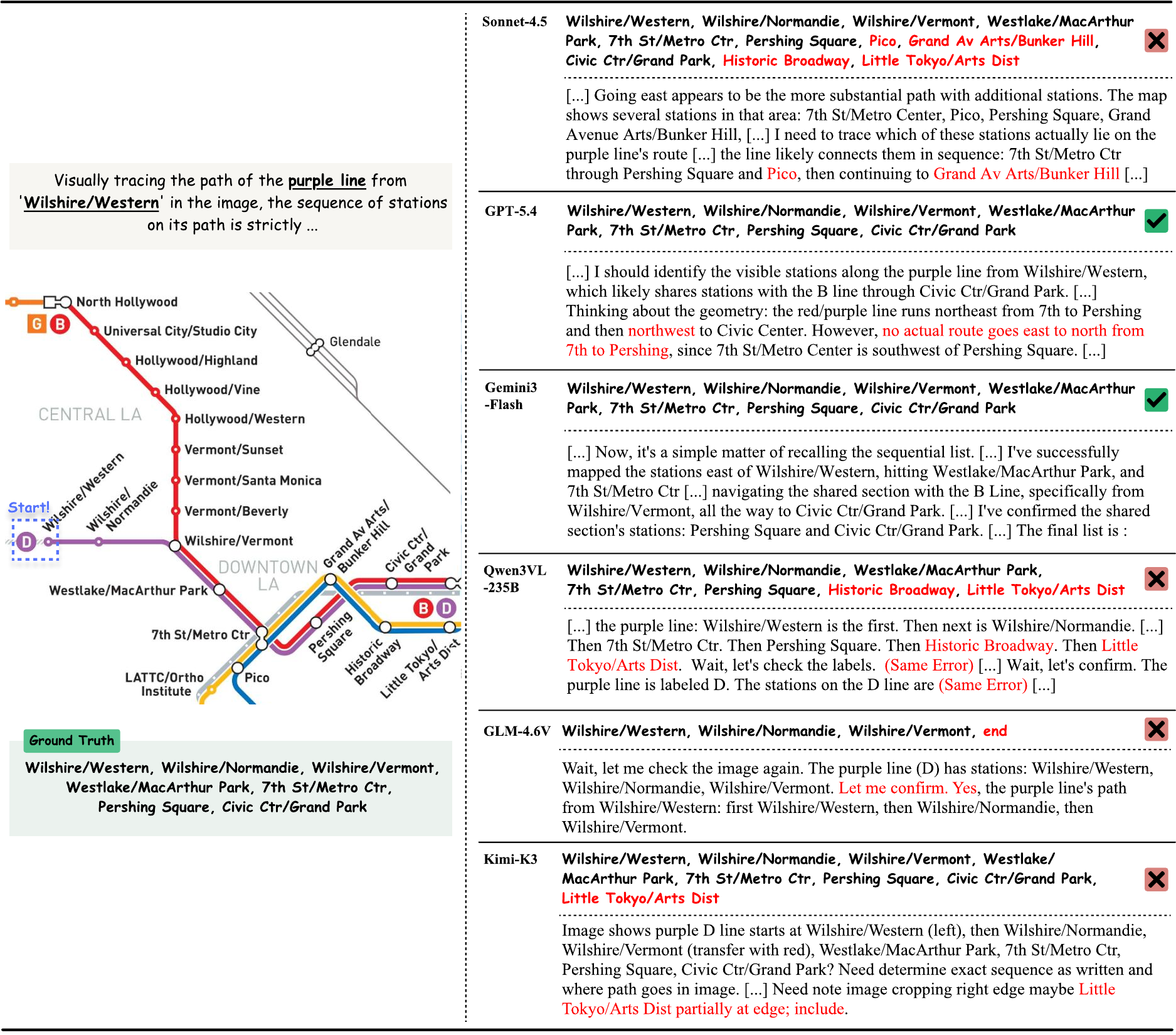}
\caption{
\textbf{Reasoning traces on a Los Angeles metro-map crop.}
Models are asked to trace the purple line from \textit{Wilshire/Western} and list the stations in order. Red text marks incorrect station names or erroneous reasoning claims. Sonnet-4.5 and Qwen3-VL-235B produce incorrect answers in the dense downtown portion of the map: Sonnet shifts from local line following to selecting among plausible nearby downtown stations, while Qwen skips a local transition and repeats the same incorrect route after self-checking. GPT-5.4 and Gemini3-Flash return the correct station sequence, but their traces are not fully explicit local traversals: GPT-5.4 includes an inconsistent geometric claim, and Gemini3-Flash assembles the answer through larger route chunks and shared-corridor cues. GLM-4.6V stops at \textit{Wilshire/Vermont} after explicitly checking and confirming the truncated sequence. Kimi-K3 reconstructs the route through \textit{Civic Ctr/Grand Park} but then extends it to \textit{Little Tokyo/Arts Dist}, despite noting uncertainty at the edge of the crop.
}
\label{fig:metro_reasoning_example_10}
\end{figure*}

\begin{figure*}[p]
\centering
\includegraphics[width=\textwidth,height=0.82\textheight,keepaspectratio]{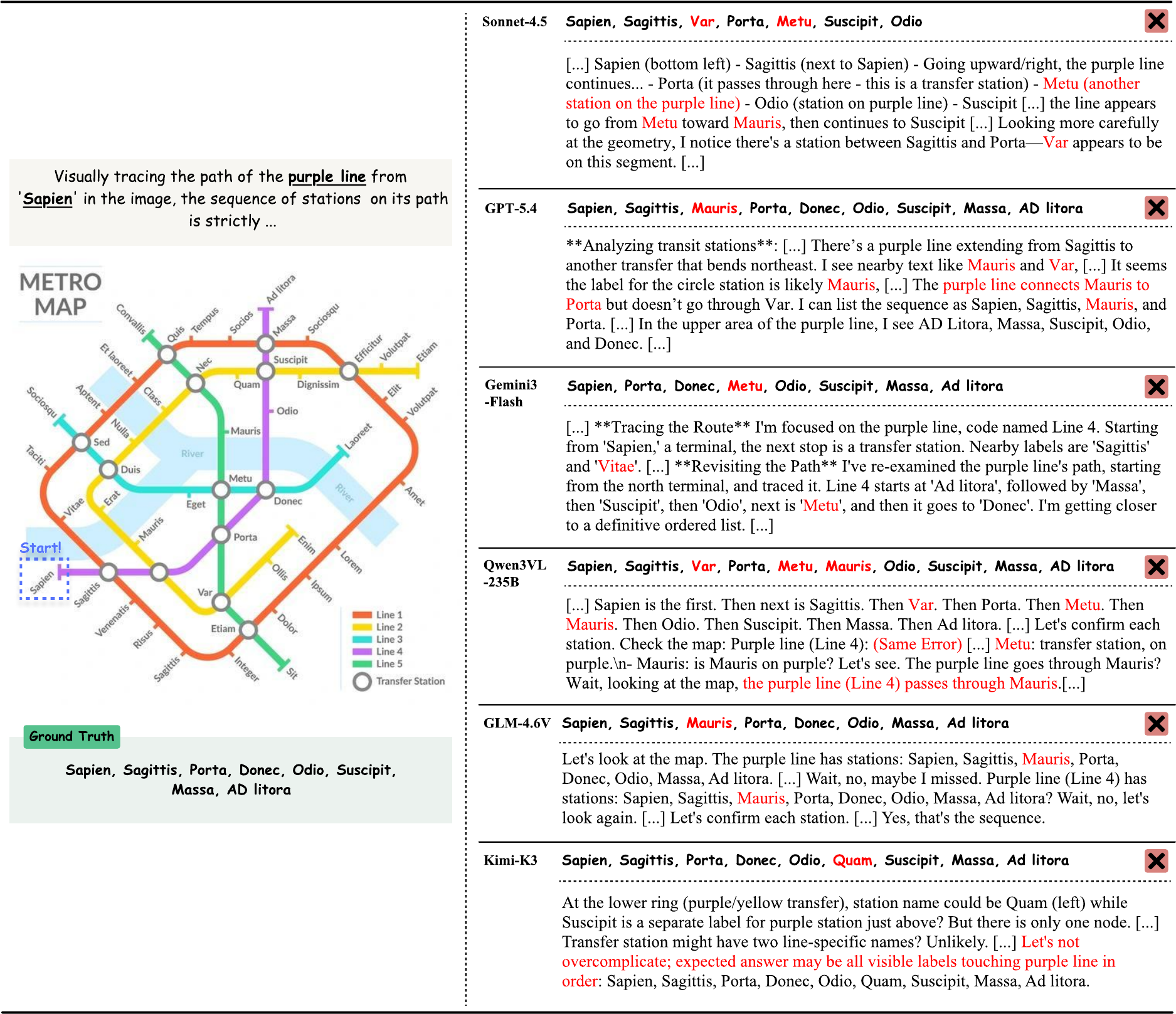}
\caption{
\textbf{Reasoning traces on a synthetic metro-map image.}
Models trace the purple line from \textit{Sapien}. Red text marks incorrect station names or erroneous reasoning claims. All six reasoning-enabled models fail on this example. Several traces are drawn toward nearby off-route stations in the central transfer region, especially \textit{Metu}: Sonnet-4.5, Gemini3-Flash, and Qwen3-VL-235B insert \textit{Metu} and other nearby labels, while GPT-5.4 mistakes nearby text such as \textit{Mauris} or \textit{Var} for the continuation after \textit{Sagittis}. Gemini3-Flash skips the early \textit{Sagittis} transition, inserts \textit{Metu}, and briefly verifies the path from the opposite terminal rather than maintaining a forward local traversal from the starting station. GLM-4.6V likewise inserts \textit{Mauris} and, despite repeatedly checking the route, omits \textit{Suscipit}. Kimi-K3 explicitly considers the ambiguity between \textit{Quam} and \textit{Suscipit} but retains both, thereby adding the off-route station \textit{Quam}. The example illustrates how plausible reasoning and explicit rechecking can still fail to preserve the queried line through nearby station-label distractors.
}
\label{fig:metro_reasoning_example_15}
\end{figure*}

\begin{figure*}[p]
\centering
\includegraphics[width=\textwidth,height=0.82\textheight,keepaspectratio]{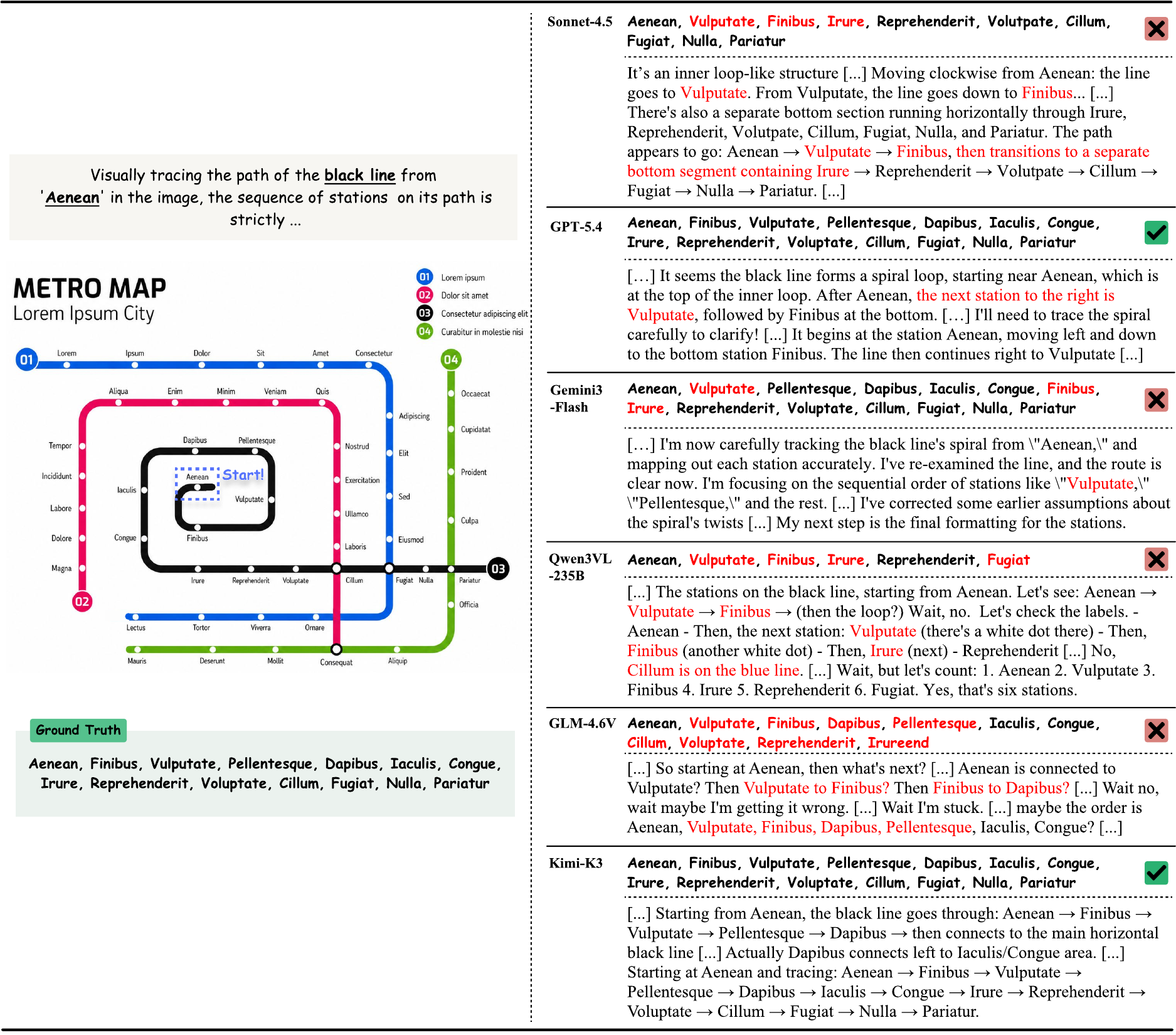}
\caption{
\textbf{Reasoning traces on a synthetic metro-map image.}
Models trace the black line from \textit{Aenean}. Red text marks incorrect station names or erroneous reasoning claims. Most models fail by taking a nearby but incorrect continuation within the inner loop: Sonnet-4.5, Gemini3-Flash, Qwen3-VL-235B, and GLM-4.6V begin by moving from \textit{Aenean} toward \textit{Vulputate} rather than the correct next station \textit{Finibus}. Their subsequent self-checks do not recover the missed local transition; GLM-4.6V in particular repeatedly questions the ordering without repairing it. GPT-5.4 and Kimi-K3 instead arrive at the correct sequence. GPT-5.4 initially describes the wrong continuation toward \textit{Vulputate} but later revisits the loop geometry, while Kimi-K3 restores the initially omitted \textit{Iaculis} after reconsidering the connection from \textit{Dapibus}. This example echoes the swirl setting: in a loop-like local structure, reasoning can be drawn to nearby plausible continuations, and only a correction grounded in the specific local connection restores faithful step-by-step tracing.
}
\label{fig:metro_reasoning_example_48}
\end{figure*}

For completeness, we include representative metro-map reasoning traces from
all six reasoning-enabled models in
\Cref{fig:metro_reasoning_example_10,fig:metro_reasoning_example_15,fig:metro_reasoning_example_48}.
The captions summarize the main patterns in each example. Broadly, the traces exhibit plausible route-level descriptions, reconstructions of shared corridors, distraction from nearby station labels, and repeated self-checks. Yet none of these behaviors reliably ensures correct station-by-station continuation along the queried line. These examples further show that path preservation failures persist in the metro-map setting, particularly around locally ambiguous continuations. Overall, these traces suggest that additional reasoning and self-checking do not by themselves resolve tracing errors; successful correction requires identifying and revising the specific local connection at which the trace went wrong.

\clearpage

\clearpage

\subsubsection{Data Sources and Licensing}
\label{app:metro_licensing}
The metro-map set combines three sources. First, we use synthetic metro-map images from \emph{BabyVision}~\citep{chen2026babyvision}, whose public dataset release lists the MIT license, and we follow its stated license and terms of use. Second, we crop a small number of real transit-map images from the \emph{RewardMap}/ReasonMap-Plus resources~\citep{feng2025rewardmap}; the ReasonMap-Plus dataset is listed under Apache License 2.0, and these crops are used only for evaluation with the original attribution and licensing information preserved. Third, we create one additional synthetic metro-map image ourselves. The dataset does not contain personal or sensitive information.

\section{LLM Usage}
\label{app:llm_usage}
We used LLM-based tools as auxiliary aids for writing, grammar and clarity edits, formatting, and routine coding support, such as debugging, refactoring, and generating boilerplate code. Their use was limited to improving presentation and workflow efficiency, and the authors retained full responsibility for the technical content, implementation, experimental results, analyses, and claims presented in the paper.

\section{Broader Impacts}
\label{app:broader_impacts}
This work studies a basic visual limitation of current vision-language models: their difficulty in maintaining a selected visual path under nearby local competition. Its potential positive impact is primarily diagnostic, as it provides a concrete way to evaluate whether VLMs can reliably follow connected visual structures in settings such as diagrams, maps, and deformable objects. The main potential negative impact comes from over-reliance on current VLMs in applications where path-switching errors could lead to incorrect downstream decisions. Our work does not introduce a deployed system, automate decisions about people, or release a model with new capabilities for misuse; rather, it highlights a limitation and suggests that VLM outputs on tracing-dependent tasks should be treated as assistive or diagnostic unless validated by task-specific checks or human review.




\end{document}